\documentclass[conference]{IEEEtran}
\usepackage{times}

\usepackage{epsfig}
\usepackage{graphicx}

\usepackage{color}
\usepackage{multirow}
\usepackage{rotating}
\usepackage{gensymb}
\usepackage{wrapfig}
\usepackage{soul}
\usepackage{booktabs}
\usepackage{graphicx}
\usepackage{amsmath,amssymb} 

\usepackage{color}
\usepackage{times}
\usepackage{epsfig}
\usepackage{epstopdf}
\usepackage{subfig}
\usepackage{bm}
\usepackage{verbatim}
\usepackage{multirow}
\usepackage{svg}

\usepackage[ruled]{algorithm2e}

\usepackage[numbers]{natbib}
\usepackage{multicol}
\usepackage[pagebackref=true,breaklinks=true,letterpaper=true,colorlinks,bookmarks=false]{hyperref}

\newcommand{\figLabel}{Figure\xspace}
\newcommand{\eqLabel}{Equation\xspace}
\newcommand{\secLabel}{Section\xspace}
\newcommand{\tblLabel}{Table\xspace}
\newcommand{\eg}{e.g.\xspace}
\newcommand{\etal}{et al.\xspace}
\newcommand{\ie}{i.e.\xspace}
\newcommand{\mysection}[1]{\vspace{3pt}\noindent\textbf{#1}}
\newcommand{\supp}{\textbf{appendix} \ref{sec: supp}\xspace}

\definecolor{turquoise}{cmyk}{0.65,0,0.1,0.1}
\definecolor{purple}{rgb}{0.65,0,0.65}
\definecolor{darkgreen}{rgb}{0.0, 0.5, 0.0}
\definecolor{darkred}{rgb}{0.5, 0.0, 0.0}
\definecolor{darkblue}{rgb}{0.0, 0.0, 0.5}
\definecolor{blue}{rgb}{0.0, 0.0, 1.0}
\definecolor{orange}{rgb}{1.0, 0.5, 0.0}

\begin{document}

\title{OIL: Observational Imitation Learning
\\\small\url{https://sites.google.com/kaust.edu.sa/oil/}}

\author{Guohao Li\authorrefmark{1}, Matthias M\"uller\authorrefmark{1}, Vincent Casser, Neil Smith, Dominik L. Michels, Bernard Ghanem\\
		Visual Computing Center, KAUST, Thuwal, Saudi Arabia\\
		{\tt\footnotesize \{guohao.li, matthias.mueller.2, vincent.casser, neil.smith, dominik.michels, bernard.ghanem\}@kaust.edu.sa}}


\maketitle

\begin{abstract}
Recent work has explored the problem of autonomous navigation by imitating a teacher and learning an end-to-end policy, which directly predicts controls from raw images. However, these approaches tend to be sensitive to mistakes by the teacher and do not scale well to other environments or vehicles. To this end, we propose Observational Imitation Learning (OIL), a novel imitation learning variant that supports online training and automatic selection of optimal behavior by observing multiple imperfect teachers. We apply our proposed methodology to the challenging problems of autonomous driving and UAV racing. For both tasks, we utilize the Sim4CV simulator \cite{sim4cv} that enables the generation of large amounts of synthetic training data and also allows for online learning and evaluation. We train a perception network to predict waypoints from raw image data and use OIL to train another network to predict controls from these waypoints. Extensive experiments demonstrate that our trained network outperforms its teachers, conventional imitation learning (IL) and reinforcement learning (RL) baselines and even humans in simulation.
\end{abstract}

\IEEEpeerreviewmaketitle

\section{Introduction} \label{sec: intro}
In the machine learning community, solving complex sequential prediction problems usually follows one of two different paradigms: reinforcement learning (RL) or supervised learning (SL), more specifically imitation learning (IL). On the one hand, the learner in conventional IL is required to trust and replicate authoritative behaviors of a teacher. The drawbacks are primarily the need for extensive manually collected training data and the inherent subjectivity to potential negative behaviors of teachers, since in many realistic scenarios they are imperfect. 
On the other hand, RL does not specifically require supervision by a teacher, as it searches for an optimal policy that leads to the highest eventual reward. However, a good reward function, which offers the agent the opportunity to learn desirable behaviors, requires tedious and meticulous reward shaping \cite{ng1999policy}. Recent methods have used  RL to learn simpler tasks without supervision \cite{DosovitskiyK16}, but they require excessive training time and a very fast simulation (\eg~$1\,000$\,fps). In this paper, we demonstrate that state-of-the art performance can be achieved by incorporating RL concepts into direct imitation to learn only the successful actions of multiple teachers. We call this approach Observational Imitation Learning (OIL). Unlike conventional IL, OIL enables learning from multiple teachers with a method for discarding bad maneuvers by using a reward based online evaluation of the teachers at training time. Furthermore, our approach allows for a modular architecture that abstracts perception and control, which allows for more flexibility when training in diverse environments with different visual properties and control dynamics. In our experiments, it is shown that this approach leads to greater robustness and improved performance, as compared to various state-of-the-art IL and RL methods. Moreover, OIL allows for control networks to be trained in a fully automatic fashion requiring no human annotation but rather can be trained using automated agents.
We demonstrate that our approach outperforms other state-of-the-art end-to-end network architectures and purely IL and RL based approaches.

\begin{figure}
\centering
\includegraphics[width=\columnwidth]{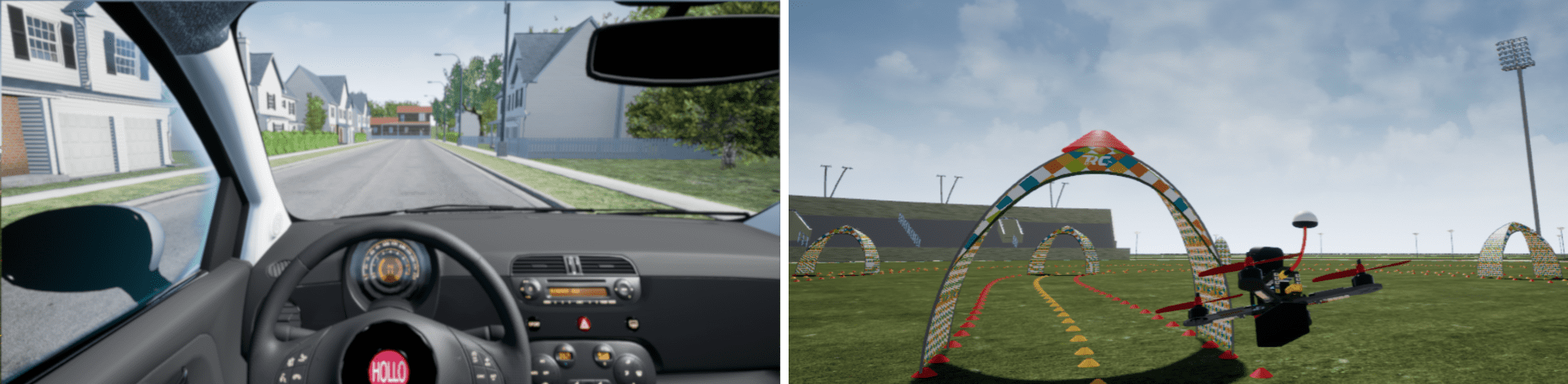}
\caption{OIL-trained Autonomous Driving  (\emph{left}) and OIL-trained UAV Racing (\emph {right}) on test tracks created using Sim4CV \cite{sim4cv}.}
\vspace{-8pt}
\label{fig:teaser}
\end{figure}

\begin{figure*}
  \includegraphics[width=18cm]{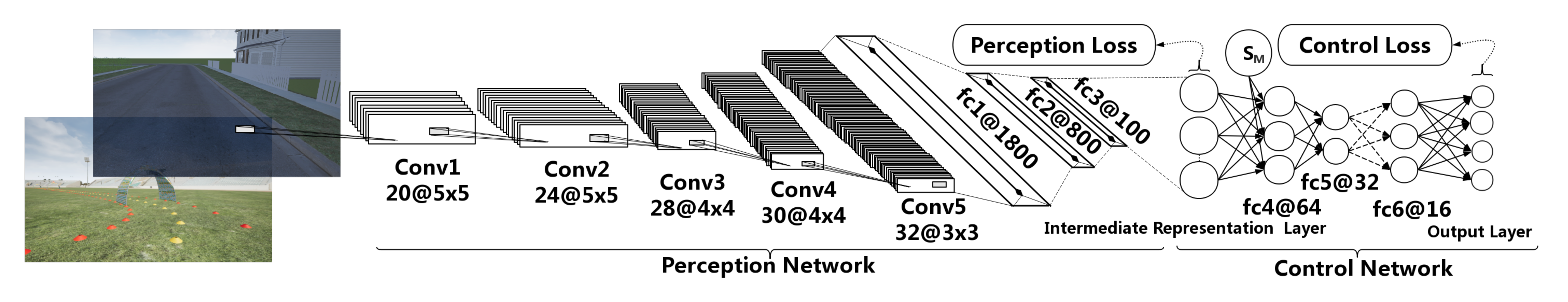}
	\caption{Pipeline of our modular network for autonomous navigation. The perception network $\phi$ takes the raw image as input and predicts waypoints. The control network $\varphi$ takes waypoints and vehicle state as input and outputs an appropriate control signal, \eg throttle (T), aileron (A), elevator (E), and rudder (R) for the UAV and only gas (E) and steering (R) for the car.}
\label{fig:pipeline}
\end{figure*}

We apply OIL to both autonomous driving and UAV racing in order to demonstrate the diverse scenarios in which it can be applied to solve sequential prediction problems.
We follow recent work \cite{DosovitskiyK16} that tests AI systems through the use of computer games. We use Sim4CV based on the Unreal Engine 4 (UE4), which has both a residential driving environment with a physics based car and gated racing tracks for UAV racing \cite{sim4cv}. 
The simulator is multi-purpose, as it enables the generation of synthetic image data, reinforcement based training in real-time, and evaluation on unseen tracks. We demonstrate that using OIL enables to train a modular neural network predicting controls for autonomous driving and the more complex task of UAV racing in the simulated Sim4CV environment. Through extensive experiments, we show that OIL outperforms its teachers, conventional IL and RL approaches and even humans in simulation.

\mysection{Contributions.} 
\textbf{(1)} We propose Observational Imitation Learning (OIL) as a new approach for training a stationary deterministic policy that overcomes shortcomings of conventional imitation learning by incorporating reinforcement learning ideas. It learns from an ensemble of imperfect teachers, but only updates the policy with the best maneuvers of each teacher, eventually outperforming all of them. 
\textbf{(2)} We use a flexible network architecture which adapts well to different perception and control scenarios. We show that it is suitable for solving complex navigation tasks (\eg autonomous driving and UAV racing). 
\textbf{(3)} To the best of our knowledge, this paper is the first to apply imitation learning to multiple teachers while being robust to teachers that exhibit bad behavior. 

\section{Related Work} \label{sec: related work}
The task of training an actor (\eg ground vehicle, human, or UAV) to physically navigate through an unknown environment has traditionally been approached either through supervised learning (SL) and in particular imitation learning (IL), reinforcement learning (RL), or a combination of the two. A key challenge is learning a high dimensional representation of raw sensory input within a complex 3D environment. Similar to our approach, many recent works such as \cite{gaidon2016virtual,GtaV,RosCVPR16,AirSim,UE4simulator,BlukisBBKA18} use modern game engines or realistic physics-based simulators to evaluate complex navigation tasks.

\mysection{Imitation Learning (IL).} 
In physics based navigation, IL can be advantageous when high dimensional feedback can be recorded. It has been applied to both autonomous driving \cite{NvidiaCar,deepDriving,ForestTrail,Xu_2017_CVPR,Pan_RSS2018} and UAV navigation \cite{ForestTrail,pmlr-v87-kaufmann18a,TeachingUAVstoRace,BlukisBBKA18,SpicaFCMSS18}. 
In particular, DAGGER \cite{Dagger} and its variants \cite{BlukisBBKA18} \cite{Pan_RSS2018} have been widely used for many robotic control tasks. However, limitations such as requirements of significant fine-tuning, inability to predict human-like controls, data augmentation being only corrective and the requirement for near-optimal expert data makes DAGGER hard to scale up to more complex problems. Moreover, the flaws and mistakes of the teachers are learned along with differing responses. See Section \ref{sec: baselines} for the comparison of two popular IL methods (Behavioral Cloning \cite{torabi2018behavioral} and DAGGER) to OIL. Follow-up work such as AGGREVATE \cite{ross2014reinforcement} and Deeply AggreVaTeD \cite{sun2017deeply} attempt to mitigate these problems by introducing exploratory actions and measuring actual induced cost instead of optimizing for expert imitation only. They also claim  exponentially higher sample efficiency than many classic RL methods. A number of improvements in other respects have been published, such as SafeDAgger \cite{zhang2016query} that aims to make DAGGER more (policy) query-efficient and LOLS \cite{chang2015learning} that aims to improve upon cases where the reference policy is sub-optimal.

\mysection{Reinforcement Learning (RL).} 
RL provides an alternative to IL by using rewards and many iterations of exploration to help discover the proper response through interactive trial and error. Recent work on autonomous driving has employed RL \cite{mnih2016asynchronous, deepReinforcementSimulator,Koutnik2014,Sallab2017, Sallab2016EndtoEndDR,Iros17_IseleCSF17,carla-dosovitskiy17}. RL networks may not be able to discover the optimal outputs in higher-order control tasks. For example, Dosovitskiy \etal \cite{carla-dosovitskiy17} find RL to under-perform in vehicle navigation due to the extensive hyperparameter space. RL methods can be divided into three classes: value-based, policy-based, and actor-critic based methods \cite{sutton1998introduction}. In particular, actor-critic based methods, \eg A3C \cite{mnih2016asynchronous} and DDPG \cite{deepReinforcementSimulator}, are notably the most popular algorithms in the RL community. However, achieving strong results with RL is difficult, since it is very sensitive to the reward function, it can be sample inefficient, and it requires extensive training time due to the large policy space (see Section \ref{sec: baselines} for comparison of DDPG to OIL). Methods such as TRPO \cite{schulman2015trust} have been developed to provide monotonic policy improvements in most cases, but still require extensive training time.

\mysection{Combined Approaches.}
Several methods exist that combine the advantages of IL and RL. Most of them focus on tackling the problem of low RL sample efficiency by providing suitable expert demonstrations (\cite{liang2018cirl,Zhuetal_RSS2018,Taylor:2011,Rajeswaran-RSS-18,Gao_RLimperfect,Yuziang2018}) and guided policy search (\cite{guidedpolicysearch,DosovitskiyK16,Gimelfarb_NIPS2018,Zhuetal_RSS2018}). Others focus on risk awareness as real-world deployment failures can be costly \cite{Andersson2017}.
Generative adversarial imitation learning \cite{GAIL_NIPS2016} avoids the costly expense of IRL by directly learning a policy from supplied data. It explores randomly to determine which actions bring a policy closer to an expert but never directly interacts with the expert as in DAGGER. The authors note a combination of both random search and interaction with the expert would lead to better performance, an insight employed in our approach. Similar to our approach, \cite{Gimelfarb_NIPS2018} apply Bayesian Model Combination to guide the policy to learn the best combination of multiple experts.
The recent CFN \cite{mller2019learning} algorithm is the most related method to our approach; trajectories of multiple controllers are filtered and then fused by a neural network in order to obtain a more robust controller.

We draw inspiration for OIL from these hybrid approaches. In contrast to pure IL, OIL can prevent itself from learning bad demonstrations from imperfect teachers by observing teachers' behaviours and estimating the advantage or disadvantage to imitate them. Unlike pure RL, it converges to a high performance policy without too much exploration since it is guided by the best teacher behaviors. While sharing the advantage of higher sample efficiency with other hybrid approaches, our method has the specific advantage of inherently dealing well with bad demonstrations, which is a common occurrence in real-world applications.

\section{Methodology} \label{sec: methodology}
After providing a brief review of related learning strategies for sequential decision making (\ie Markov Decision Process, Imitation Learning, and Reinforcement Learning), we introduce our proposed Observational Imitation Learning (OIL), which enables automatic selection of the best teacher (among multiple teachers) at each time step of online learning. 

\subsection{Markov Decision Process}
OIL is a method that enables a learner to learn from multiple sub-optimal or imperfect teachers and eventually to outperform all of them. To achieve this goal, training needs to be done by repeatedly interacting with an environment $\mathcal{E}$. We consider the problem as a Markov Decision Process (MDP) consisting of an agent and environment. At every time step $n$, the agent observes a state $s_n$ or a partial observation $o_n$ of state $s_n$. In our setting, we assume the environment is partially-observed but we use $s_n$ to represent the state or the partial observation for simplicity. Given  $s_n$, the agent performs an action $a_n$ within the available action space $\mathcal{A}$ based on its current policy $\pi$, where $\theta$ represents the parameters of the policy. Then, the environment $\mathcal{E}$ provides the agent a scalar reward $r_n$ according to a reward function $R(s_n, a_n)$ and transfers to a new state $s_{n+1}$ within the state space $\mathcal{S}$ under the environment transition distribution $p(s_{n+1}|s_n, a_n)$.
After receiving an initial state $s_1$, the agent generates a trajectory $\tau = (s_1, a_1, s_2, a_2, ..., s_N)$ after $N$ time steps. The trajectory can end  after a certain number of time steps or after the agent reaches a goal or terminal state. The objective of solving a MDP problem is to find an optimal policy $\pi^{*}$ from policy space $\Pi$ that maximizes the expected sum of discounted future rewards, $R_n$ at time $n$. It is commonly referred to as the value function:
\begin{equation}
\begin{split}
	V^{\pi}(s) &= \mathbb{E}\left[R_n\bigr{|}s_n=s, \pi\right] \\
	&= \mathbb{E}\left[\sum_{n^{\prime}=n}^{N} \gamma^{n^{\prime}-n}r_{n^{\prime}} \bigr{|}s_n=s, \pi\right],
\end{split}
\end{equation}
where $\gamma \in [0,1]$ is a discounted factor that trades off the importance of immediate and future rewards, and $N$ is the time step when the trajectory $\tau$ terminates. The optimal policy $\pi^{*}$  maximizes the value function for all $s \in \mathcal{S}$:
\begin{equation}
\begin{split}
	 \pi^{*} &= \underset{\pi \in \Pi}{\mathrm{argmax}}~~ V^{\pi}(s)
\end{split}
\end{equation}

\subsection{Imitation Learning}\label{sec: IL}
Imitation learning (IL) is a supervised learning approach to solve sequential decision making problems by mimicking an expert policy $\pi^{e}$. Instead of directly optimizing the value function, IL minimizes a surrogate loss function  $\mathbb{L}(s, \pi, \pi^{e})$. Let $d_\pi$ denote the average distribution of visited observations when an arbitrary policy $\pi$ is executed for $T$ time steps. Behavioral Cloning, one of the simplest IL algorithms, trains a learner policy network $\pi^{l}$ to fit the input (observations) and output (actions) of the expert policy by minimizing the surrogate loss \cite{Dagger}:
\begin{equation}
\begin{split}
	\pi^{l} = \underset{\pi \in \Pi}{\mathrm{argmin}} ~~\mathbb{E}_{s\thicksim d_{\pi^{e}}}[\mathbb{L}(s, \pi, \pi^{e})]
\end{split}
\end{equation}
However, this leads to poor performance because the encountered observation spaces of the learner and the expert are different, thus violating the independent, identically distributed (i.i.d.) assumption of statistical learning approaches and causing compounding errors \cite{ross2010efficient}. DAGGER (Dataset Aggregation) \cite{Dagger} alleviates these  errors in an iterative fashion by collecting the state-action pairs visited by the learned policy, but labeled by the expert. Its goal is to find a policy  $\pi^{l}$ that minimizes the surrogate loss under the observation distribution $d_{\pi}$ induced by the current policy $\pi$:
\begin{equation}
\begin{split}
	\pi^{l} = \underset{\pi \in \Pi}{\mathrm{argmin}} ~~\mathbb{E}_{s\thicksim d_{\pi}}[\mathbb{L}(s, \pi, \pi^{e})]
\end{split}
\end{equation}
A major drawback of DAGGER is that it highly depends on the expert's performance, while a near-optimal expert is hard to acquire for most tasks. 

\subsection{Reinforcement Learning}\label{sec: RL}
Reinforcement Learning (RL) is another category of methods to solve the MDP problem by trial and error. RL methods can be divided into three classes: value-based, policy-based, and actor-critic-based methods \cite{sutton1998introduction}. Specifically, actor-critic-based methods, \eg A3C \cite{mnih2016asynchronous} and DDPG \cite{deepReinforcementSimulator}, are currently the most popular algorithms in the RL community. The most related actor-critic method to our work is the Advantage Actor-critic approach \cite{mnih2016asynchronous}, which uses an advantage function for policy updates instead of the typical value function or Q-function \cite{mnih2015human}. Intuitively, this advantage function evaluates how much improvement is obtained if action $a$ is taken at state $s$ as compared to the expected value. It is formally defined as follows \cite{peters2005natural}:
\begin{equation} \label{rl_ad}
\begin{split}
	A^{\pi}(s, a) &= Q^{\pi}(s, a) - V^{\pi}(s)
\end{split}
\end{equation}

\subsection{Observational Imitation Learning (OIL)}
As discussed in \secLabel \ref{sec: IL}, imitation learning requires a near-optimal teacher and extensive augmentation for exploration. Getting labeled expert data is expensive and not scalable. While RL approaches (\secLabel \ref{sec: RL}) do not require supervision and can freely explore an environment, they are time-consuming and may never learn a good policy unless the reward is very well designed. In an effort to combine the strengths of both approaches we propose \emph{Observational Imitation Learning} (OIL). Inspired by advantage actor-critic learning \cite{mnih2016asynchronous}, we learn to imitate only the best behaviours of several sub-optimal teachers. We do so by estimating the value function of each teacher and only keeping the best to imitate. As a result, we can learn a policy that outperforms each of its teachers. Learning from multiple teachers also allows for exploration, but only of feasible states, leading to faster learning than in RL.

Since we do not require expert teachers we can obtain labeled data much more cheaply. We assume easy access to $K$ cheap sub-optimal teacher policies (\eg simple PID controllers). We denote the teacher policy set as $\Pi^{t} = \{\pi^{t_1}, \pi^{t_2}, ..., \pi^{t_K}\}$, where $\pi^{t_k}$ is the teacher policy corresponding to teacher $k$ and denote the learner policy as $\pi^{l}$. The advantage function of current learner policy $\pi^{l}$ compared to teacher policy $\pi^{t}$ at state $s$ can be written as:
\begin{equation} \label{il_ad}
\begin{split}
	A(s, \pi^{l}, \pi^{t}) = V^{\pi^{l}}(s) - V^{\pi^{t}}(s)
\end{split}
\end{equation}

The advantage function $A(s, \pi^{l}, \pi^{t})$ determines the advantage of taking the learner policy $\pi^{l}$ at state $s$ compared to the teacher policy $\pi^{t}$ where $\pi^{l}$ and $\pi^{t}$ are analogous to an actor and critic policy respectively. Note that the advantage function in regular RL settings is used to cope with the learner policy. In contrast, our advantage function \eqLabel\ref{il_ad} considers both the learner policy and the teacher policy. 
In multi-teacher scenarios, we select the most critical teacher $\pi^{t_{*}}$ as the critic policy (refer to Algorithm \ref{alg: OIL}).

We define our training in terms of observation rounds, each consisting of three phases: \emph{observing}, \emph{rehearsing}, \emph{acting}. 

\mysection{Observe.}
We estimate the value functions $V^{\pi}(s)$ of the learner policy as well as of all teacher policies using Monte-Carlo sampling (rolling out the trajectory of each policy to get the returns $R_t$). We select the most critical teacher policy $\pi^{t_{*}} = \underset{\pi^{t} \in \Pi^{t}}{\mathrm{argmax}} ~~ V^{\pi^{t}}(s)$ as the critic. Then we compute the advantage function $A(s_1, \pi^{l}, \pi^{t_{*}})$ between the learner policy and the most critical teacher policy. If the advantage $A(s_1, \pi^{l}, \pi^{t_{*}}) < 0$ (\ie there exists a sub-optimal teacher policy with a higher advantage than the current learner policy), we enter the \emph{rehearsing} phase. Otherwise, we go to the \emph{acting} phase directly. 

\mysection{Rehearse.}
After computing $A(s_1, \pi^{l}, \pi^{t_{*}})$, we can optimize the policy by actor-critic methods or by optimizing the surrogate loss. In order to benefit from the fast convergence of IL instead of optimizing the advantage function directly, we optimize the surrogate loss in \eqLabel \ref{suro loss} iteratively as follows:
\begin{equation} \label{suro loss}
\begin{split}
	\pi^{l(i+1)} = \underset{\pi \in \Pi}{\mathrm{argmin}} ~~\mathbb{E}_{s\thicksim d_{\pi^{l(i)}}}[\mathbb{L}(s, \pi^{l(i)}, \pi^{t_{*}})]
\end{split}
\end{equation}
where $\pi^{l(i)}$ is the learner policy at the $i$-th iteration. In our implementation, we use a DNN to represent our learner policy as $\pi^{l}(s|\theta)$, where $\theta$ represents the parameters of the neural network. In order to minimize the surrogate loss, we roll out the learner (actor) policy and use the selected teacher (critic) to correct the learner's actions. In other words, we minimize the surrogate loss with states encountered by the learner policy and actions labeled by the most critical teacher policy. We minimize the surrogate loss by performing gradient descent with respect to $\theta$ on collected data $\mathcal{D}$ and update the learner policy until $A(s_1, \pi^{l}, \pi^{t_{*}}) > \epsilon$ or $I$ episodes. In practice, the advantage function $A$ can be estimated using \emph{Monte Carlo Methods} or \emph{Bootstrapping Methods}. In our experiment, we use \emph{Monte Carlo Methods} to estimate the advantage function by rolling out policies (see \secLabel \ref{para:training}).

\mysection{Act.}
After rehearsing, the learner policy will perform well from the initial state $s_1$; we roll out the current policy $\pi^{l}$ to the new state $s_1^{\prime}$ by acting $J$ steps.

\begin{algorithm}[t] 
\caption{Observational Imitation Learning (OIL).}\label{alg: OIL}
\small
\SetAlgoLined
 Initialize Learner training database $\mathcal{D} \leftarrow \emptyset$\;
 Initialize Learner network $\pi^{l}(s|\theta)$ with random weights $\theta$\;
 \For{observation round m $\leftarrow 1$ \KwTo $M$}{
   Receive initial state $s_1$ from the environment\;
   Estimate learner value function $V^{\pi^{l}}(s_1)$ \;
   Estimate teacher value functions $V^{\pi^{t}}(s_1)$, $\forall \pi^{t} \in \Pi^{t}$\;
   Choose $\pi^{t_{*}} = \underset{\pi^{t} \in \Pi^{t}}{\mathrm{argmax}}~~ V^{\pi^{t}}(s_1)$\;
   Compute advantage function $A(s_1, \pi^{l}, \pi^{t_{*}})$\;
   \While{$A(s_1, \pi^{l}, \pi^{t_{*}}) < 0$}{
        \Repeat{$A(s_1, \pi^{l}, \pi^{t_{*}}) > \epsilon$ or repeat $I$ episodes}{
            Sample N-step trajectories using learner policy\;
            \While{$n<N$}{
                \Repeat{$s_n$ is a terminal state}{
                    $a_n \thicksim  \pi^{l}(s_n|\theta) $\;
                    Take action $a_n$, observe $r_n$\, $s_{n+1}$ \;
                    Add state-action $(s_n, \pi^{t_*}(s_n))$ to $\mathcal{D}$\;
                    Update $\pi^{l}(s|\theta)$ by minimizing $\mathbb{L}(s, \pi^{l}, \pi^{t_{*}})$\ from $\mathcal{D}$\;
                    $s_n \leftarrow s_{n+1}$\;
                }
            }
        }
    }
    Sample $s_1^{\prime}$ by acting updated $\pi^l$ policy $J$ steps\;
    $s_1 \leftarrow s_1^{\prime}$
 }
\end{algorithm}

\section{Network and Training Details}
In this section we present a modular network architecture based on CFN \cite{mller2019learning} for autonomous driving and UAV racing that solves the navigation task as a high dimensional perception module (trained with automatically annotated data) and a low dimensional control module (trained with OIL).

\subsection{Modular Architecture}
The fundamental modules of our proposed system are summarized in \figLabel\ref{fig:pipeline}. We use a modular architecture \cite{pmlr-v87-kaufmann18a, mueller18a, mller2019learning} to reduce the required training time by introducing an intermediate representation layer. The overall neural network consists of two modules: a \emph{Perception Network} $\phi$ and a \emph{Control Network} $\varphi$. The input state includes image $\mathcal{I}$ and the physical measurements $\mathcal{M}$ of the vehicle's state (\ie current orientation and velocity). The action is a control signal for the car ($G$: Gas/Brake, $S$: Steering) or UAV ($T$: Throttle, $A$: Aileron/Roll, $E$: Elevator/Pitch, $R$: Rudder/Yaw). One scalar value is regressed for each control signal. The perception network is parameterized by $\theta_p$ and the control network is parameterized by $\theta_c$. The control network takes the intermediate representation predictions of the perception network and vehicle's state as input and outputs the final control predictions. In our experiments, we choose waypoints as the intermediate representation. Note that it can also be segmentation maps \cite{mueller18a}, depth images, affordances \cite{deepDriving} \cite{sauer2018conditional} or the combination of them. The overall navigation policy can be described as follows:
\begin{equation} \label{eq2}
\begin{split}
	\pi^{l}(s|\theta) = \pi^{l}(s|\theta_p,\theta_c) = \varphi(\phi(s_\mathcal{I}|\theta_p),s_\mathcal{M}|\theta_c)
\end{split}
\end{equation}

The overall loss is defined in \eqLabel\ref{eq3} as a weighted sum of perception and control loss. The perception loss relates to the intermediate layer by minimizing the difference between the ground truth intermediate representation $z^{*}$ and predicted intermediate representation $\phi(s_\mathcal{I}|\theta_p)$. The control loss in \eqLabel\ref{eq4} comes from applying OIL to learn from multiple imperfect teachers (automated PID controllers in our case) by minimizing the surrogate loss in \eqLabel\ref{suro loss}, where $s\thicksim d_{\pi^{l}}$ and $a^{*} \thicksim \pi^{t_{*}}$.
\begin{align} 
&\mathbb{L} = \underbrace{ \mathbb{L}_c(\pi(s|\theta),a^{*}) }_{\text{control loss}} + \lambda \underbrace{\mathbb{L}_p(\phi(s_\mathcal{I}|\theta_p),z^{*}) }_{\text{perception loss}}
\label{eq3}\\
&\mathbb{L}_c(\pi(s|\theta),a^{*}) = \mathbb{L}_c(\varphi(\phi(s_\mathcal{I}|\theta_p),s_\mathcal{M}|\theta_c),a^{*})\label{eq4}
\end{align}

In general, this optimization problem can be solved by minimizing the overall loss with respect to $\theta_p$ and $\theta_c$ at the same time. The gradients are as follows:
\begin{align} 
\frac{\partial \mathbb{L}}{\partial \theta_p} &= \frac{\partial \mathbb{L}_c}{\partial \theta_p} + \lambda \frac{\partial \mathbb{L}_p}{\partial \theta_p}=\frac{\partial \mathbb{L}_c}{\partial \varphi}\frac{\partial \varphi}{\partial \phi}\frac{\partial \phi}{\partial \theta_p} + \lambda \frac{\partial \mathbb{L}_p}{\partial \phi}\frac{\partial \phi}{\partial \theta_p} \label{eq5}\\
\frac{\partial \mathbb{L}}{\partial \theta_c} &= \frac{\partial \mathbb{L}_c}{\partial \varphi}\frac{\partial \varphi}{\partial \theta_c}\label{eq6}
\end{align}

A good perception network is essential to achieve good controls. To maintain modularity and reduce training time, we first optimize only for $\theta_p$ and ignore the control loss. As the perception network converges, we fix $\theta_p$ and optimize for $\theta_c$.

This modular approach has several advantages over an end-to-end approach (see also \cite{pmlr-v87-kaufmann18a, mueller18a, mller2019learning}). Since only the control module is specific to the vehicle's dynamics, the perception module can simply be swapped out, allowing the vehicle to navigate in completely different environments without any modification to the control module. Similarly, if the reward function is changed in different tasks, we can simply retrain the control module to learn a different policy. Naturally, one can add links between the perception and control networks and then finetune the joint network in an end-to-end fashion. It is also possible to connect the two networks and use the waypoint labels as intermediate supervision for the perception part while training the joint model end-to-end. While these variants are interesting, we specifically refrain from such connections to safeguard the attractive modular properties.

In what follows, we provide details of the architecture, implementation, and training procedure for both the perception and control modules. Note that OIL and the proposed architecture can also be applied to other types of MDP problems (\eg other vision-based sequential decision making problems).

\subsection{Perception} \label{sec: perception}
In our case the perception module takes raw RGB images as an input and predicts a trajectory of waypoints relative to the vehicle's current position, which remains unknown in 3D. The waypoints predicted by the perception module are input into the control network along with the current vehicle state (velocity and orientation). 

\mysection{Drawbacks of Predicting Controls Directly.~} Note that related work commonly proposed to frame the navigation task as an end-to-end learning problem, predicting controls directly from single images. However, this has several limitations: 

\textbf{(i)} A teacher is required for data collection and the controls are strongly dependent on the teacher and vehicle dynamics, as opposed to our modular DNN. \textbf{(ii)} There is no unique mapping from images to controls, since different sequences of controls can lead to the same image. This can result in direct discrepancies and less stable training, if data from multiple teachers is used. \textbf{(iii)} When using camera views in addition to the views used for data acquisition (a common augmentation method to increase robustness), it is unclear how the controls corresponding to these augmented views should be adjusted, given the underlying complex nature of the task. For the case of driving, it might be sufficient to only predict a steering angle and acceleration. Since the car is confined to a 2D plane (road) and the friction of the tires limits drift, one can design a highly simplified model, \eg by offsetting the steering by the rotation of the augmented camera view. However, for more complex scenarios where the vehicle moves in 3D and is less constrained by friction, such simplifying assumptions quickly become invalid.

\begin{figure}[!thb]
   \centering
	   \includegraphics[width=\linewidth]{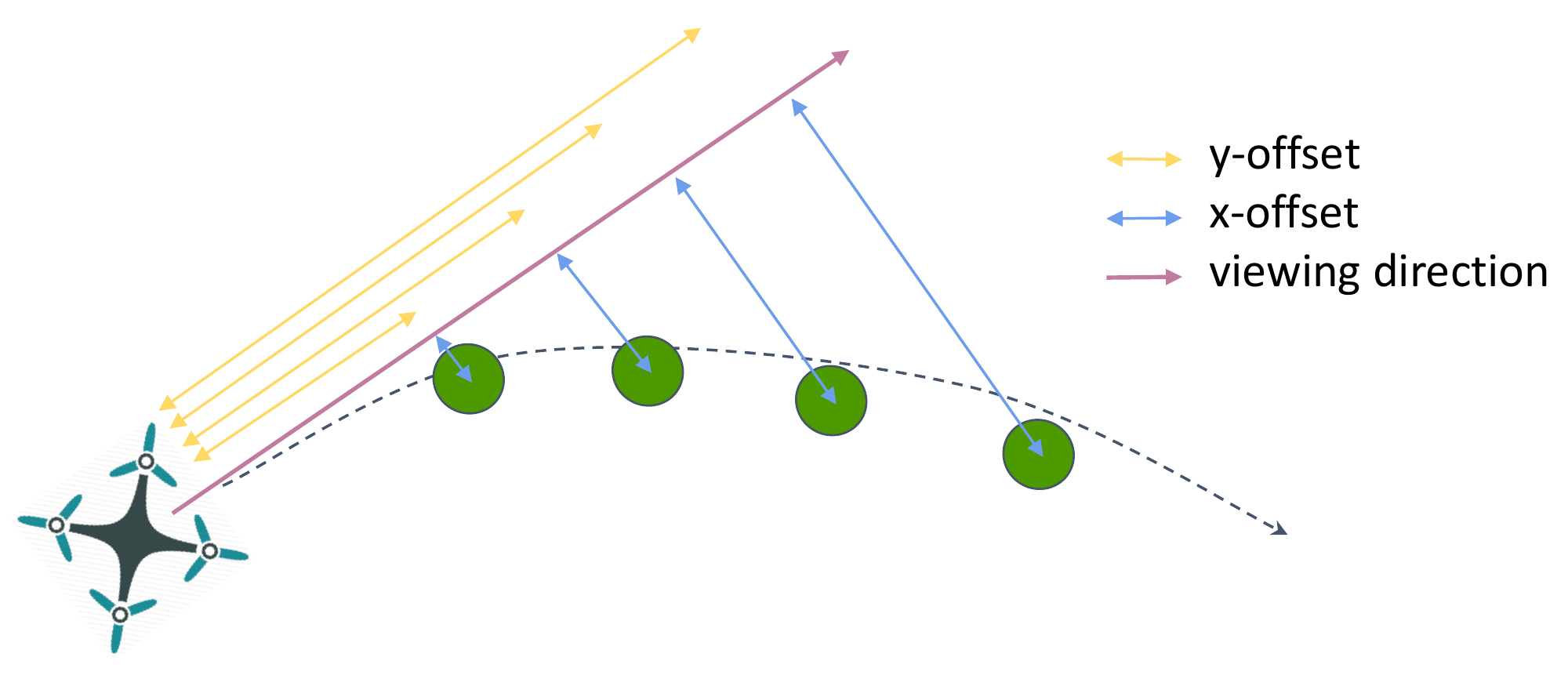}
     \caption{Illustration of waypoint encoding.}
     \label{fig:wp_encoding}
\end{figure}

\mysection{Waypoint Encoding.~} The mapping from image to waypoints is deterministic and unique. For every camera view, the corresponding waypoints can easily be determined and are independent of the vehicle state. We define waypoints along the track as a vertical offset that is measured as the distance between the vehicle position and the projected point along the viewing axis, and a horizontal offset that is defined as the distance between the original and projected point along the viewing axis normal. We then encode these waypoints relative to the vehicle position and orientation by projecting them onto the viewing axis. Predicting waypoints rather than controls does not only facilitate network training, but it also allows for the automatic collection of training data without human intervention. Within the simulator, we simply sample/render the entire training track from multiple views and calculate the corresponding waypoints along the track. Note that it is still possible to use recordings from teachers, as one can use future positions to determine the waypoint for the current frame similar to \cite{Xu2016EndtoendLO}. Please refer to \figLabel \ref{fig:wp_encoding} for a visualization of the encoding method.

\mysection{Network Architecture.~} We choose a regression network architecture similar to the one used by Bojarski \etal\cite{NvidiaCar}. 
Our DNN architecture is shown in \figLabel \ref{fig:pipeline} as the perception module. It consists of eight layers: five convolutional with $\{20,24,28,30,32\}$ filters and three fully-connected with $\{1800,800,100\}$ hidden units. 
The DNN takes in a single RGB-image with $180\times320$ pixel resolution and is trained to regress the next five waypoints ($x$-offset and $y$-offset with respect to the local position of the vehicle) using a standard $\ell_2$-loss and dropout ratio of $0.5$ in the fully-connected layers. As compared to related methods \cite{NvidiaCar,ForestTrail}, we find that the relatively high input resolution is useful to improve the network's ability to look further ahead, and increasing long-term trajectory stability. For more details and a visualization of the learned perception network, please refer to the \supp. 

\subsection{OIL for Control} \label{sec: controller}
Here, we present details of the network architecture and learning strategy to train the control network using OIL. 

\mysection{Teachers and Learner.~} In our experiments, we use multiple naive PID controllers as teachers for our control policy. The PID parameters are tuned within a few minutes and validated on a training track to perform well. As the system is very robust to learn from imperfect teachers, we do not need to spend much effort tuning these parameters or achieve optimal performance of the teachers on all training tracks. Although an unlimited number of PID based teachers can be created, we empirically find five to be sufficient for the two control scenarios (autonomous driving and UAV racing). We refer to our evaluation section to see the effect of the number of teachers on learning. The five PID teachers have different properties. Some of them are fast at straightaways but are prone to crash in tight turns. Others drive or fly more conservatively and are precise at curves but slow at straightaways as a result. OIL enables the agent to learn from such imperfect teachers by selecting only the best maneuvers among them to learn.

We use a three-layer fully connected network to approximate the control policy $\varphi$ of the learner. The MDP state of the learner is a vector concatenation of the predicted intermediate representation (\eg waypoints) $\hat{z} = \phi(s_\mathcal{I}|\theta_p)$ and the vehicle state (physical measurements of the vehicle's speed and orientation) $s_\mathcal{M}$.

\mysection {Network Architecture.~} The goal of the control network is to find a control policy $\varphi$ that minimizes the control loss $\mathbb{L}_c$:
\begin{align} 
\varphi^* &= \underset{\varphi}{\arg\min}~~ \mathbb{L}_c(\varphi(\hat{z}, s_{\mathcal{M}}|\theta_c),a^{*}) \label{eq9}
\end{align}
It consists of three fully-connected layers with hidden units $\{64, 32, 16\}$, where we apply dropout in the second layer with a ratio of $0.5$ for regularization. The loss function $\mathbb{L}_c$ is a standard $\ell_2$-loss optimized by the Adam optimizer with a learning rate of $10^{-4}$. The control network is updated by OIL in an online fashion, while the vehicle runs through a set of training tracks.

As such, the control network is learning from experiences (akin to reinforcement learning), but it is supervised throughout by multiple teachers to minimize the surrogate loss. An advantage of our approach is that  multiple teachers are able to teach throughout the control network's exploration. The control network never becomes dependent on the teachers, but gradually becomes independent and eventually learns to outperform them. 

\section{Experiments} \label{sec: results}

\begin{figure*}
\centering
\begin{tabular}{cccc}
\includegraphics[width=0.22\textwidth]{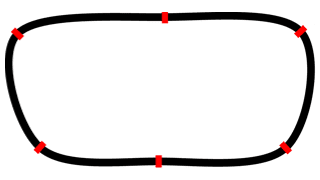} &
\includegraphics[width=0.22\textwidth]{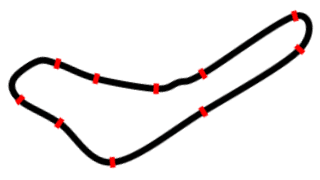} &
\includegraphics[width=0.22\textwidth]{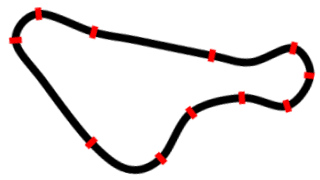} &
\includegraphics[width=0.22\textwidth]{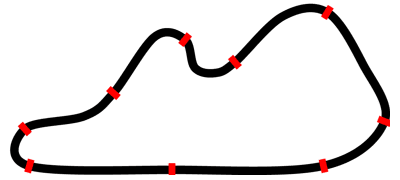}
\end{tabular}
\caption{Tracks used to evaluate the flying policy.}
\label{fig:tracks}
\end{figure*}

\begin{figure*}
\centering
\begin{tabular}{cccc}
\includegraphics[width=0.22\textwidth]{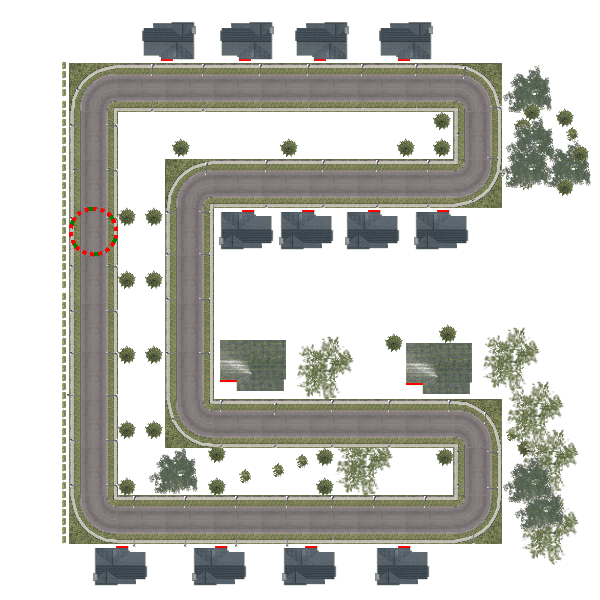} &
\includegraphics[width=0.22\textwidth]{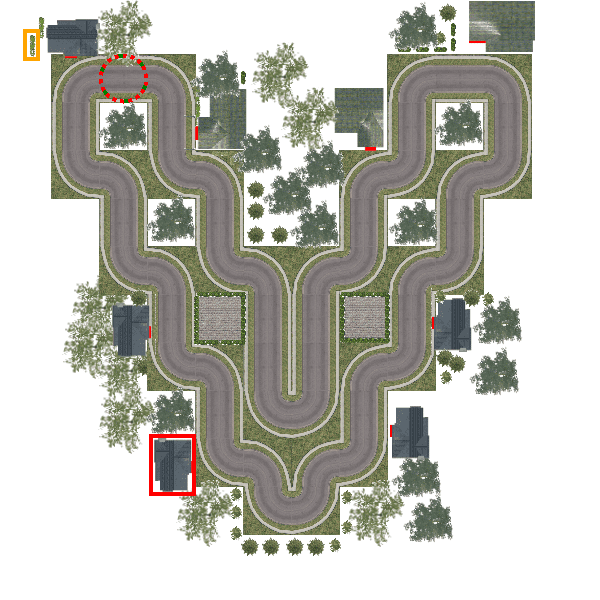} &
\includegraphics[width=0.22\textwidth]{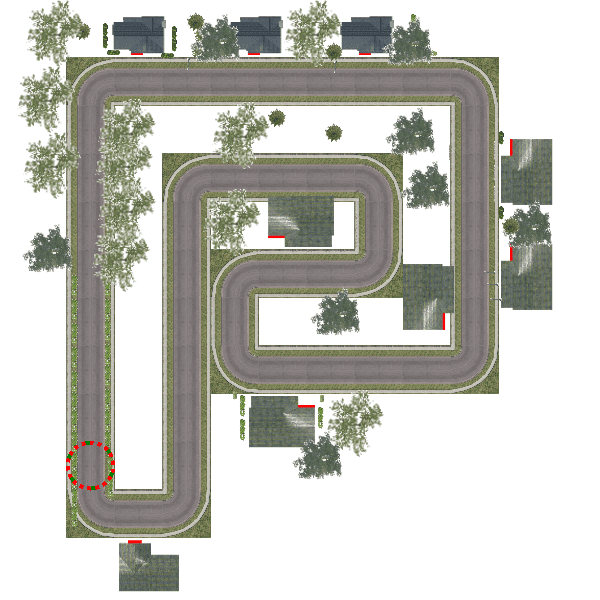} &
\includegraphics[width=0.22\textwidth]{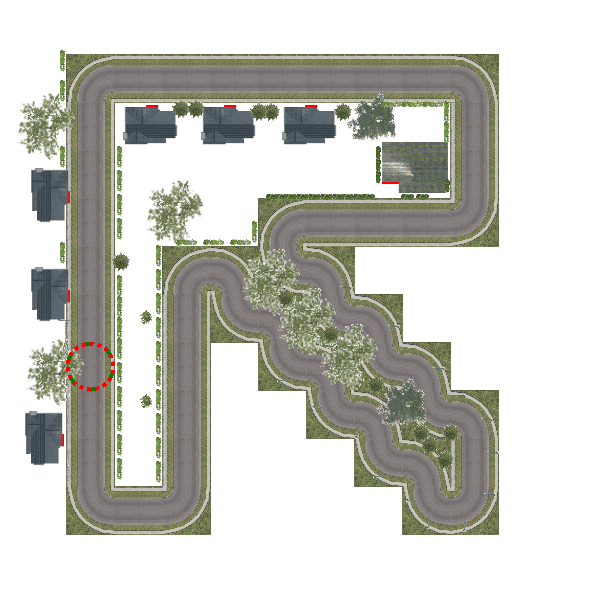}
\end{tabular}
\caption{Maps used to evaluate the driving policy.}
\label{fig:maps}
\end{figure*}

\subsection{Experimental Setup~} \label{sec: experimental_setup}
The Sim4CV \cite{sim4cv} (see Figure \ref{fig:teaser}) environment provides capabilities to generate labeled datasets for offline training (\eg imitation learning), and interact with the simulation environment for online learning (\eg OIL or reinforcement learning) as well as online evaluation. 
For each application, we design six environments for training and four for testing. 
For fair comparison, human drivers/pilots are given as much time as needed to practice on the training tracks, before attempting the test tracks.
Please refer to \figLabel \ref{fig:tracks} and \figLabel \ref{fig:maps} for the test environments and to the \supp for the training environments.

For the autonomous driving scenario, the PID teachers, learned baselines and human drivers have to complete one lap. They are scored based on the average error to the center line and the time needed to complete the course. In addition, they need to pass through invisible checkpoints placed every 50 meters to make sure they stay on the course and do not take any shortcuts. The vehicle is reset at the next checkpoint if it did not reach it within 15 seconds. 

For the UAV racing task, all pilots have to complete two laps on the test tracks and are scored based on the percentage of gates they maneuver through and the overall time. Similar to the autonomous driving scenario we reset the UAV at the next checkpoint if it was not reached within 10 seconds. Here, the checkpoints correspond to the racing gates. Visualizations of the vehicle trajectory for all models on each track are provided in the \supp.

\mysection{Training Details.~} \label{para:training}
For fair comparison we train each network until convergence or for at most $800$k steps. We estimate the value functions for all the teachers and the learner to select the most critical teacher with $N$ step rollouts ($N=300$ for the car, $N=200$ for the UAV) during the \emph{observing} phase. During the \emph{rehearsing} phase, we execute the learner policy for $N$ steps until the advantage function $A(s_1, \pi^{l}, \pi^{t_{*}}) > \epsilon$ or $I=50$ episodes are reached. We choose $\epsilon = -0.1V^{\pi^{t_{*}}}$ for the multi teacher experiments, and $\epsilon = 0$ for the single teacher experiments. We choose $J=60$ as the \emph{acting} step size.

\mysection{Reward Function of OIL.~}
The reward function of OIL is used to compute the value functions and advantage functions of learner and teachers. For OIL, we use reward functions that score the whole trajectory. Assuming a trajectory of length $N$, let $\tau = (s_1, a_1, s_2, a_2, ..., s_N)$ denote the generated trajectory and $S_{\text{end}}$ denote all the terminal states when the vehicles leave the course/track or hit an obstacle. 

For autonomous driving, preciseness in lane keeping is an important measurement that reflects the smoothness and safety of the algorithm. Hence, we use a reward function that considers both speed and trajectory error as follows:
\begin{equation}
R(\tau) = \begin{cases}
\frac{\zeta} {\alpha\sum_{n=1}^{N}e_{n}+1} &s_N \notin S_{\text{end}}\\
\frac{\zeta} {\alpha\sum_{n=1}^{N}e_{n}+1} + r_{\text{penalty}} &s_{N} \in S_{\text{end}}
\end{cases}
\end{equation}
where $\zeta$ is the trajectory length of the vehicle along the center of the lane and $e_n$ is the distance error to the center of the lane at state $s_{n}$. We choose $\alpha=0.5$ and a large negative reward $r_{\text{penalty}}=-15000$ for violations (\eg crossing the lane boundaries).

For UAV racing, it is not necessary to stay in the center of the track. Hence, we use the following reward function that considers forward speed $v_{f}$ and terminal state $S_{\text{end}}$:
\begin{equation}
R(\tau) = \begin{cases}
\sum_{n=1}^{N} v_{f_{n}} &s_N \notin S_{\text{end}}\\
\sum_{n=1}^{N} v_{f_{n}} + r_{\text{penalty}} &s_{N} \in S_{\text{end}}
\end{cases}
\end{equation}
where $v_{f_{n}}$ is the forward speed of the vehicle after executing $a_{n}$ at state $s_{n}$ and $r_{\text{penalty}}$ is a large negative reward. We choose the penalty to be $-15000$.

\mysection{Reward function of DDPG.~}
In our experiments, DDPG cannot converge using the same simple reward function as OIL. Therefore, we design a more complex reward function to induce a dense normalized reward as follows.
\begin{equation}
r_{s_{n}, a_{n}} = \begin{cases}
\frac{e_{n}-e_{n+1}}{e_{\text{norm}}} + \frac{1}{e_{n+1} + 1} + \beta \frac{v_{f_{n}}}{v_{\text{norm}}} &s_{n+1} \notin S_{\text{end}}\\
r_{\text{penalty}} &s_{n+1} \in S_{\text{end}}
\end{cases}
\end{equation}
where $e_{n}$ and $e_{n+1}$ are the distance errors to the center of the track/lane at state $s_{n}$ and state $s_{n+1}$ respectively, $e_{\text{norm}}$ is a distance error normalization factor, $v_{f_{n}}$ is the forward speed the vehicle, $v_{\text{norm}}$ is a velocity normalization factor and $r_{\text{penalty}}$ is a negative reward for when the vehicles reaches a terminal state. Furthermore, we find it is hard to learn both acceleration and steering controls for the car. Therefore, we fix the acceleration value to $E=0.5$ and choose $\beta=0$ to only score the agent by the distance error. 
For the UAV, we choose $\beta=1$ to score the agent by both the velocity and distance error and choose $v_{\text{norm}}=800$. We set $e_{\text{norm}}=15$ and $r_{\text{penalty}}=-0.2$ for both the UAV and the car.

\mysection{Controller Architecture of Learned Baselines.~}
For fair comparisons, we use the same control policy network with hidden units $\{64, 32, 16\}$ for all the learned baselines. As for the critic network of DDPG, we use the same number of hidden units and use an output layer with a one-dimensional output to predict the Q value.

\subsection{Results}

\mysection{Comparison to State-of-the-Art Baselines.~} \label{sec: baselines}
We compare OIL for both autonomous driving and UAV racing to several DNN baselines. These include both IL approaches (Behaviour Cloning, DAGGER) and RL approaches (DDPG). For each comparison we implement both learning from the single best teacher and an ensemble of teachers. This essentially allows a broad baseline comparison of 6 different state-of-the-art learning approaches evaluating various IL and RL approaches. 

In Table \ref{tbl:uav_car}, the learned approaches are compared to OIL. Our results demonstrate that OIL outperforms all learned baselines in both accuracy scores and timings. Behaviour Cloning and DAGGER both perform worse in terms of score and timings as the number of teachers increases. This is because they do not distinguish between good and bad behaviour and learn from both. Moreover, they are limited by their teachers and unable to outperform them. This can be seen in single teacher training, where they do not achieve scores better than their teacher. In contrast, OIL improves upon scores both with a single teacher and multiple teachers. In comparison to DDPG, OIL converges quickly without extensive hyperparameter search and still learns to fly/drive much more precisely and faster.

\begin{table}[!htb]
\centering
\resizebox{\columnwidth}{!}{
\begin{tabular}{lrr}
\toprule
\multicolumn{3}{c}{\textbf{Results for the UAV}}            \\ 
\midrule
\multicolumn{3}{c}{Teachers: PID controllers}            \\ 
\midrule
\multicolumn{1}{l|}{Teacher~\#1}          & 100\%  & 131.9 \\
\multicolumn{1}{l|}{Teacher~\#2}          & 76.4\% & 87.0  \\
\multicolumn{1}{l|}{Teacher~\#3}          & 97.2\% & 87.6  \\
\multicolumn{1}{l|}{Teacher~\#4}          & 80.6\% & 90.3  \\
\multicolumn{1}{l|}{Teacher~\#5}          & 69.4\% & 99.7  \\ 
\midrule
\multicolumn{3}{c}{Baseline: Human}              \\ 
\midrule
\multicolumn{1}{l|}{Novice}    & 97.2\% & 124.6           \\
\multicolumn{1}{l|}{Intermediate}      &   100\%     & 81.2       \\ 
\multicolumn{1}{l|}{Expert}            &   100\%     & 46.9       \\ 
\midrule
\multicolumn{3}{c}{Learned Policy: Best Teacher (1)}          \\ 
\midrule
\multicolumn{1}{l|}{Behaviour Cl.} &   94.4\%     &        139.6 \\
\multicolumn{1}{l|}{DAGGER}            &   100\%     &     134.1   \\
\multicolumn{1}{l|}{DDPG}              &   95.8\%     &    84.6    \\
\multicolumn{1}{l|}{OIL (single)}               &   100\%     &      133.9  \\ 
\midrule
\multicolumn{3}{c}{Learned Policy: All Teachers}         \\ 
\midrule
\multicolumn{1}{l|}{Behaviour Cl.} &   72.2\%     &   101.6     \\
\multicolumn{1}{l|}{DAGGER}            &  58.3\%      &   140.1     \\
\multicolumn{1}{l|}{DDPG}     &    95.8\%     &    84.6   \\
\multicolumn{1}{l|}{\textbf{OIL (multi)}} &    \textbf{100\%} & \textbf{81.3}       \\ \bottomrule
\end{tabular}%
~
\begin{tabular}{lrr}
\toprule
\multicolumn{3}{c}{\textbf{Results for the car}}            \\ 
\midrule
\multicolumn{3}{c}{Teachers: PID controllers}            \\ 
\midrule
\multicolumn{1}{l|}{Teacher~\#1}          &   24.1     & 151.4 \\
\multicolumn{1}{l|}{Teacher~\#2}          &    539.6    & 110.0   \\
\multicolumn{1}{l|}{Teacher~\#3}          &   19.5     & 84.5  \\
\multicolumn{1}{l|}{Teacher~\#4}          &    76.7    & 76.7  \\
\multicolumn{1}{l|}{Teacher~\#5}          &    568.4    & 102.2   \\ 
\midrule
\multicolumn{3}{c}{Baseline: Human}                      \\ 
\midrule
\multicolumn{1}{l|}{Novice}            &     85.3   & 100.7       \\
\multicolumn{1}{l|}{Intermediate}      &   80.6     & 88.3       \\
\multicolumn{1}{l|}{Expert}            &   49.0     & 70.4       \\ 
\midrule
\multicolumn{3}{c}{Learned Policy: Best Teacher (3)}          \\ 
\midrule
\multicolumn{1}{l|}{Behaviour Cl.} &    13.9    & 88.6        \\
\multicolumn{1}{l|}{DAGGER}            &     38.6   & 88.3       \\
\multicolumn{1}{l|}{DDPG}              &  57.4      &   139.6     \\
\multicolumn{1}{l|}{OIL (single)}   &    12.4    & 88.5       \\ 
\midrule
\multicolumn{3}{c}{Learned Policy: All Teachers}         \\ 
\midrule
\multicolumn{1}{l|}{Behaviour Cl.} &    388.6    & 112.7        \\
\multicolumn{1}{l|}{DAGGER}            &   25.9     & 87.8       \\
\multicolumn{1}{l|}{DDPG}              &    57.4      &   139.6   \\
\multicolumn{1}{l|}{\textbf{OIL (multi)}}   &  \textbf{17.6}   & \textbf{74.2}     \\ \bottomrule
\end{tabular}

}
\caption{\emph{Left:} Results for the UAV. Columns show the number of gates passed and time to complete two laps. \emph{Right:} Results for the car. Columns show average error to center of the road and time to complete one round. All results are averaged over all 4 test tracks/maps. Please refer to the \supp for detailed results per track/map.}
\label{tbl:uav_car}
\end{table}

\begin{table}[!htb]
\centering
\begin{tabular}{lll}
\toprule
\multicolumn{3}{c}{\textbf{Ablation study for the car}}                \\ 
\midrule
\multicolumn{1}{l|}{OIL (Teachers \#1-5, 60 steps)} & 15.3 & 80.5 \\
\multicolumn{1}{l|}{OIL (Teachers \#1-5, 180 steps)} & 16.0 & 80.7 \\
\multicolumn{1}{l|}{OIL (Teachers \#1-5, 300 steps)} &  17.6 & 74.2 \\
\multicolumn{1}{l|}{OIL (Teachers \#1-5, 600 steps)} & 13.8 & 82.2 \\
\multicolumn{1}{l|}{OIL (Teachers \#1,3,4, 300 steps)} & 25.8 & 73.6 \\
\bottomrule
\end{tabular}
\caption{Ablation study for the car. \emph{Left column:} average error to center of the road. \emph{Right column:} time to complete one round. All results are averaged over all 4 test tracks/maps. Please refer to the \supp for the detailed results per track/map.}
\vspace{-8pt}
\label{tbl:ablation}
\end{table}

\mysection{Comparison to Teachers and Human Performance.~} We compare our OIL trained control network to the teachers it learned from and human performance. The perception network is kept the same for all learned models. The summary of this comparison to OIL is given in Table \ref{tbl:uav_car}. Our results demonstrate that OIL outperforms all teachers and novice to intermediate human pilots/drivers. 

\emph{Teacher~\#1} has the least error of all teachers and is the only one to perfectly complete all gates in the UAV racing evaluation. However, OIL not only completes all gates but is also much faster when learning from multiple teachers. In the autonomous driving evaluation, it is more precise in centering along the middle of the road and also much faster when learning from multiple teachers. In comparison to humans, OIL is better than novice and intermediate levels but slower than an expert in both applications. A notable difference between the expert driver and OIL is that OIL has a much lower error in driving. It is able to maintain high speeds while staying most accurately in the center of the tracks. Compared to the expert driver, OIL is $3.83$ seconds slower but only has $35.91\%$ of the error in terms of the distance to the center.

\mysection{Ablation Study.~} We investigate the importance of trajectory length and the number of teachers, and report results in \tblLabel \ref{tbl:ablation}. Our experiments show that OIL is robust to different trajectory lengths and varying numbers of teachers. One observation is that different trajectory length balances the average error and speed. In our car experiment, when we set trajectory length to $300$ steps and use five teachers, OIL learns a fast and precise (compared to human experts and teachers) driving policy. Moreover, it is observed that increasing the number of teachers reduces the average error and makes the learned policy more stable. When OIL only learns from three teachers, the time to complete one round reduces slightly but the average error increases significantly compared to learning from five teachers.

\section{Conclusion}\label{sec: conclusion}
In this paper, we present Observational Imitation Learning (OIL), a new approach for training a stationary deterministic policy that is not bound by imperfect or inexperienced teachers since its policy is updated by selecting only the best maneuvers at different states. OIL can be regarded as a generalization of DAGGER for multiple imperfect teachers. 

We demonstrate the ability of OIL by training a control network to autonomously drive a car and to fly an unmanned aerial vehicle (UAV) through challenging race tracks. Extensive experiments demonstrate that OIL outperforms single and multiple-teacher learned IL methods (Behavior Cloning, DAGGER), RL approaches (DDPG), its teachers and experienced humans pilots/drivers. 

While our method works well in the navigation tasks with access to non-expert controllers, a question arises concerning how this method would scale to more complex tasks where controller design is very time-consuming or intractable, \eg Atari games. OIL is general in nature and only requires access to teachers and a function to score them; one approach to avoid hand-designed controllers would be to use learned ones instead (\eg trained using different RL algorithms) and use the reward of the environment as a score for OIL.

\section{Future Work}
OIL provides a new learning-based approach that can replace traditional control methods especially in robotics and control systems. We expect OIL to expand its reach to other areas of autonomous navigation (\eg obstacle avoidance) and benefit other robotic tasks (\eg visual grasping or placing). 

One very interesting avenue for future work is to apply OIL outside simulation. Although Sim4CV uses a high fidelity game engine for rendering and physics, the differences between the simulated world and the real world will need to be reconciled. Real-world physics, weather and road conditions, as well as sensor noise will present new challenges for adapting/training the control network.

\section*{Acknowledgments}
This work was supported by the King Abdullah University of Science and Technology (KAUST) Office of Sponsored Research through the Visual Computing Center (VCC) funding.


\bibliographystyle{plainnat}
\bibliography{references}

\appendix
\label{sec: supp}

\begin{figure}[!b]
	   \includegraphics[width=\linewidth]{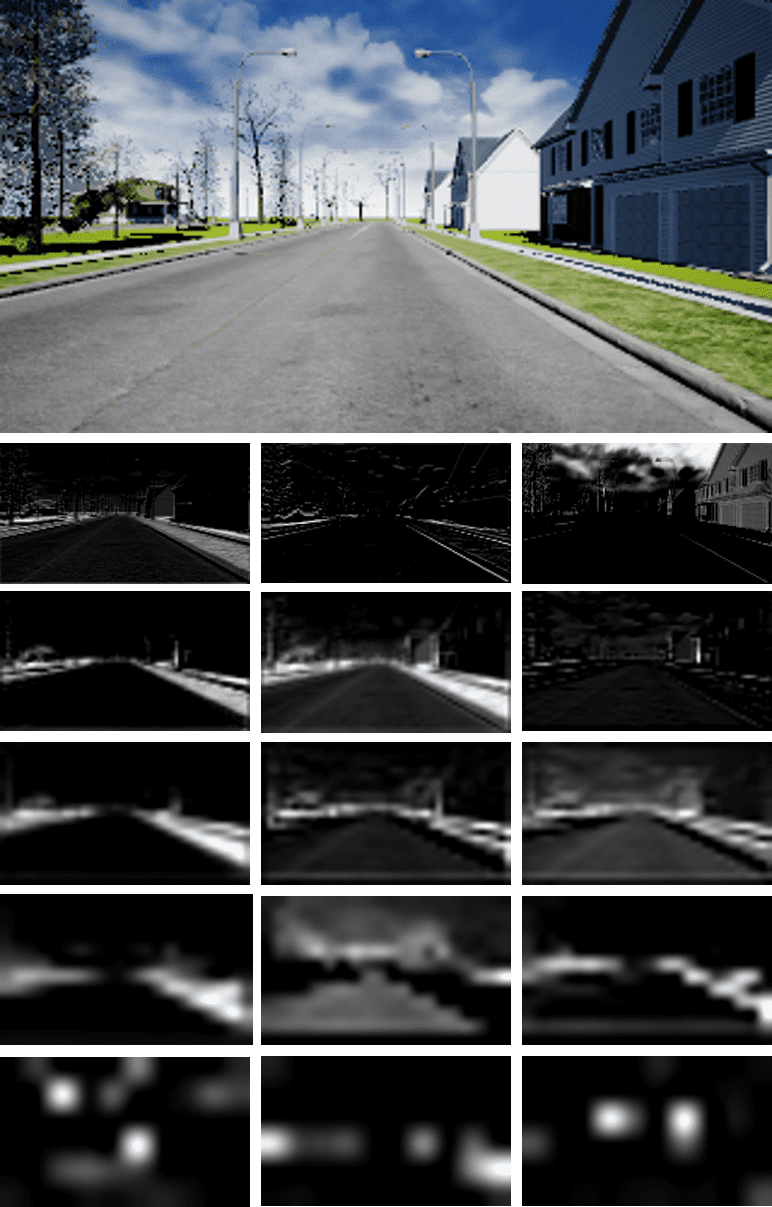}
     \caption{Visualization of selected feature maps from convolutional layers 1 to 5 (top to bottom) in our trained perception network. Top image is input at 180$\times$320 resolution. Notice how the network activates in locations of semantic meaning for the task of autonomous driving, namely the road boundaries.}
     \label{fig:featuremaps_car}
\end{figure}

\begin{figure}[!b]
	   \includegraphics[width=\linewidth]{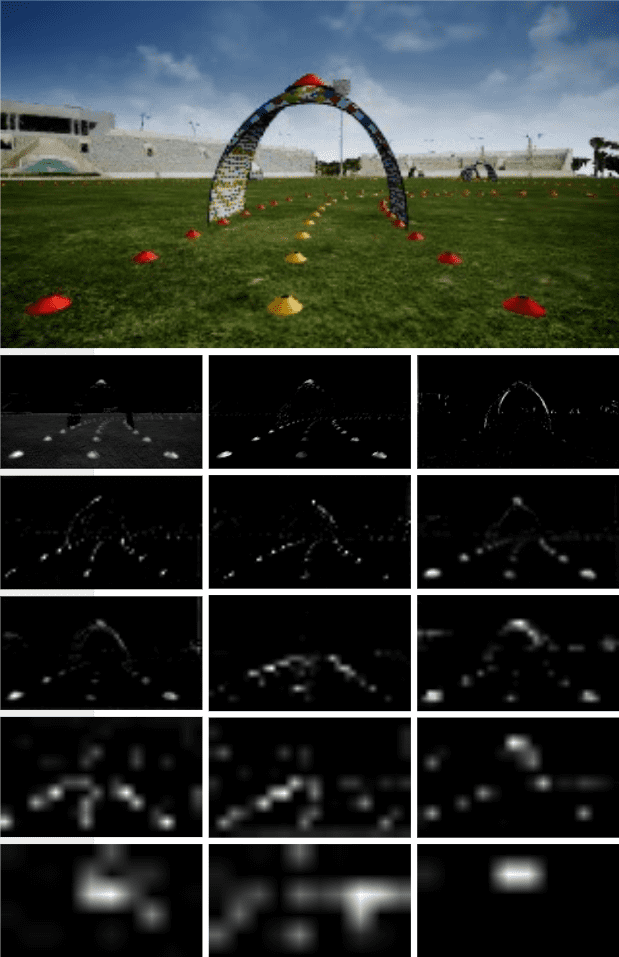}
     \caption{Visualization of selected feature maps from convolutional layers 1 to 5 (top to bottom) in our trained perception network. Top image is input at 180$\times$320 resolution. Notice how the network activates in locations of semantic meaning for the task of UAV racing, namely the gates and cones.}
     \label{fig:featuremaps_uav}
\end{figure}

\mysection{Perception Network Visualization.} After training the perception network to convergence, we visualize how parts of the network behave in order to get additional insights on what it has intrinsically learned. Figures \ref{fig:featuremaps_car} and \ref{fig:featuremaps_uav} show some feature maps for each convolutional layer of the trained networks for an input image. Note how the learned filters are able to extract all necessary information in the scene (\ie road boundaries or gates and cones). While filters from early layers already produce a reasonable activation heatmap, the deeper layers abstract the relevant features even more clearly. Although the feature map resolution becomes very low in the deeper layers, high activations can still be identified for parts of the image with semantic meaning for autonomous driving and UAV racing.

\begin{figure*}
\centering
\begin{tabular}{@{}c@{\hspace{1mm}}c@{\hspace{1mm}}c@{}}
		\includegraphics[height=4cm]{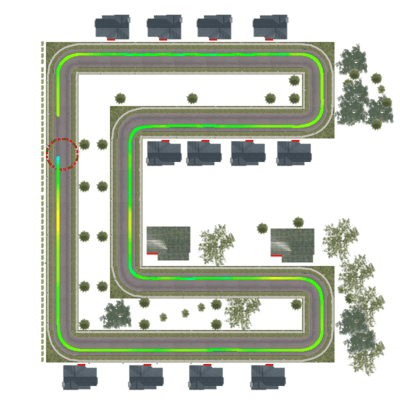} &
		\includegraphics[height=4cm]{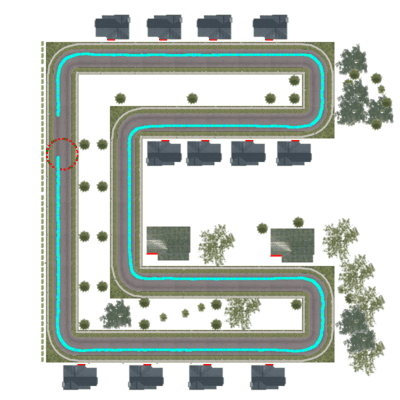} &
		\includegraphics[height=4cm]{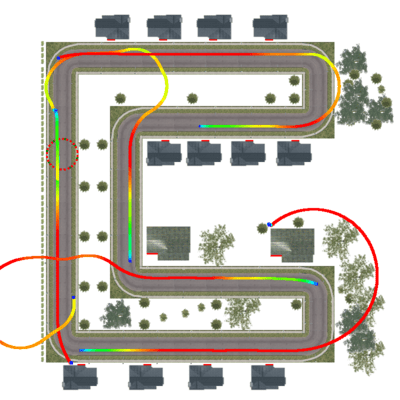} \\
		\small (a) OIL &
		\small (b) Teacher 1 &
		\small (c) Teacher 2 \\
		\includegraphics[height=4cm]{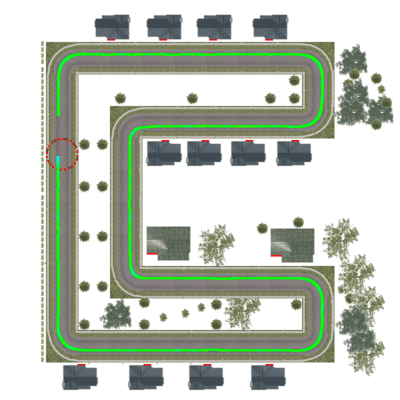} &
		\includegraphics[height=4cm]{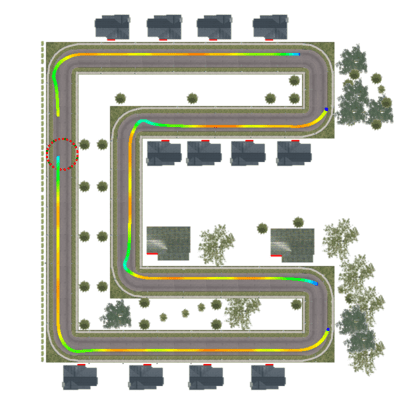} &
		\includegraphics[height=4cm]{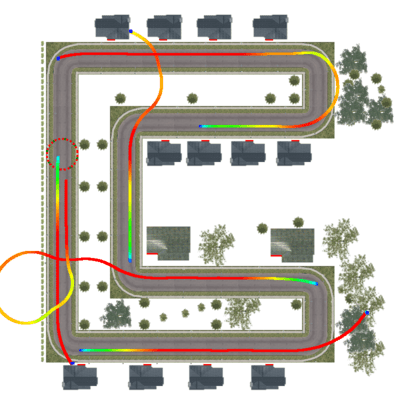} \\
		\small (d) Teacher 3 &
		\small (e) Teacher 4 &
		\small (f) Teacher 5 \\
		\includegraphics[height=4cm]{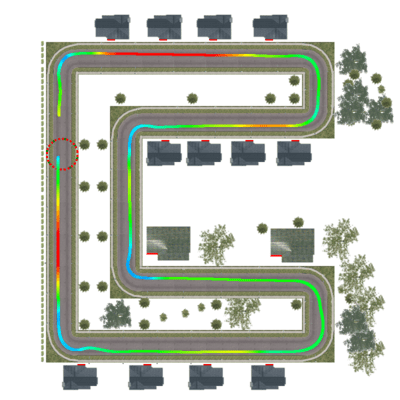} &
		\includegraphics[height=4cm]{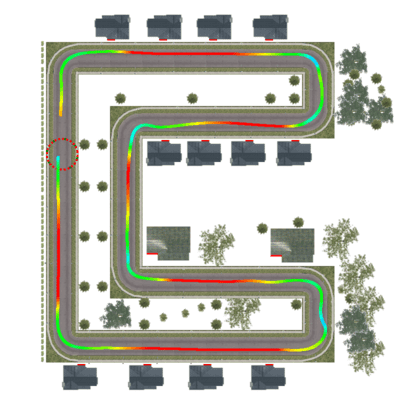} &
		\includegraphics[height=4cm]{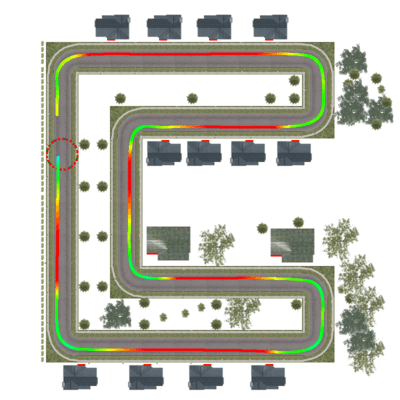}\\
		\small (g) Novice &
		\small (h) Intermediate &
		\small (i) Professional \\
		\includegraphics[height=4cm]{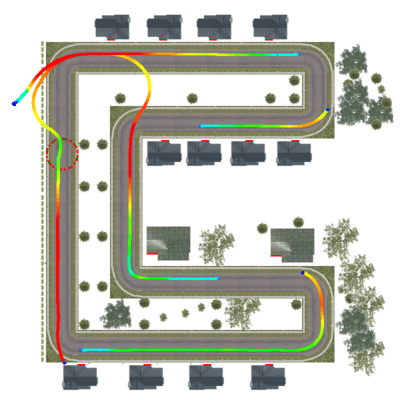} &
		\includegraphics[height=4cm]{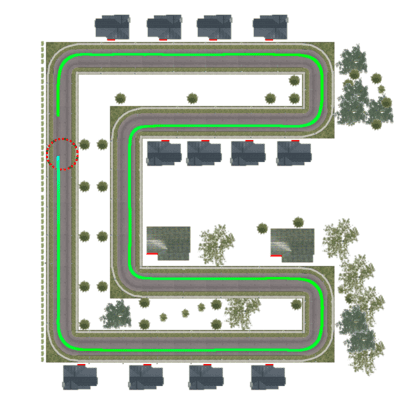} &
		\includegraphics[height=4cm]{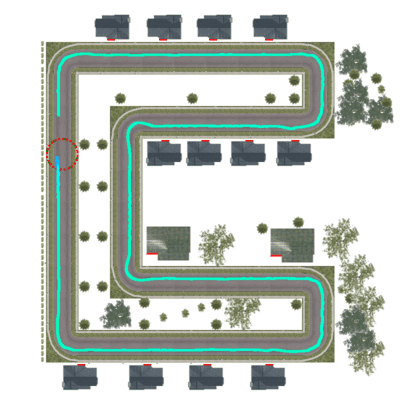}\\
		\small (j) Behaviour Cloning &
		\small (k) Dagger &
		\small (l) DDPG \\
       \multicolumn{3}{c}{\includegraphics[height=1.2cm]{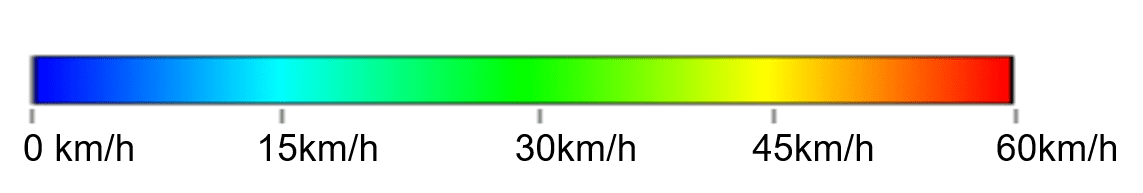}} 
\end{tabular}
\captionof{figure}{Comparison between our learned policy and its teachers (\emph{row 1,2}), human drivers (\emph{row 3}) and baselines (\emph{row 4}) on test course 1. Color encodes speed as a heatmap, where blue is the minimum speed and red is the maximum speed.}
\label{fig:qualitive_results_map1}
\end{figure*}

\begin{figure*}
\centering
\begin{tabular}{@{}c@{\hspace{1mm}}c@{\hspace{1mm}}c@{}}
		\includegraphics[height=4cm]{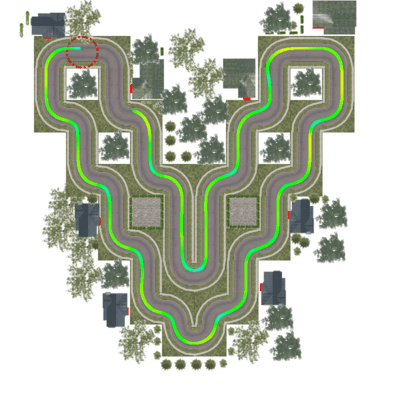} &
		\includegraphics[height=4cm]{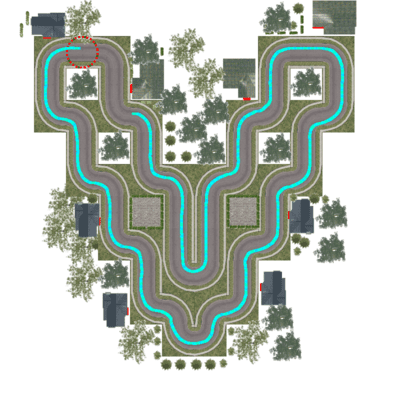} &
		\includegraphics[height=4cm]{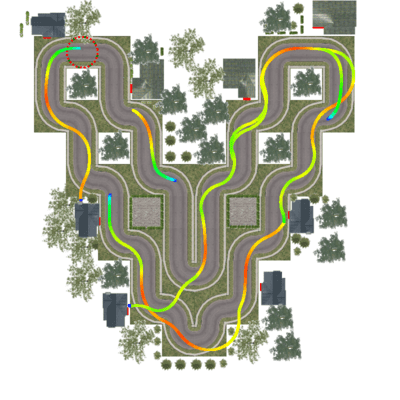} \\
		\small (a) OIL &
		\small (b) Teacher 1 &
		\small (c) Teacher 2 \\
		\includegraphics[height=4cm]{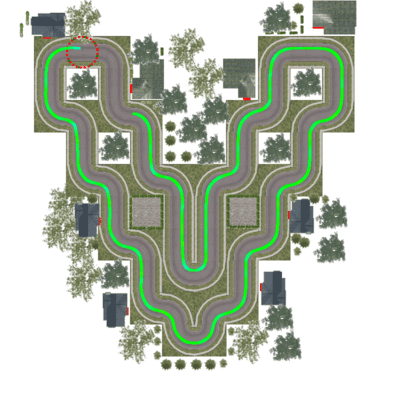} &
		\includegraphics[height=4cm]{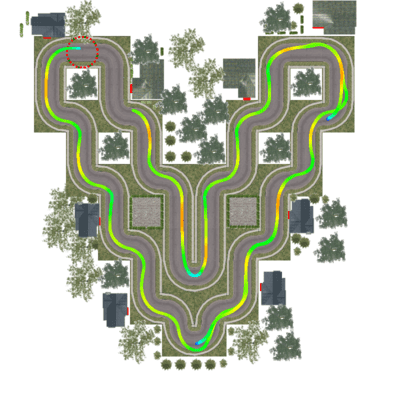} &
		\includegraphics[height=4cm]{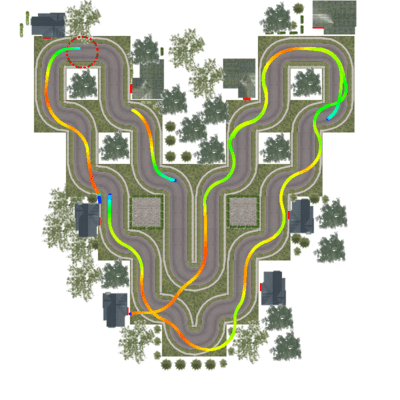} \\
		\small (d) Teacher 3 &
		\small (e) Teacher 4 &
		\small (f) Teacher 5 \\
		\includegraphics[height=4cm]{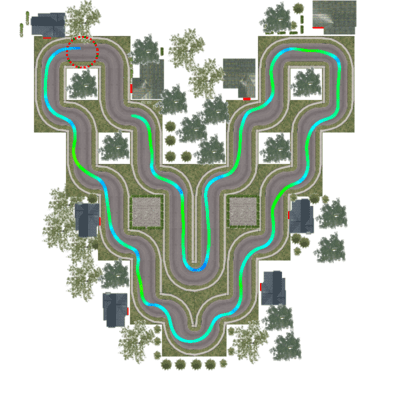} &
		\includegraphics[height=4cm]{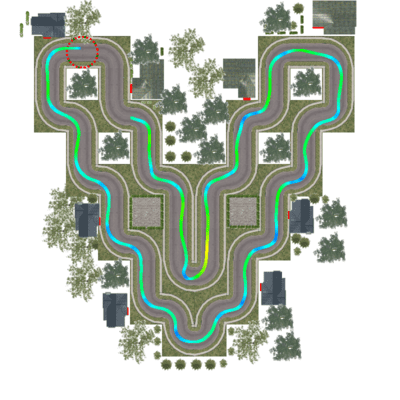} &
		\includegraphics[height=4cm]{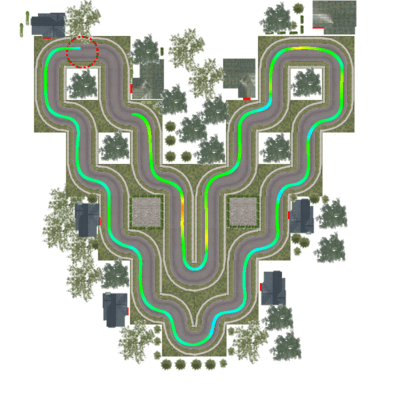}\\
		\small (g) Novice &
		\small (h) Intermediate &
		\small (i) Professional \\
		\includegraphics[height=4cm]{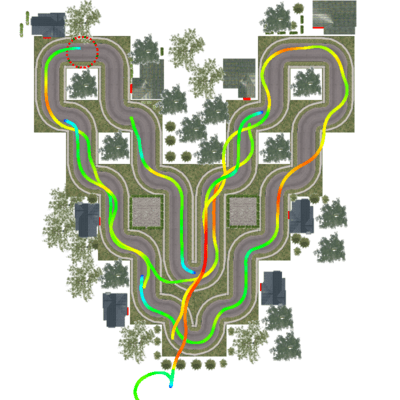} &
		\includegraphics[height=4cm]{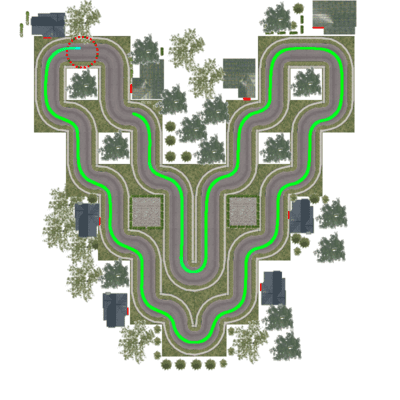} &
		\includegraphics[height=4cm]{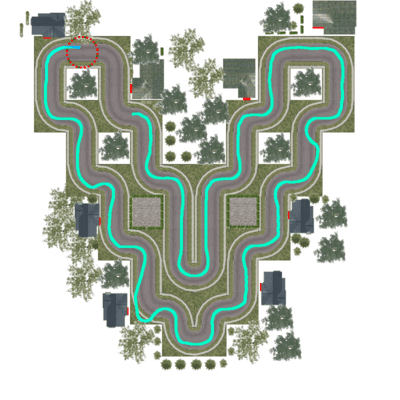}\\
		\small (j) Behaviour Cloning &
		\small (k) Dagger &
		\small (l) DDPG \\
       \multicolumn{3}{c}{\includegraphics[height=1.2cm]{figures/ColorScaleCar.png}} 
\end{tabular}
\captionof{figure}{Comparison between our learned policy and its teachers (\emph{row 1,2}), human drivers (\emph{row 3}) and baselines (\emph{row 4}) on test course 2. Color encodes speed as a heatmap, where blue is the minimum speed and red is the maximum speed.}
\label{fig:qualitive_results_map2}
\end{figure*}

\begin{figure*}
\centering
\begin{tabular}{@{}c@{\hspace{1mm}}c@{\hspace{1mm}}c@{}}
		\includegraphics[height=4cm]{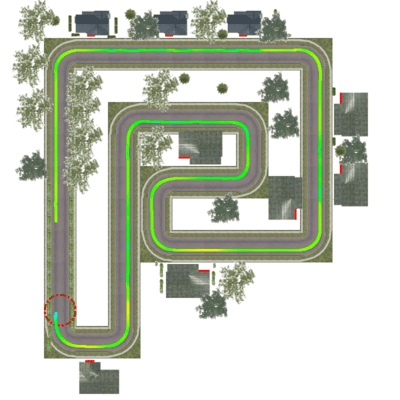} &
		\includegraphics[height=4cm]{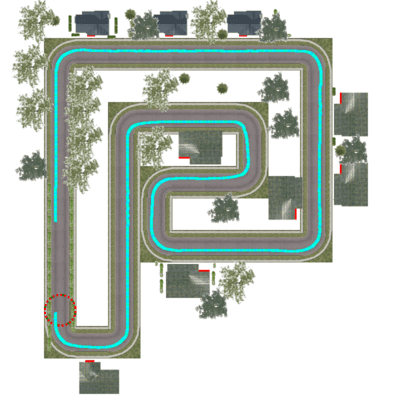} &
		\includegraphics[height=4cm]{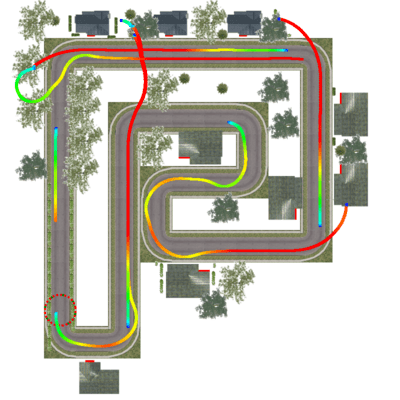} \\
		\small (a) OIL &
		\small (b) Teacher 1 &
		\small (c) Teacher 2 \\
		\includegraphics[height=4cm]{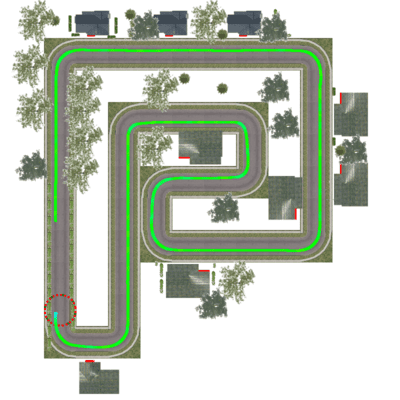} &
		\includegraphics[height=4cm]{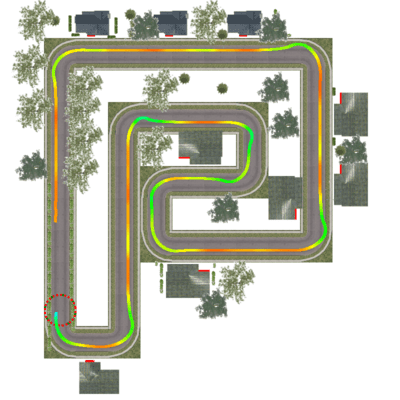} &
		\includegraphics[height=4cm]{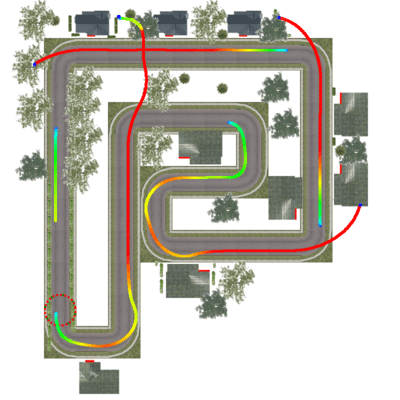} \\
		\small (d) Teacher 3 &
		\small (e) Teacher 4 &
		\small (f) Teacher 5 \\
		\includegraphics[height=4cm]{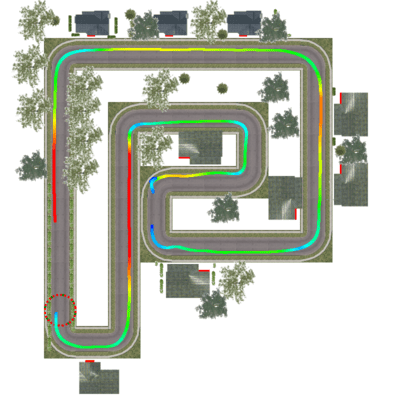} &
		\includegraphics[height=4cm]{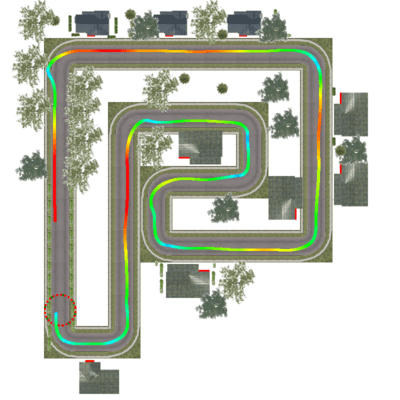} &
		\includegraphics[height=4cm]{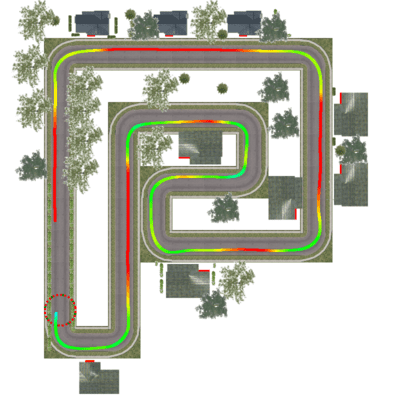}\\
		\small (g) Novice &
		\small (h) Intermediate &
		\small (i) Professional \\
		\includegraphics[height=4cm]{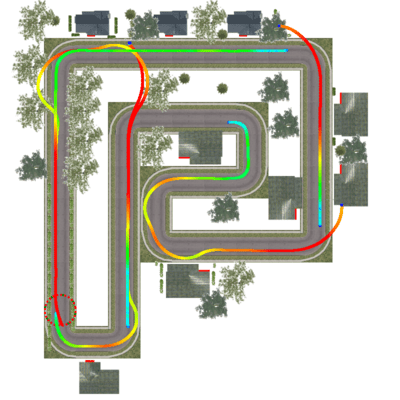} &
		\includegraphics[height=4cm]{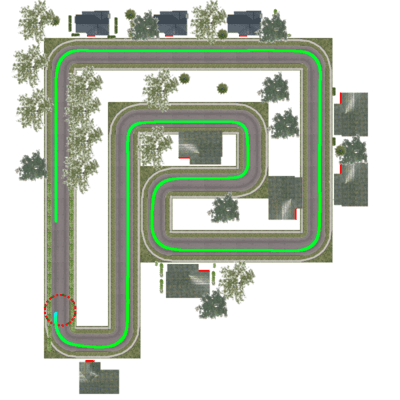} &
		\includegraphics[height=4cm]{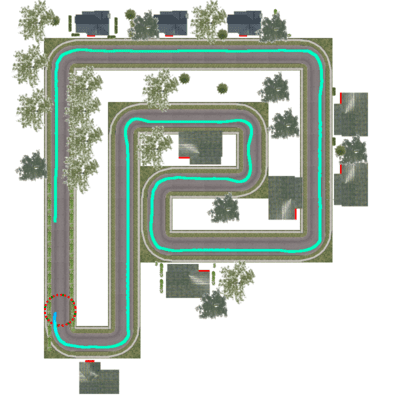}\\
		\small (j) Behaviour Cloning &
		\small (k) Dagger &
		\small (l) DDPG \\
       \multicolumn{3}{c}{\includegraphics[height=1.2cm]{figures/ColorScaleCar.png}} 
\end{tabular}
\captionof{figure}{Comparison between our learned policy and its teachers (\emph{row 1,2}), human drivers (\emph{row 3}) and baselines (\emph{row 4}) on test course 3. Color encodes speed as a heatmap, where blue is the minimum speed and red is the maximum speed.}
\label{fig:qualitive_results_map3}
\end{figure*}

\begin{figure*}
\centering
\begin{tabular}{@{}c@{\hspace{1mm}}c@{\hspace{1mm}}c@{}}
		\includegraphics[height=4cm]{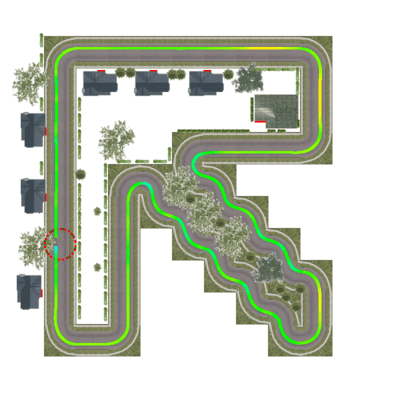} &
		\includegraphics[height=4cm]{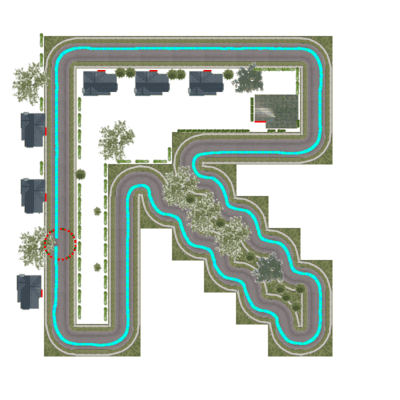} &
		\includegraphics[height=4cm]{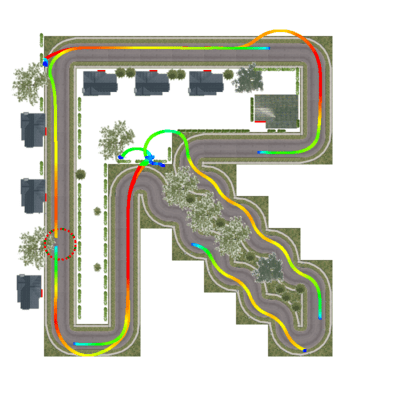} \\
		\small (a) OIL &
		\small (b) Teacher 1 &
		\small (c) Teacher 2 \\
		\includegraphics[height=4cm]{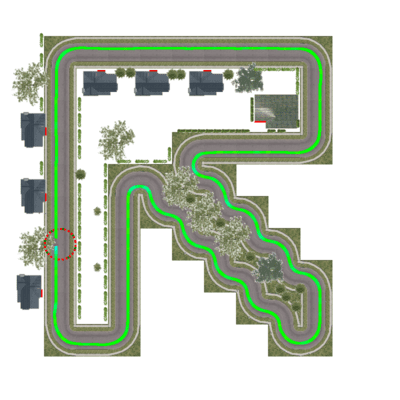} &
		\includegraphics[height=4cm]{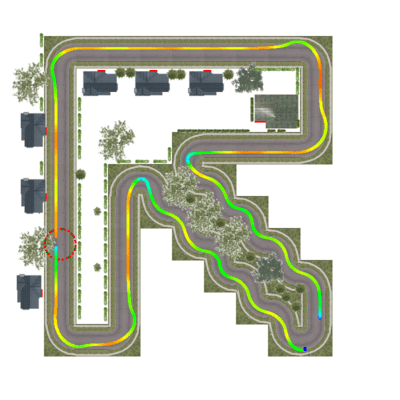} &
		\includegraphics[height=4cm]{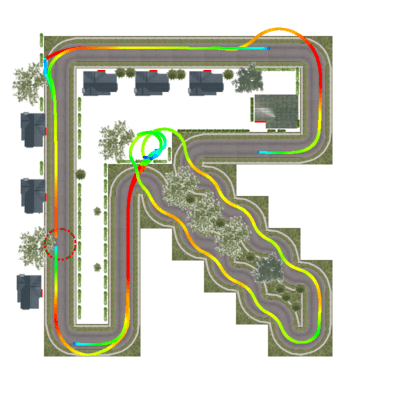} \\
		\small (d) Teacher 3 &
		\small (e) Teacher 4 &
		\small (f) Teacher 5 \\
		\includegraphics[height=4cm]{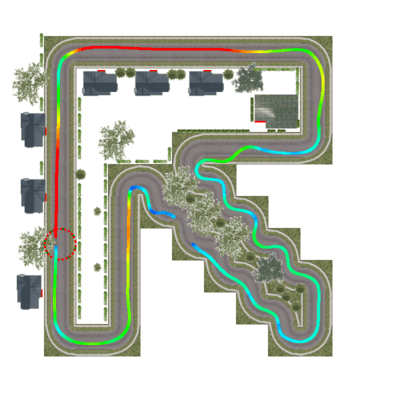} &
		\includegraphics[height=4cm]{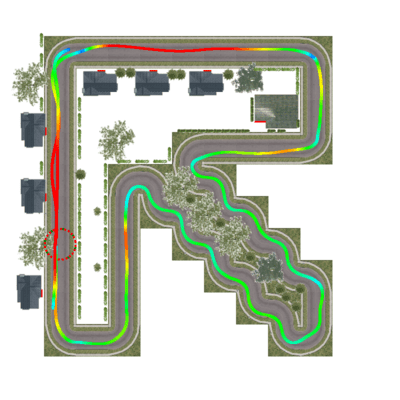} &
		\includegraphics[height=4cm]{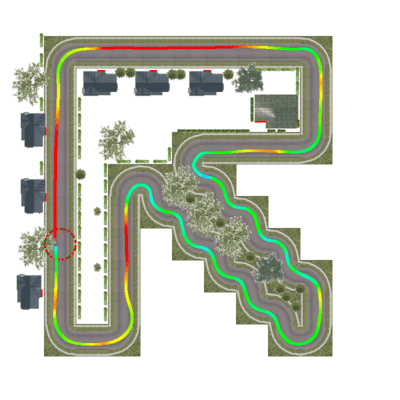}\\
		\small (g) Novice &
		\small (h) Intermediate &
		\small (i) Professional \\
		\includegraphics[height=4cm]{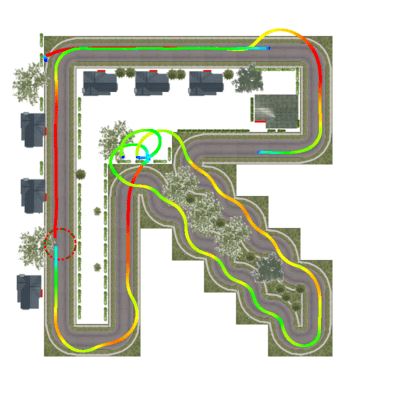} &
		\includegraphics[height=4cm]{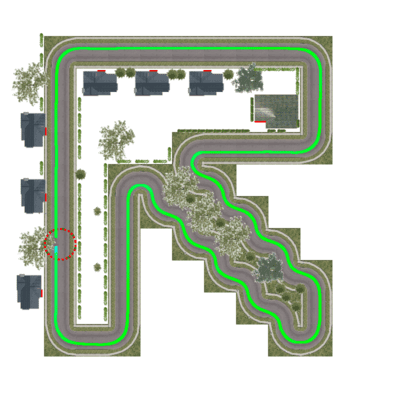} &
		\includegraphics[height=4cm]{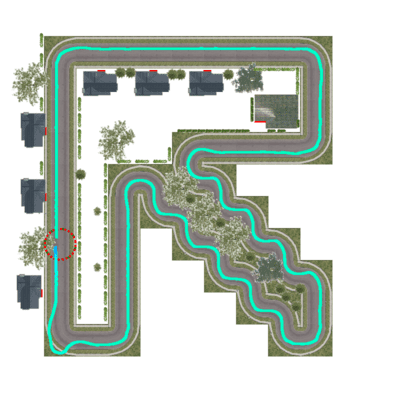}\\
		\small (j) Behaviour Cloning &
		\small (k) Dagger &
		\small (l) DDPG \\
       \multicolumn{3}{c}{\includegraphics[height=1.2cm]{figures/ColorScaleCar.png}} 
\end{tabular}
\captionof{figure}{Comparison between our learned policy and its teachers (\emph{row 1,2}), human drivers (\emph{row 3}) and baselines (\emph{row 4}) on test course 4. Color encodes speed as a heatmap, where blue is the minimum speed and red is the maximum speed.}
\label{fig:qualitive_results_map4}
\end{figure*}

\mysection{Detailed Experimental Results.}
We show detailed results for the autonomous driving experiments for each course and report the average error to the center of the lane and the time to complete the course in Tables \ref{tbl: regular baselines car}, \ref{tbl: learned baselines car} and \ref{tbl: ablation study car}. In \tblLabel \ref{tbl: regular baselines car}, we show the results for each of the teachers and provide human baselines. In \tblLabel \ref{tbl: learned baselines car}, we compare our approach \emph{OIL} to several learned baselines. In \tblLabel \ref{tbl: ablation study car}, we show an ablation study of \emph{OIL} in regard to trajectory length and number of teachers. Figures \ref{fig:qualitive_results_map1}, \ref{fig:qualitive_results_map2}, \ref{fig:qualitive_results_map3} and \ref{fig:qualitive_results_map4} visualize the results.

\begin{figure*}
\centering
\begin{tabular}{@{}c@{\hspace{1mm}}c@{\hspace{1mm}}c@{}}
		\includegraphics[height=3cm]{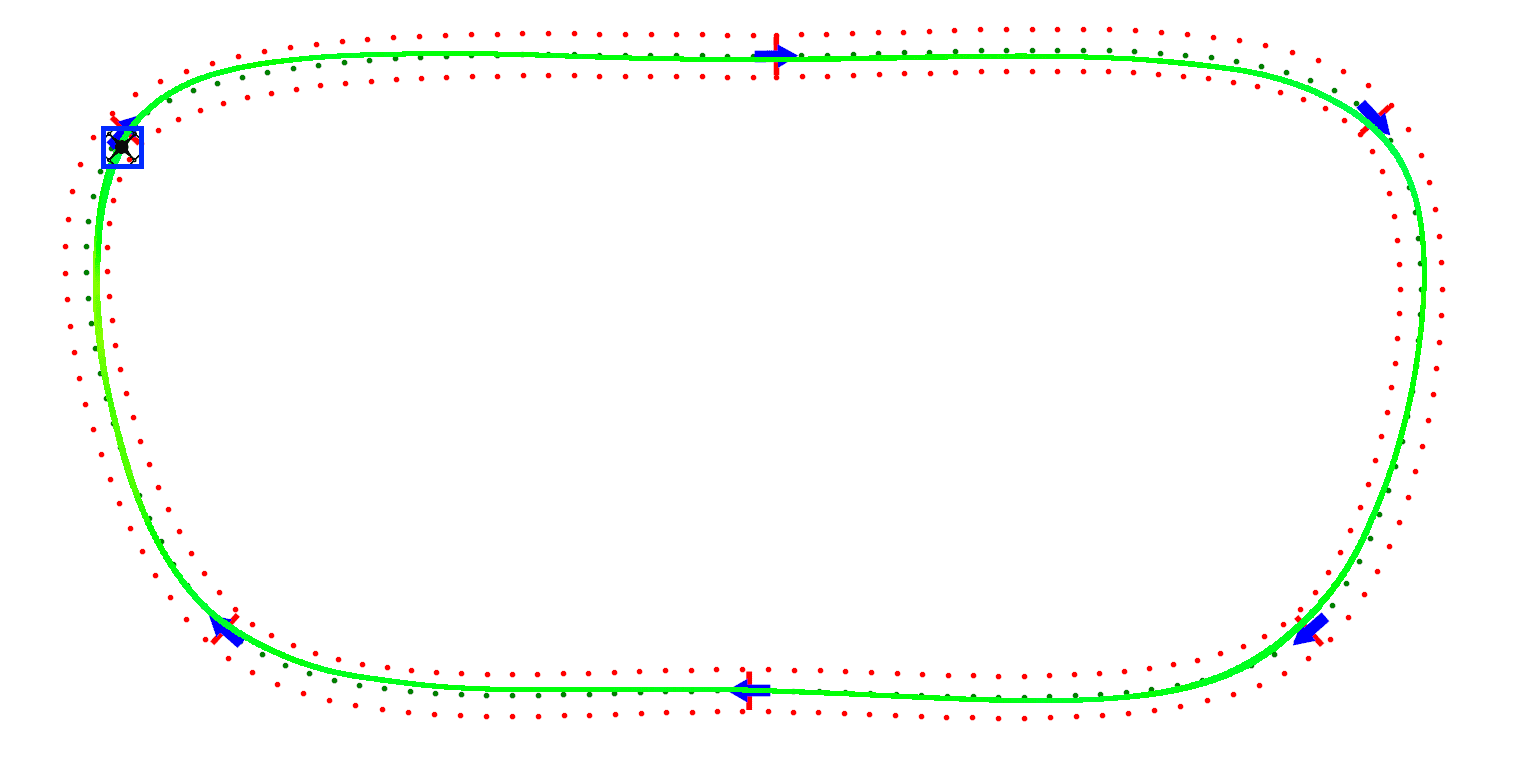} &
		\includegraphics[height=3cm]{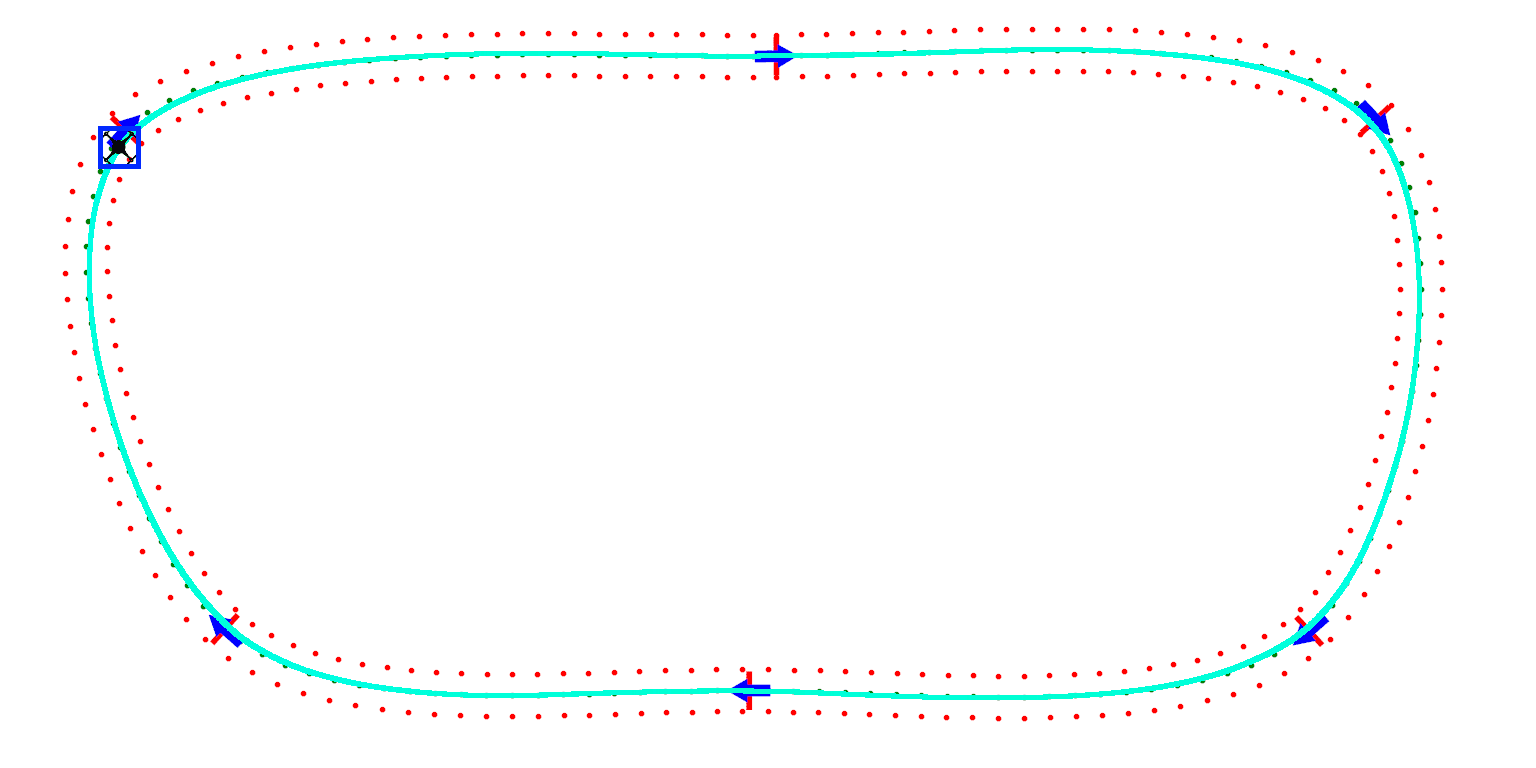} &
		\includegraphics[height=3cm]{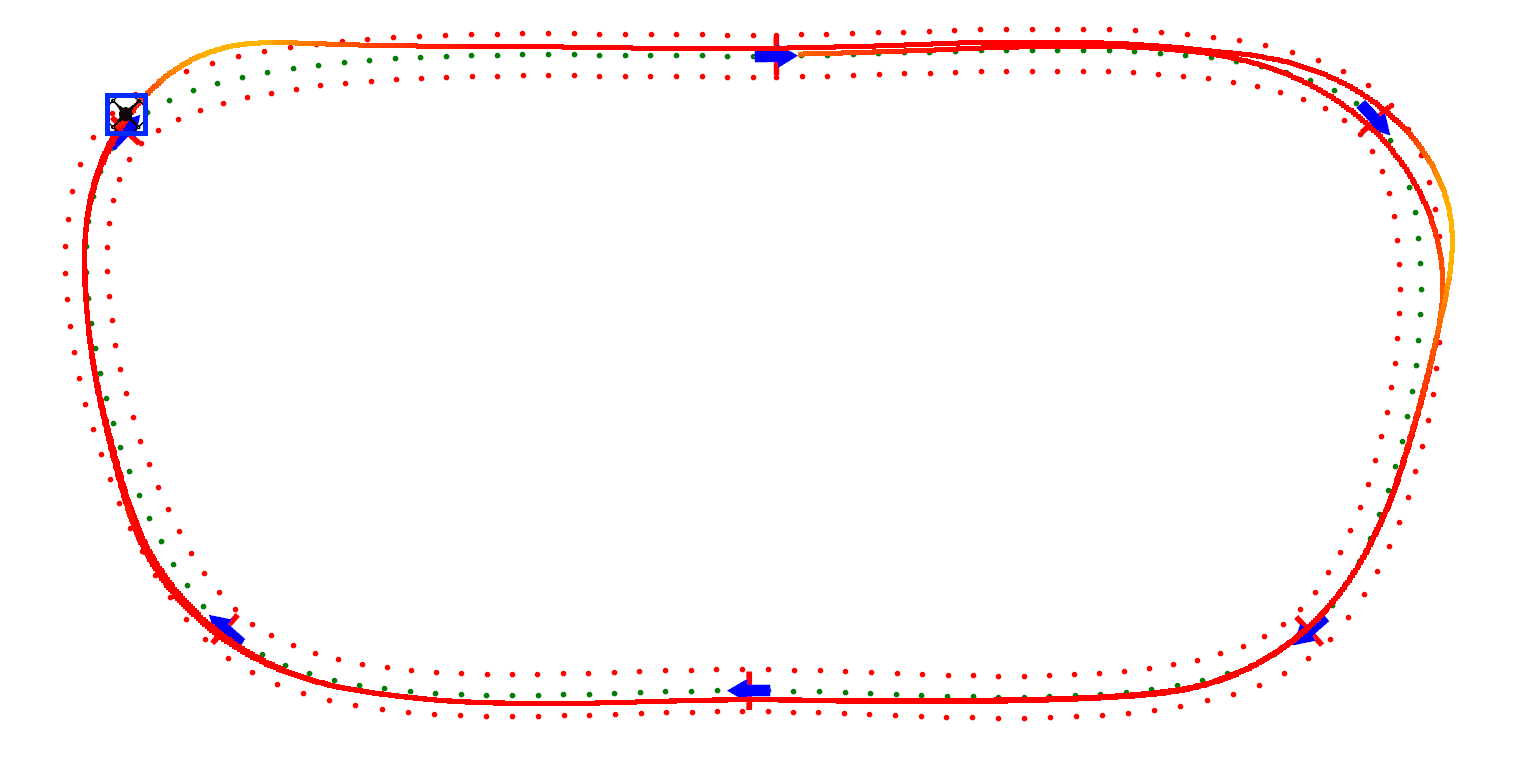} \\
		\small (a) OIL &
		\small (b) Teacher 1 &
		\small (c) Teacher 2 \\
		\includegraphics[height=3cm]{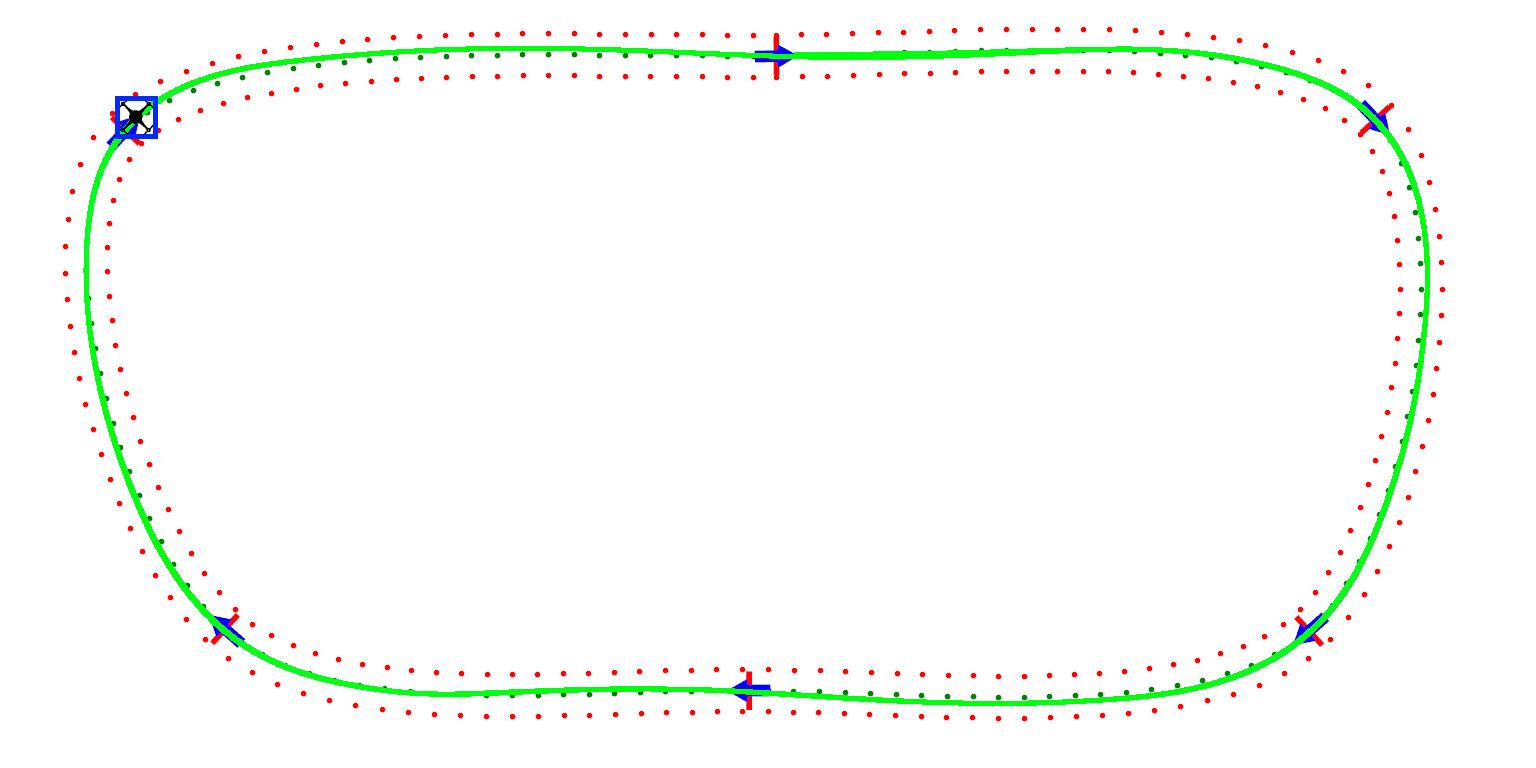} &
		\includegraphics[height=3cm]{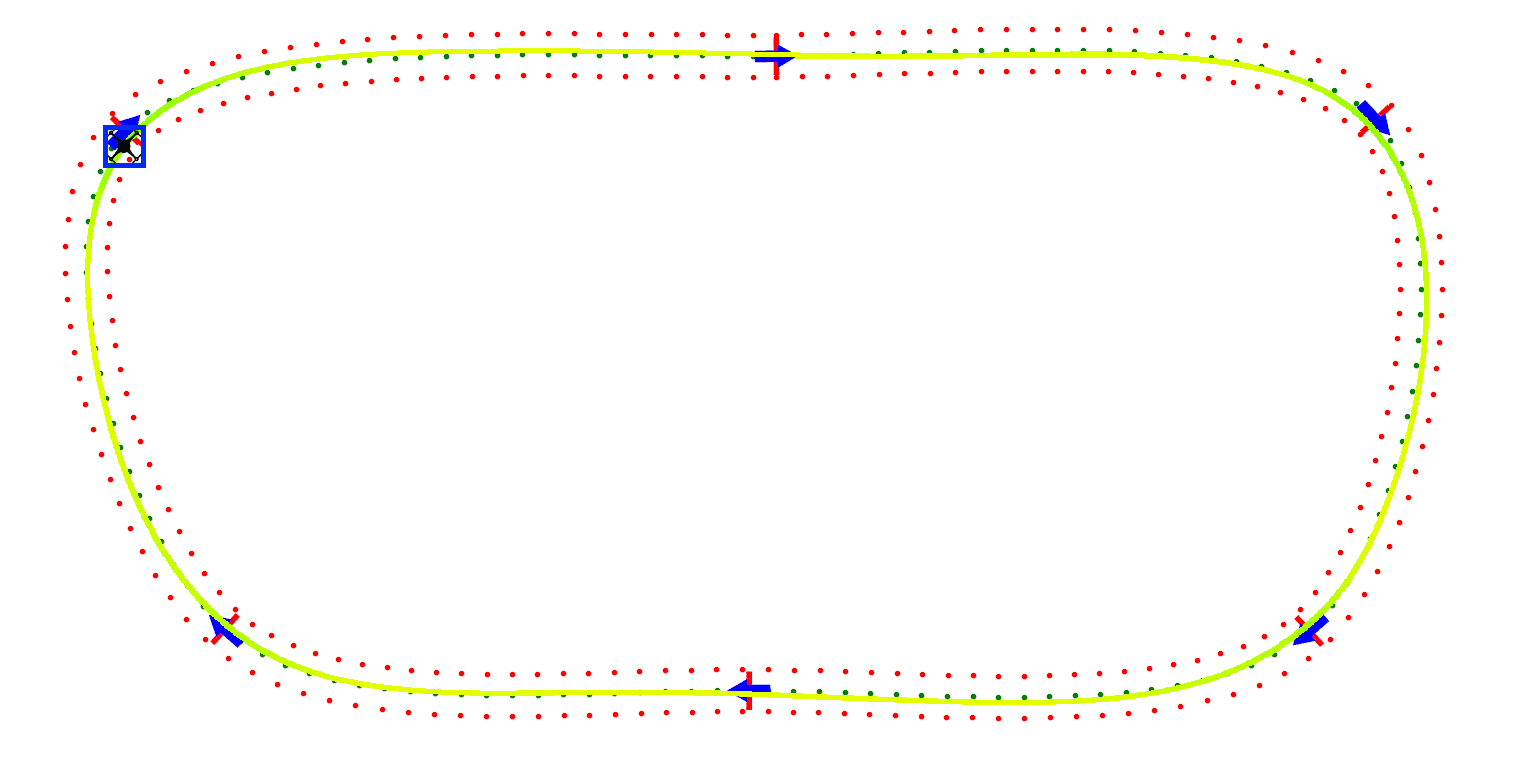} &
		\includegraphics[height=3cm]{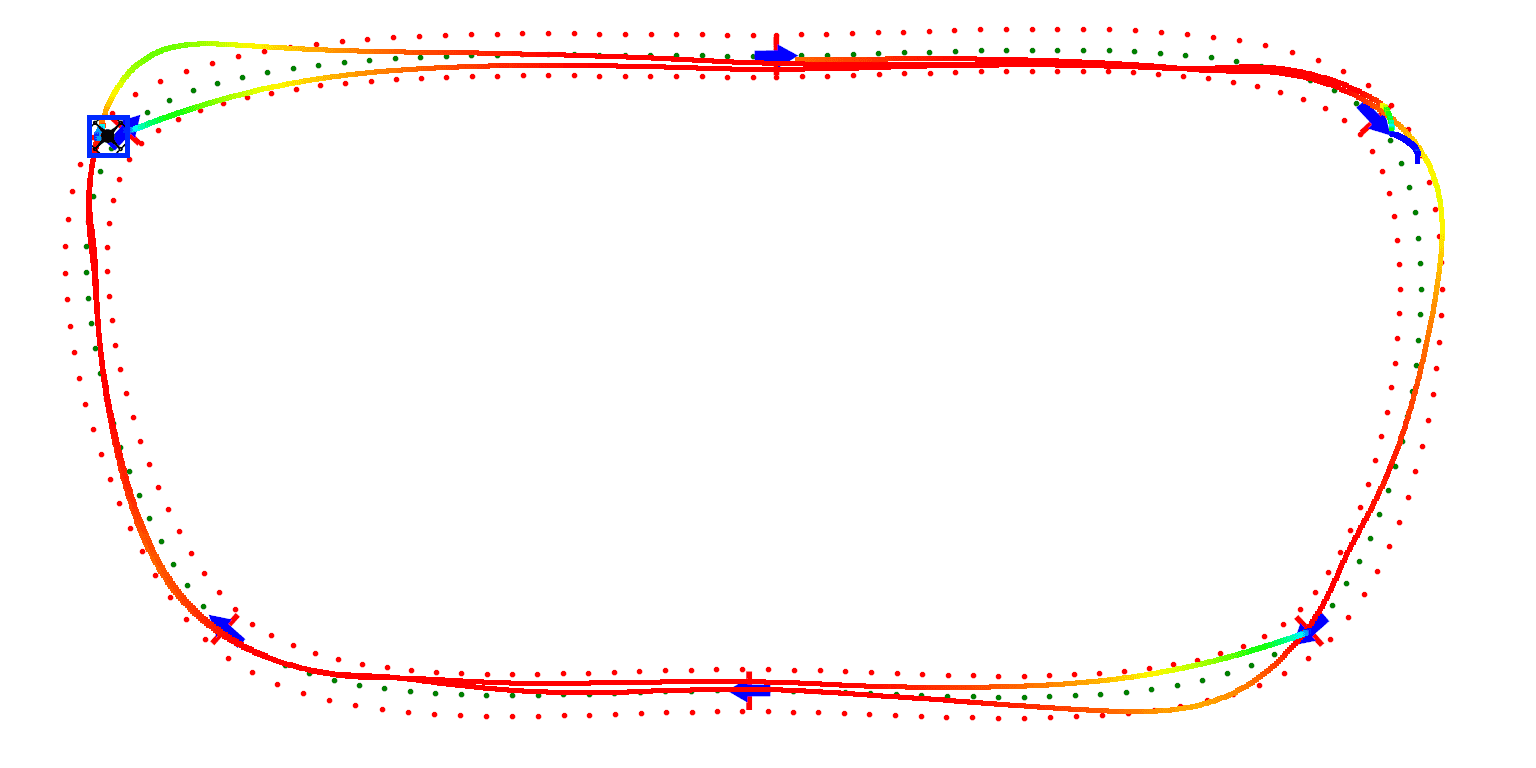} \\
		\small (d) Teacher 3 &
		\small (e) Teacher 4 &
		\small (f) Teacher 5 \\
		\includegraphics[height=3cm]{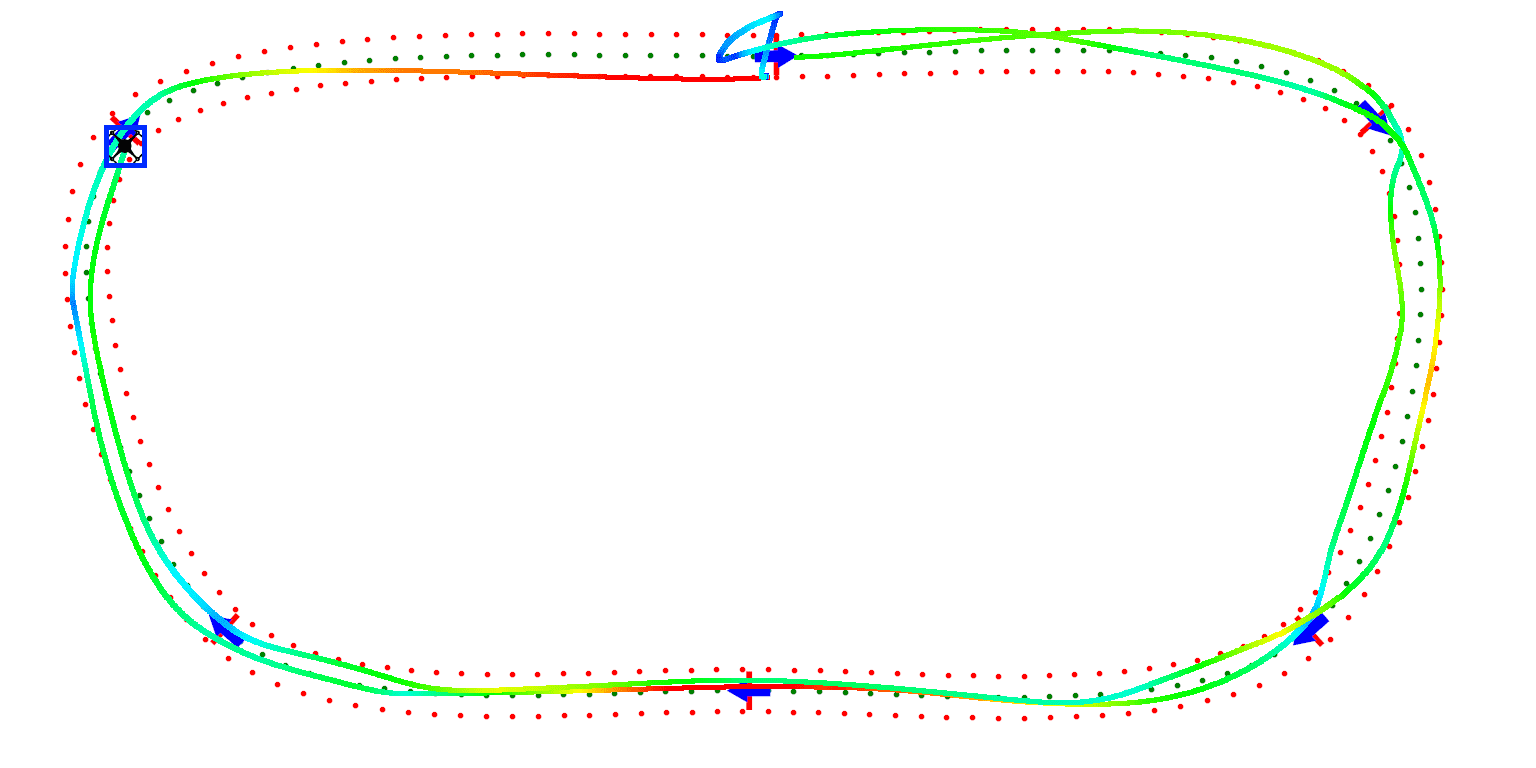} &
		\includegraphics[height=3cm]{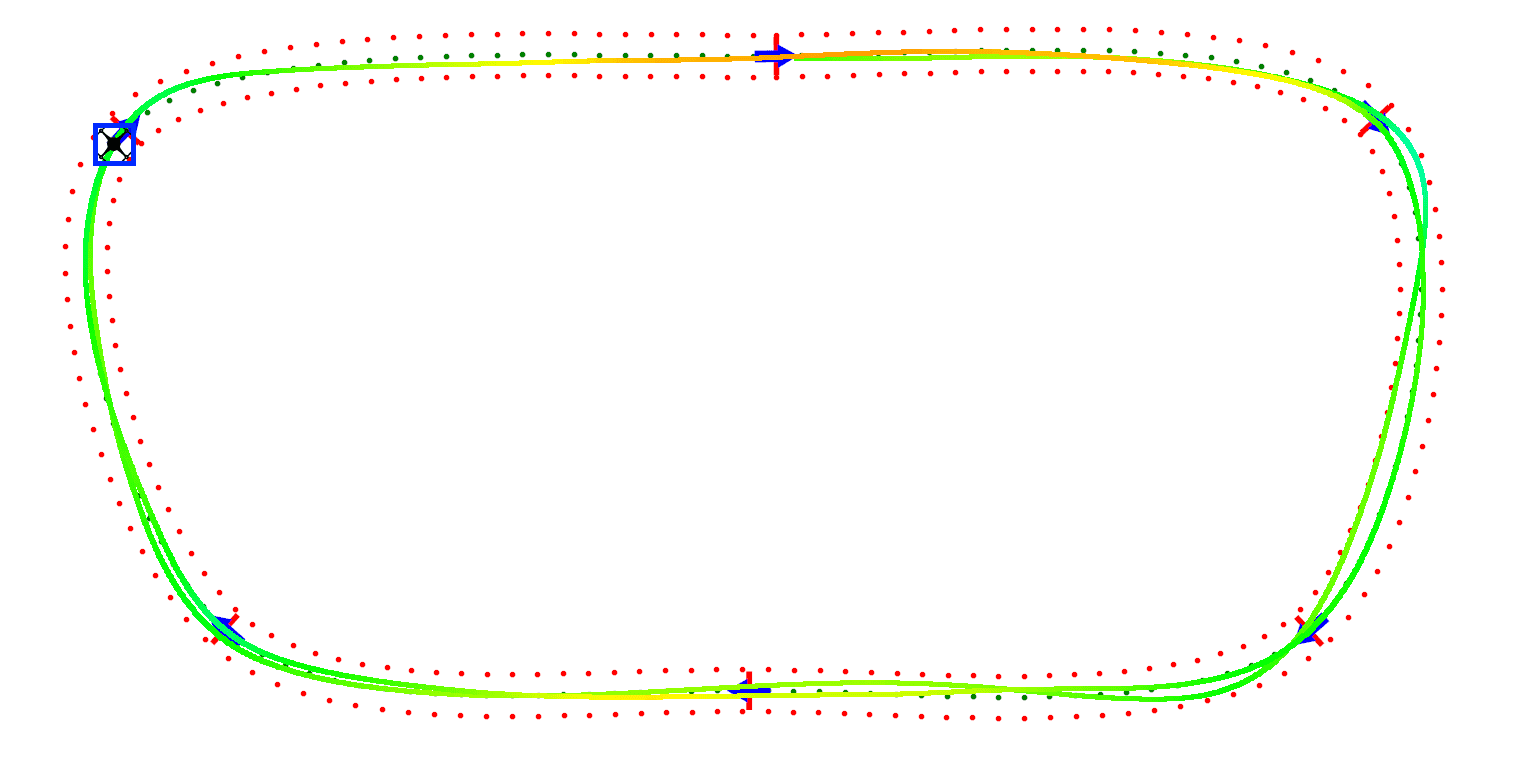} &
		\includegraphics[height=3cm]{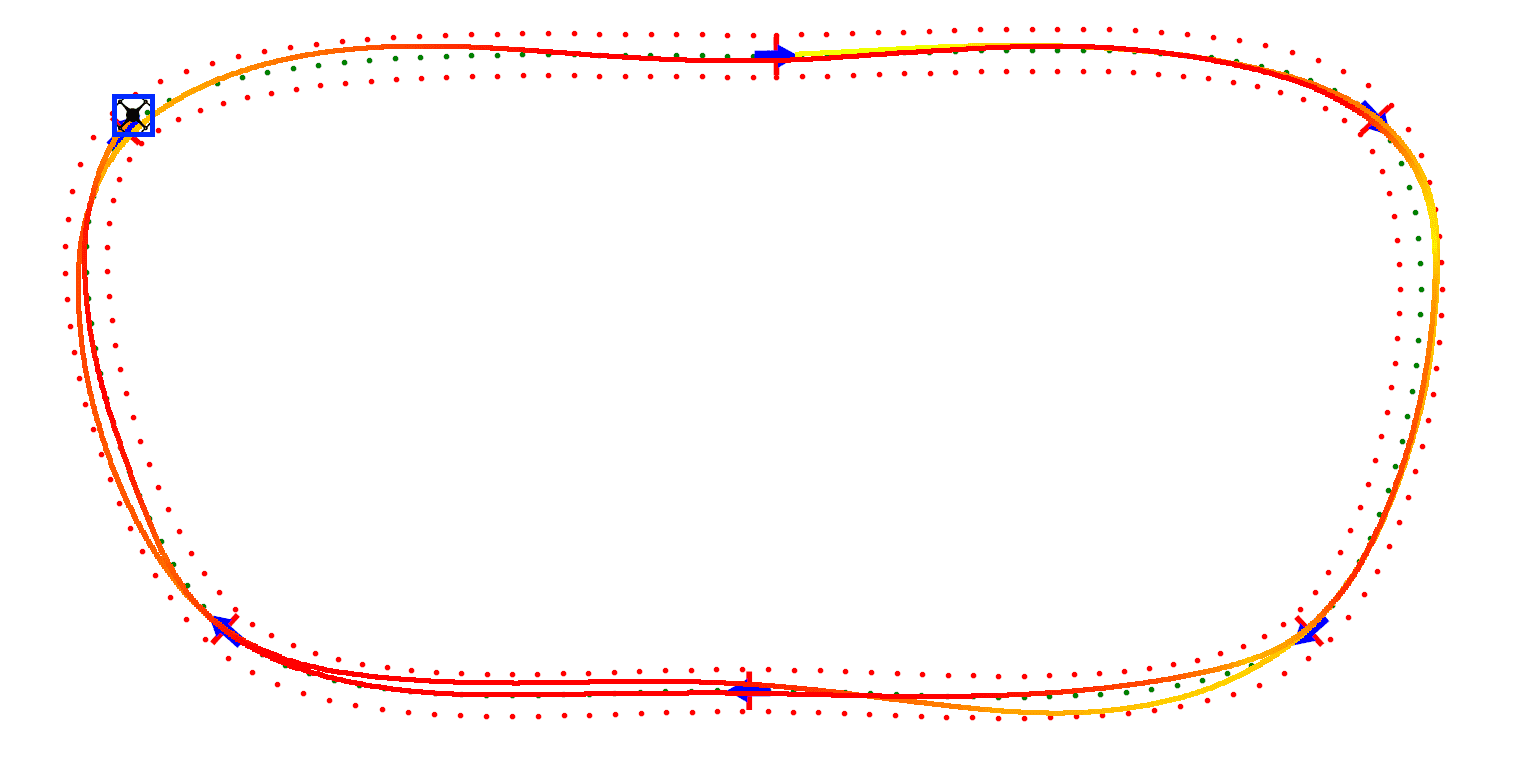}\\
		\small (g) Novice &
		\small (h) Intermediate &
		\small (i) Professional \\
		\includegraphics[height=3cm]{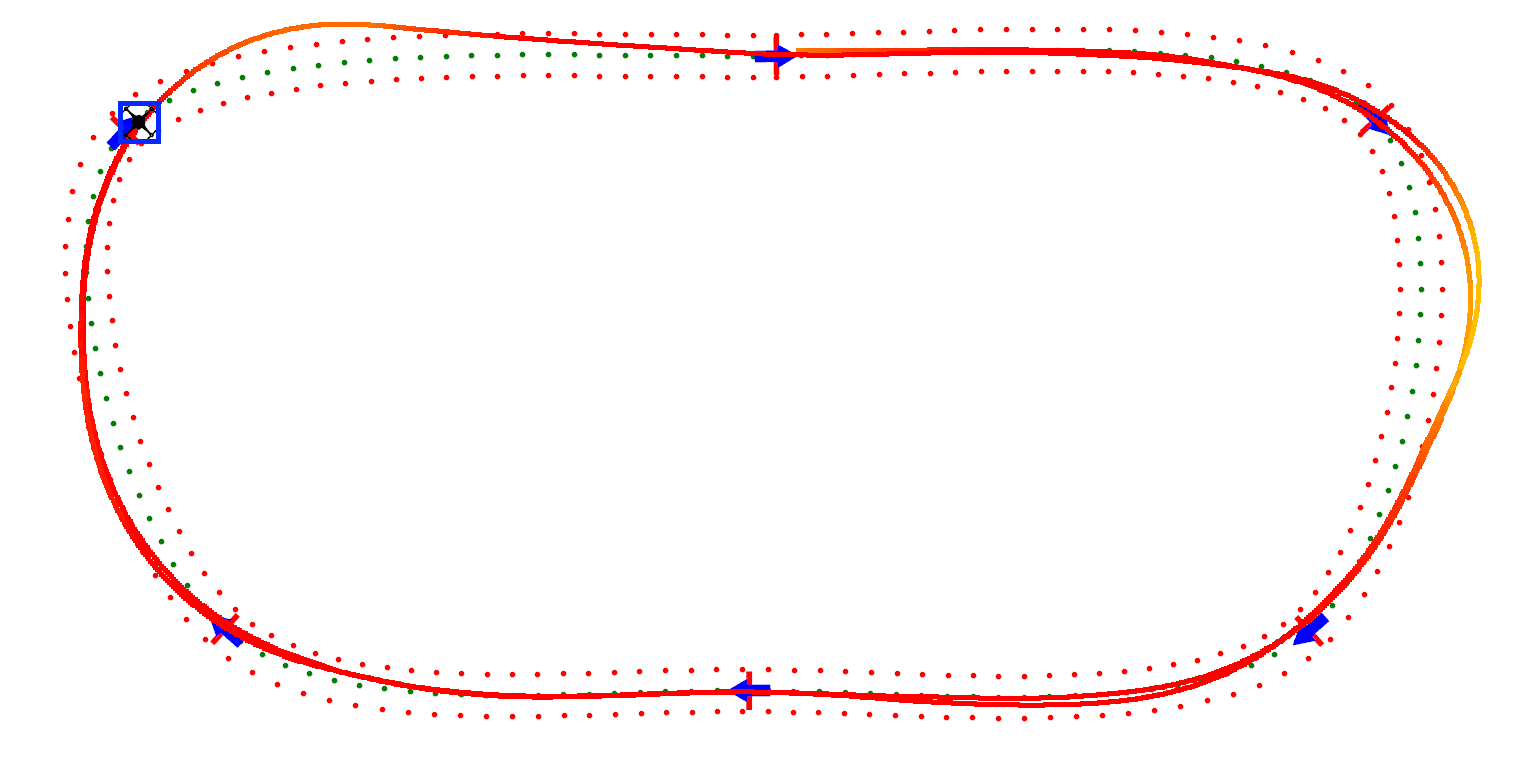} &
		\includegraphics[height=3cm]{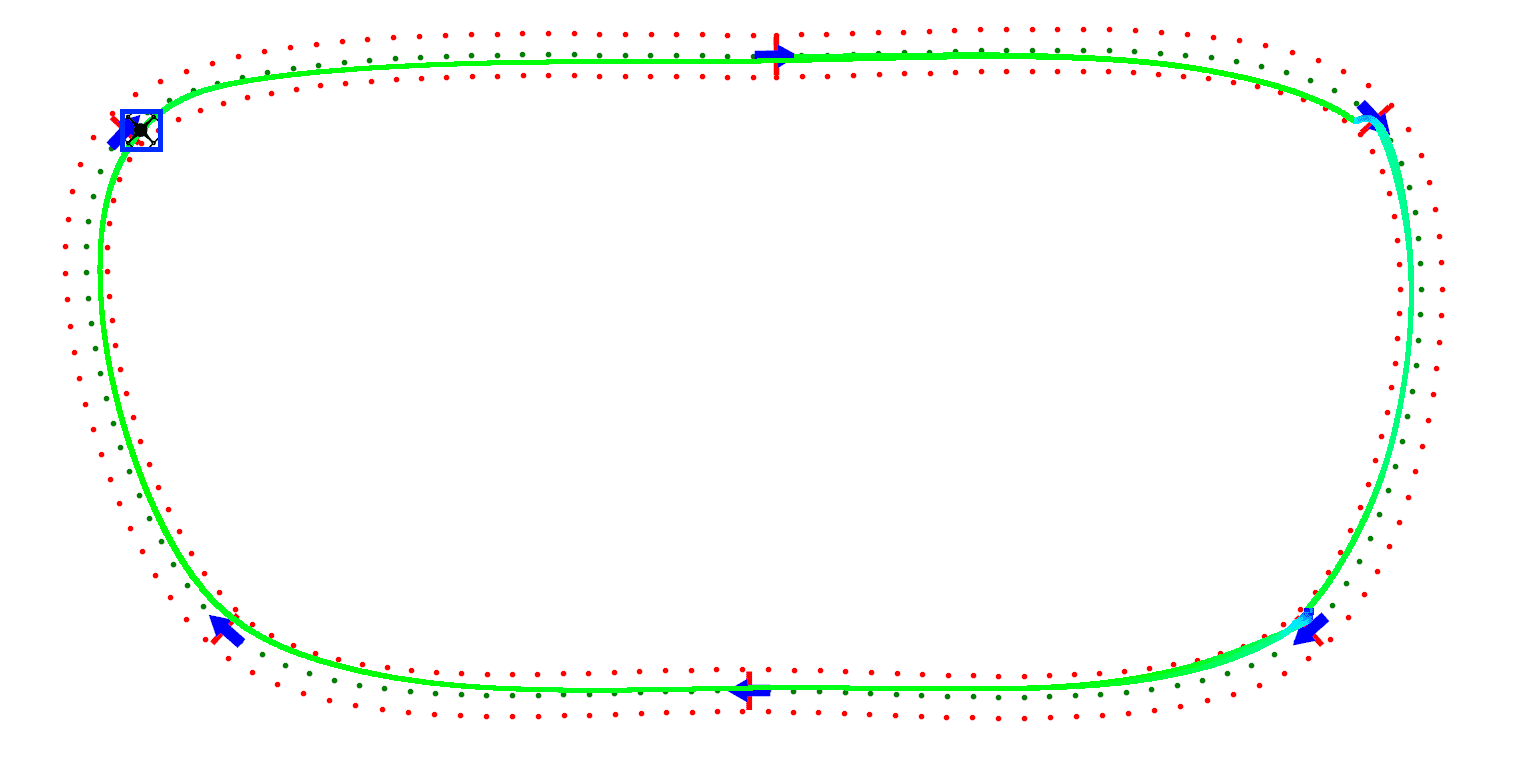} &
		\includegraphics[height=3cm]{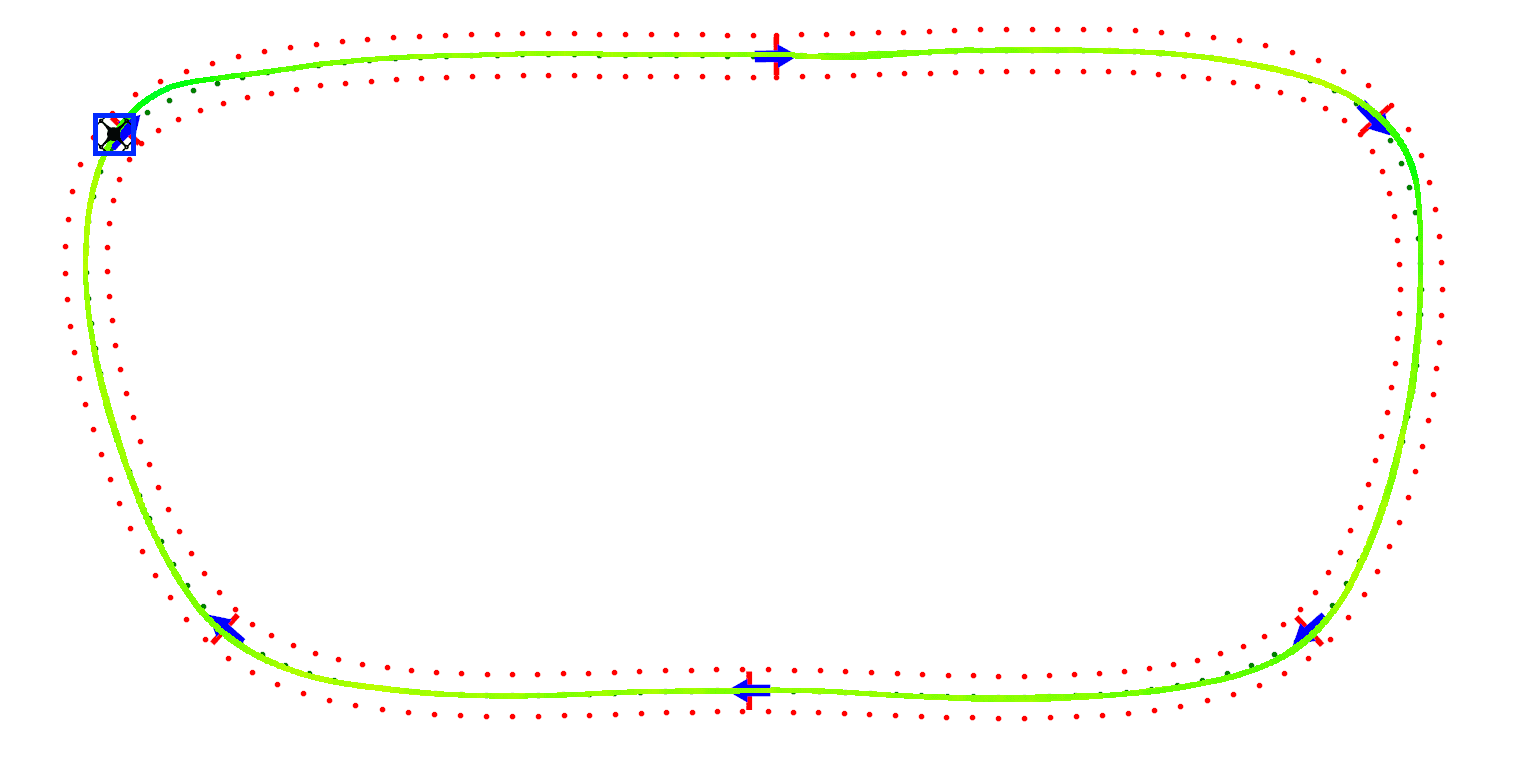}\\
		\small (j) Behaviour Cloning &
		\small (k) Dagger &
		\small (l) DDPG \\
       \multicolumn{3}{c}{\includegraphics[height=1.2cm]{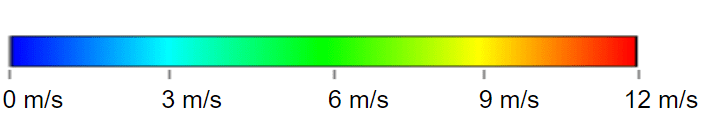}} 
\end{tabular}
\captionof{figure}{Comparison between our learned policy and its teachers (\emph{row 1,2}), human pilots (\emph{row 3}) and baselines (\emph{row 4}) on test track 1. Color encodes speed as a heatmap, where blue is the minimum speed and red is the maximum speed.}
\label{fig:qualitive_results_track1}
\end{figure*}

\begin{figure*}
\centering
\begin{tabular}{@{}c@{\hspace{1mm}}c@{\hspace{1mm}}c@{}}
		\includegraphics[height=3cm]{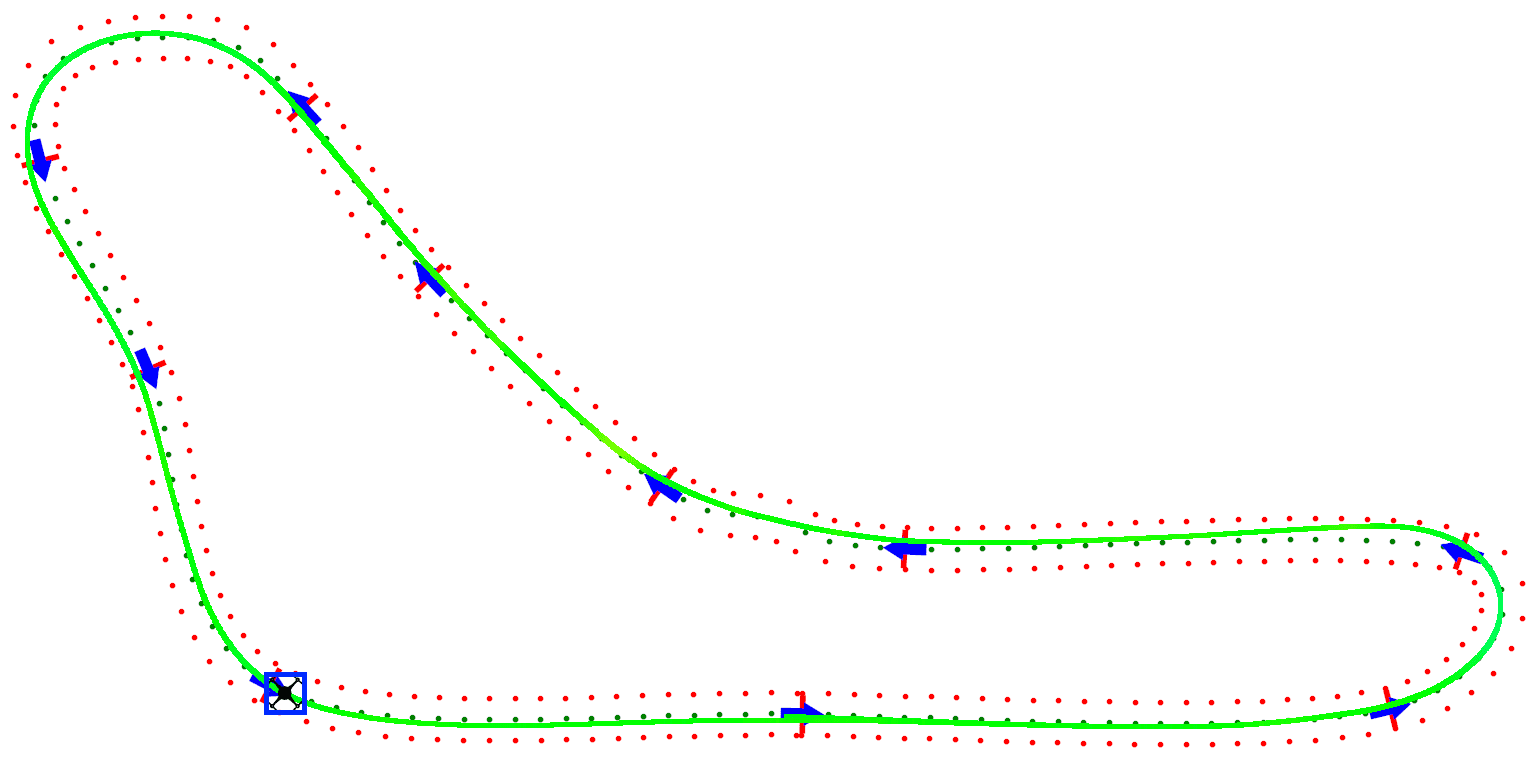} &
		\includegraphics[height=3cm]{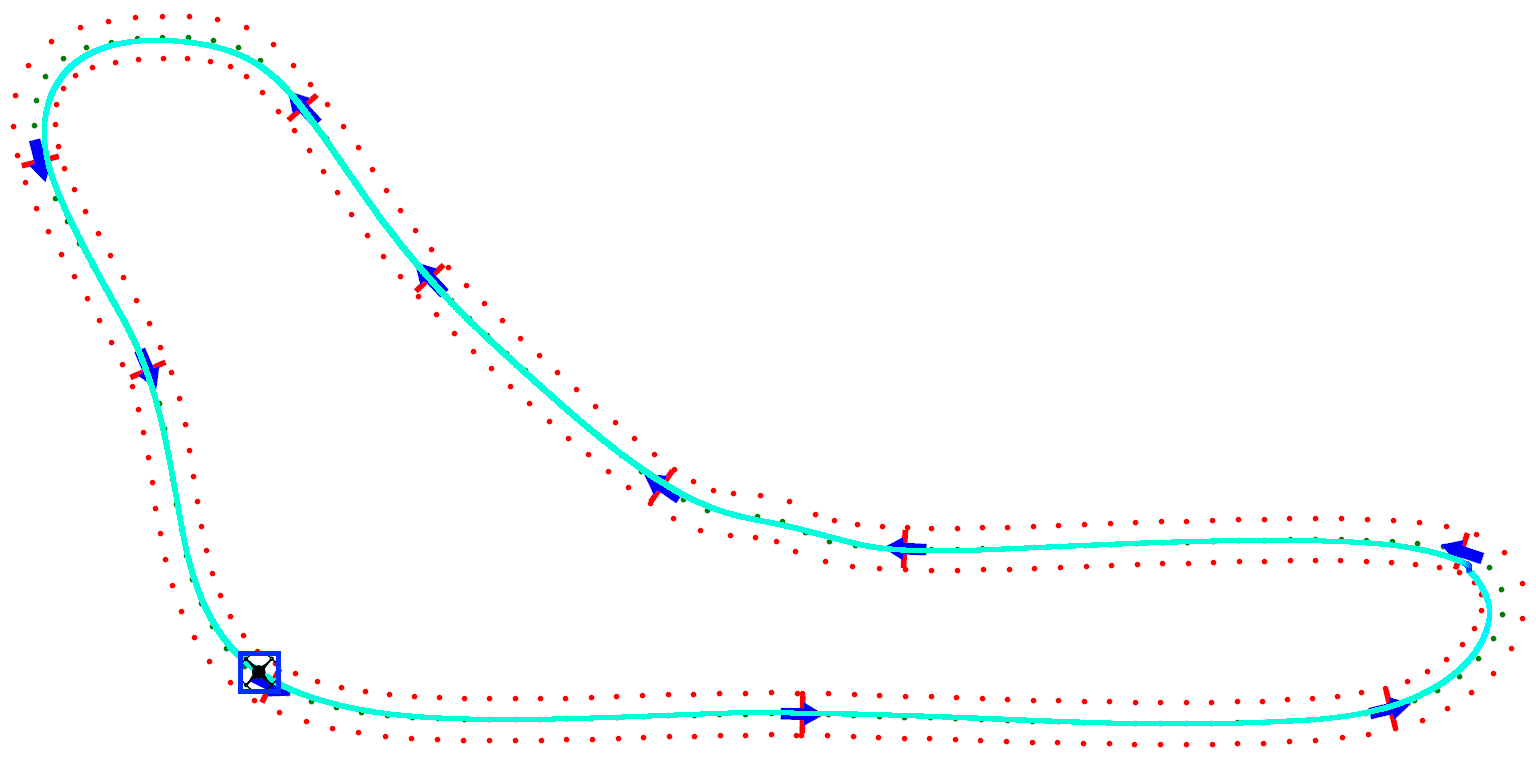} &
		\includegraphics[height=3cm]{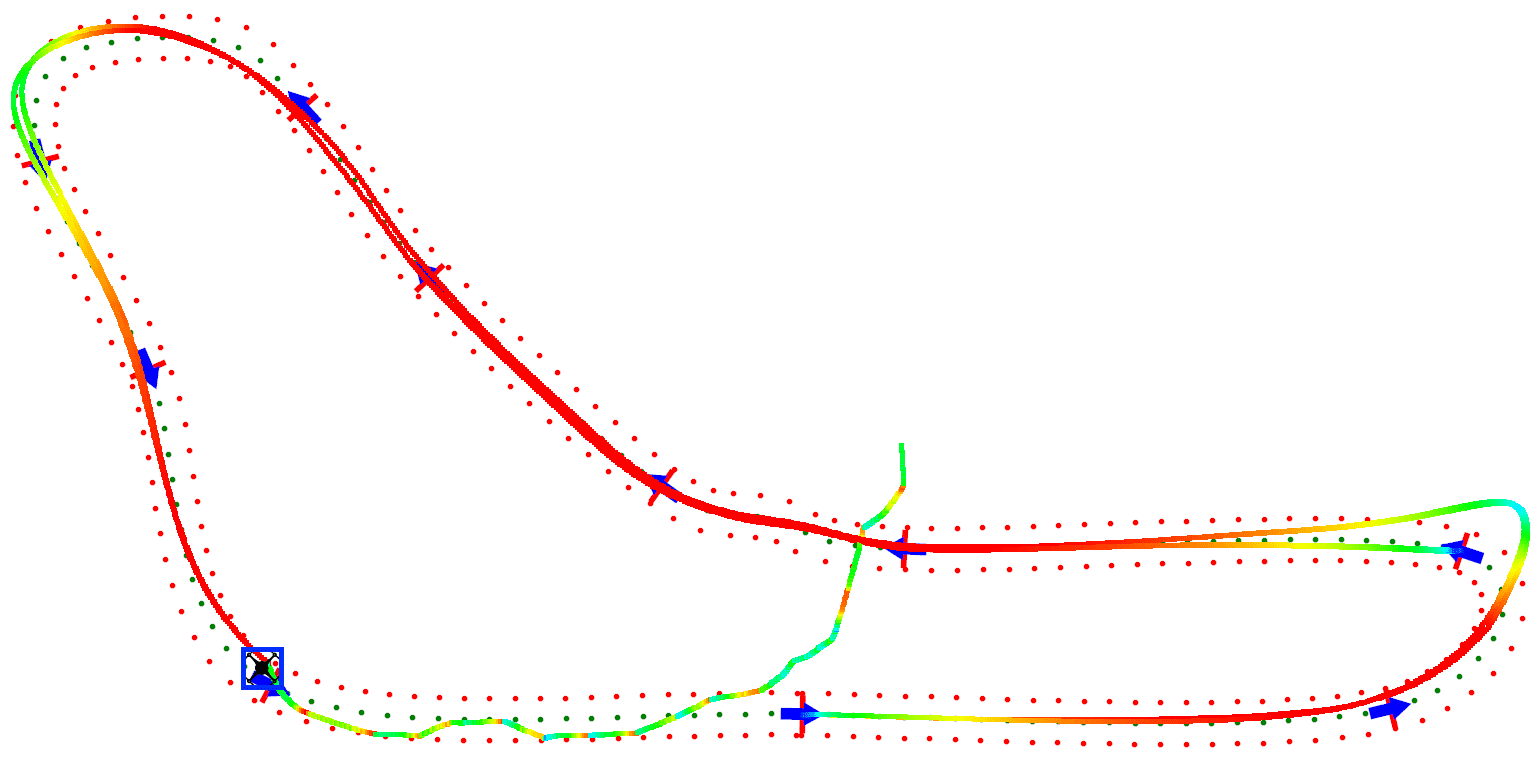} \\
		\small (a) OIL &
		\small (b) Teacher 1 &
		\small (c) Teacher 2 \\
		\includegraphics[height=3cm]{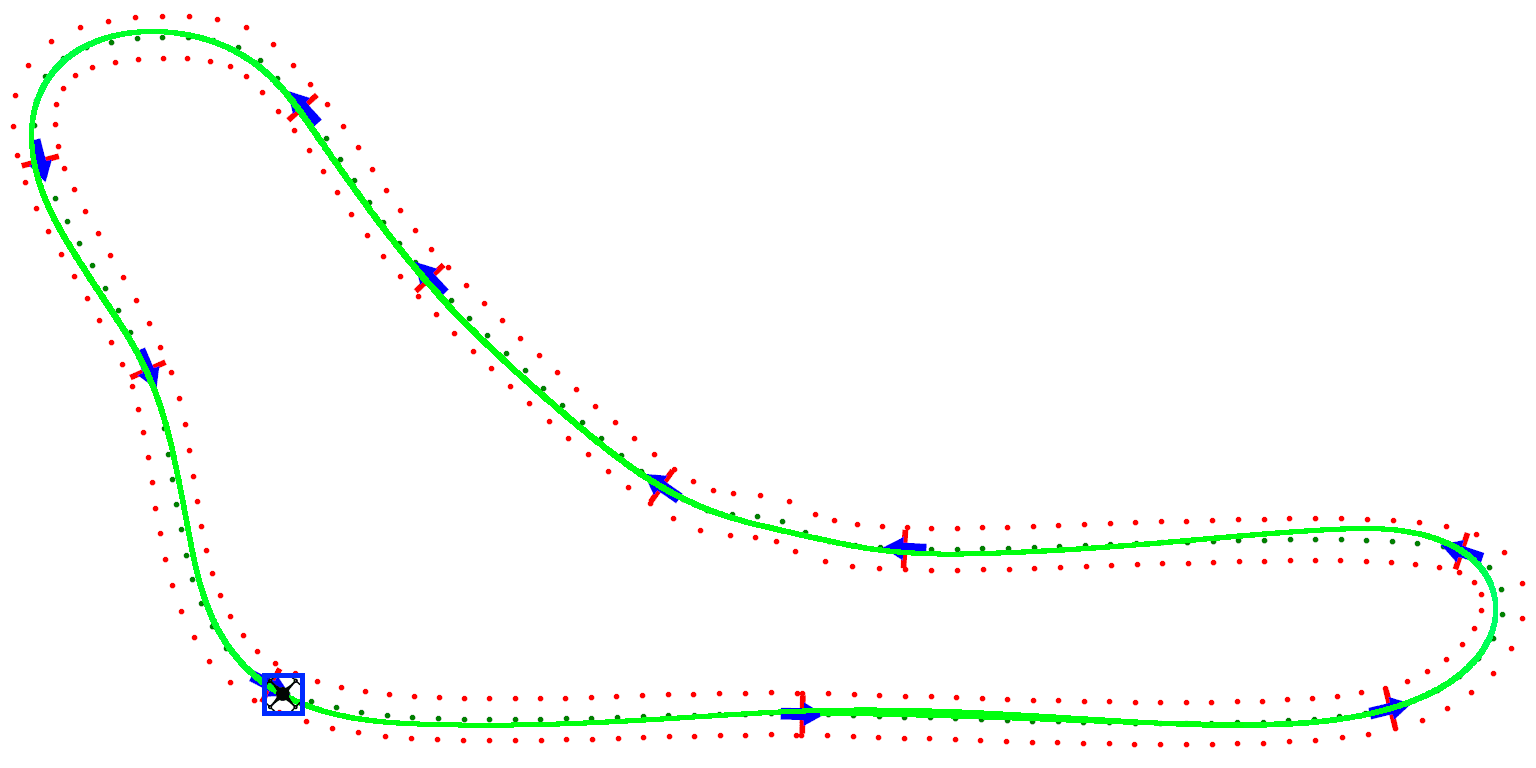} &
		\includegraphics[height=3cm]{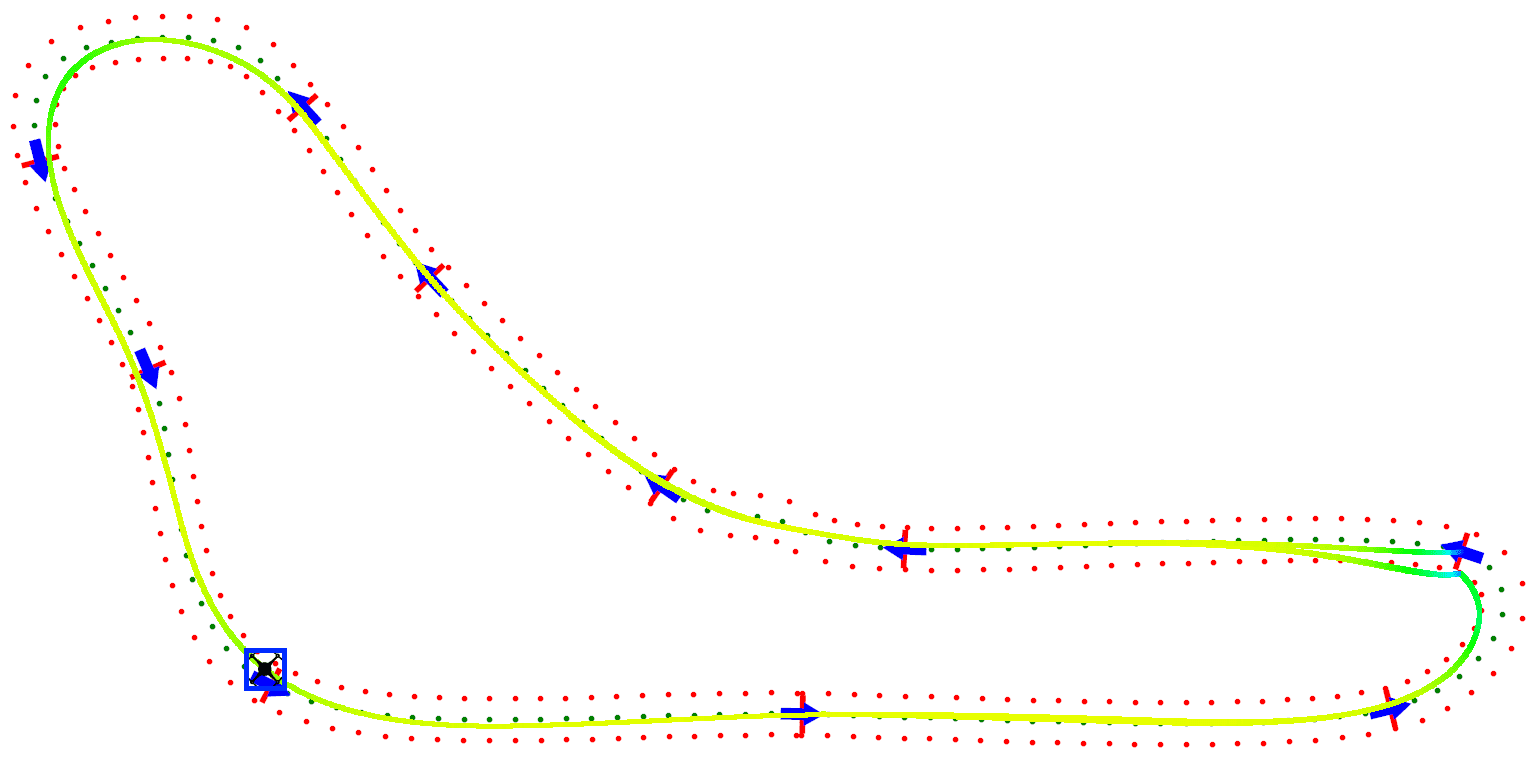} &
		\includegraphics[height=3cm]{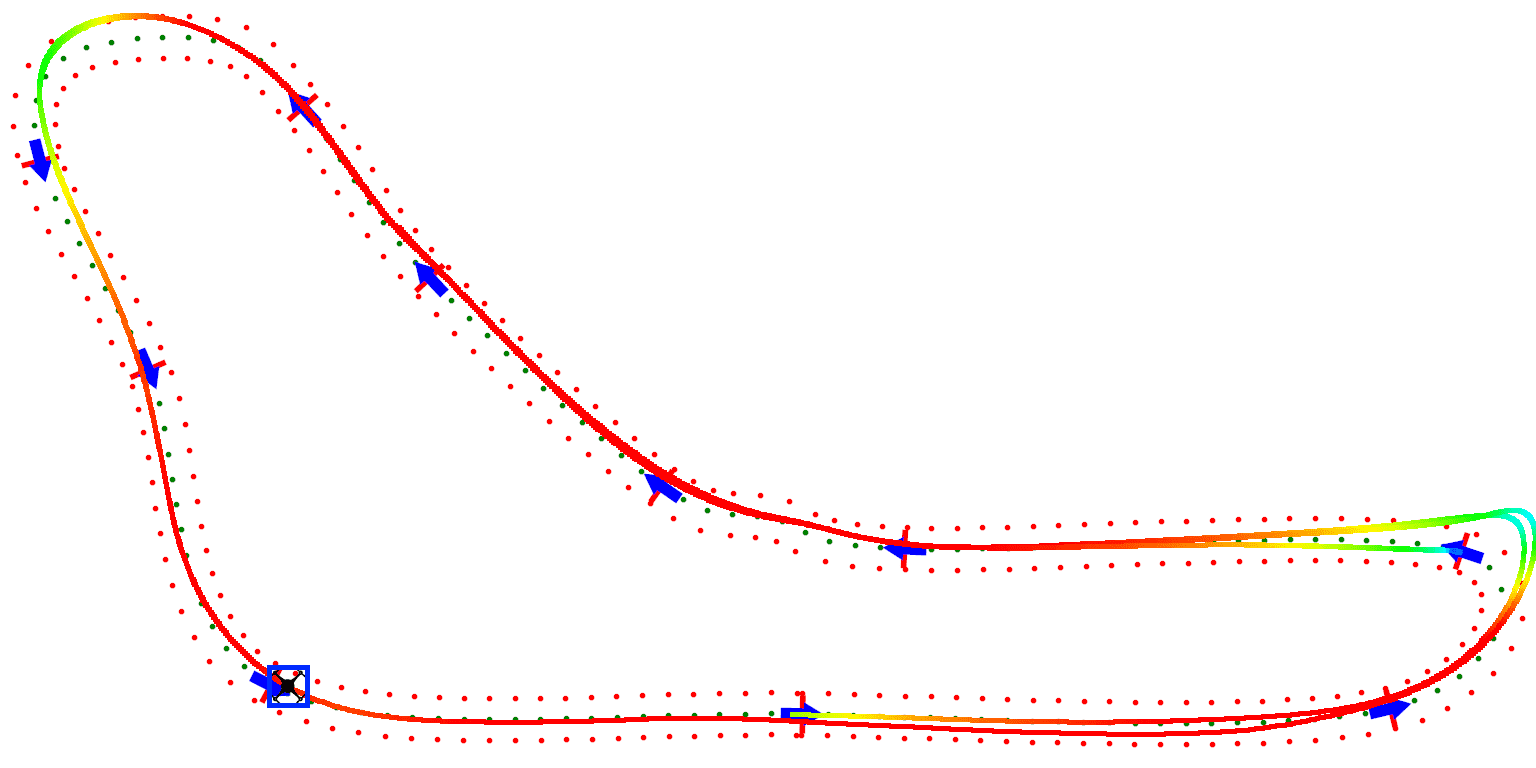} \\
		\small (d) Teacher 3 &
		\small (e) Teacher 4 &
		\small (f) Teacher 5 \\
		\includegraphics[height=3cm]{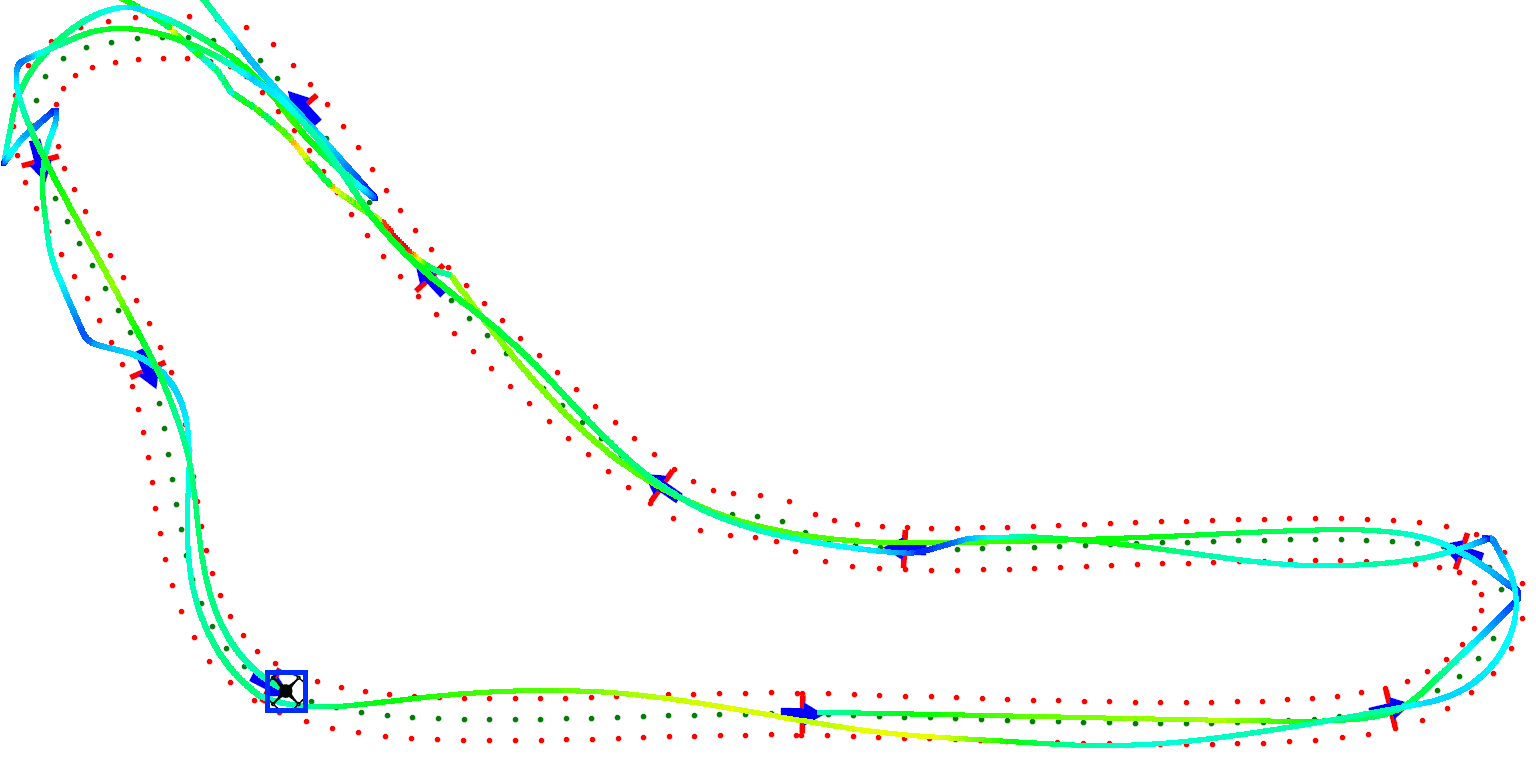} &
		\includegraphics[height=3cm]{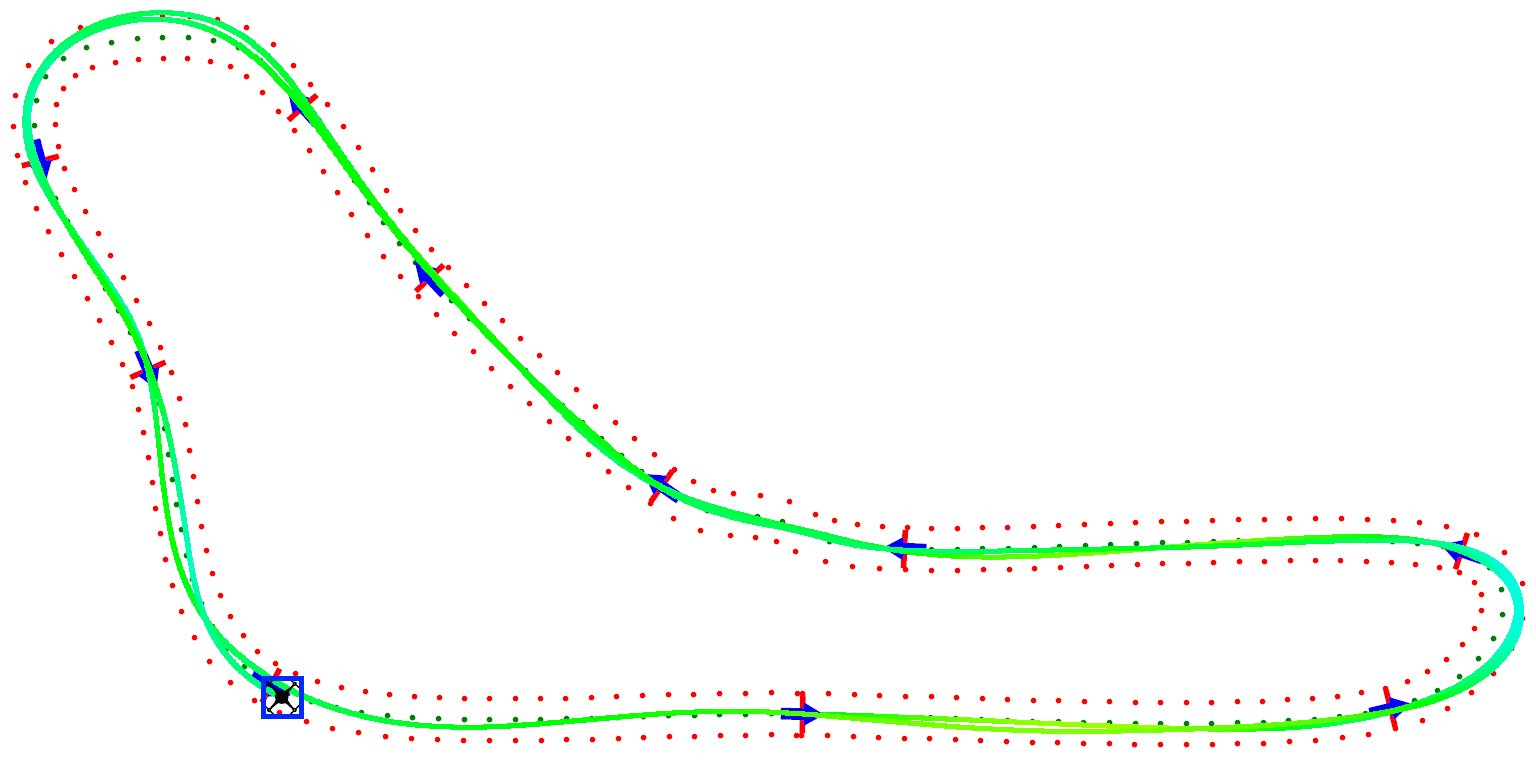} &
		\includegraphics[height=3cm]{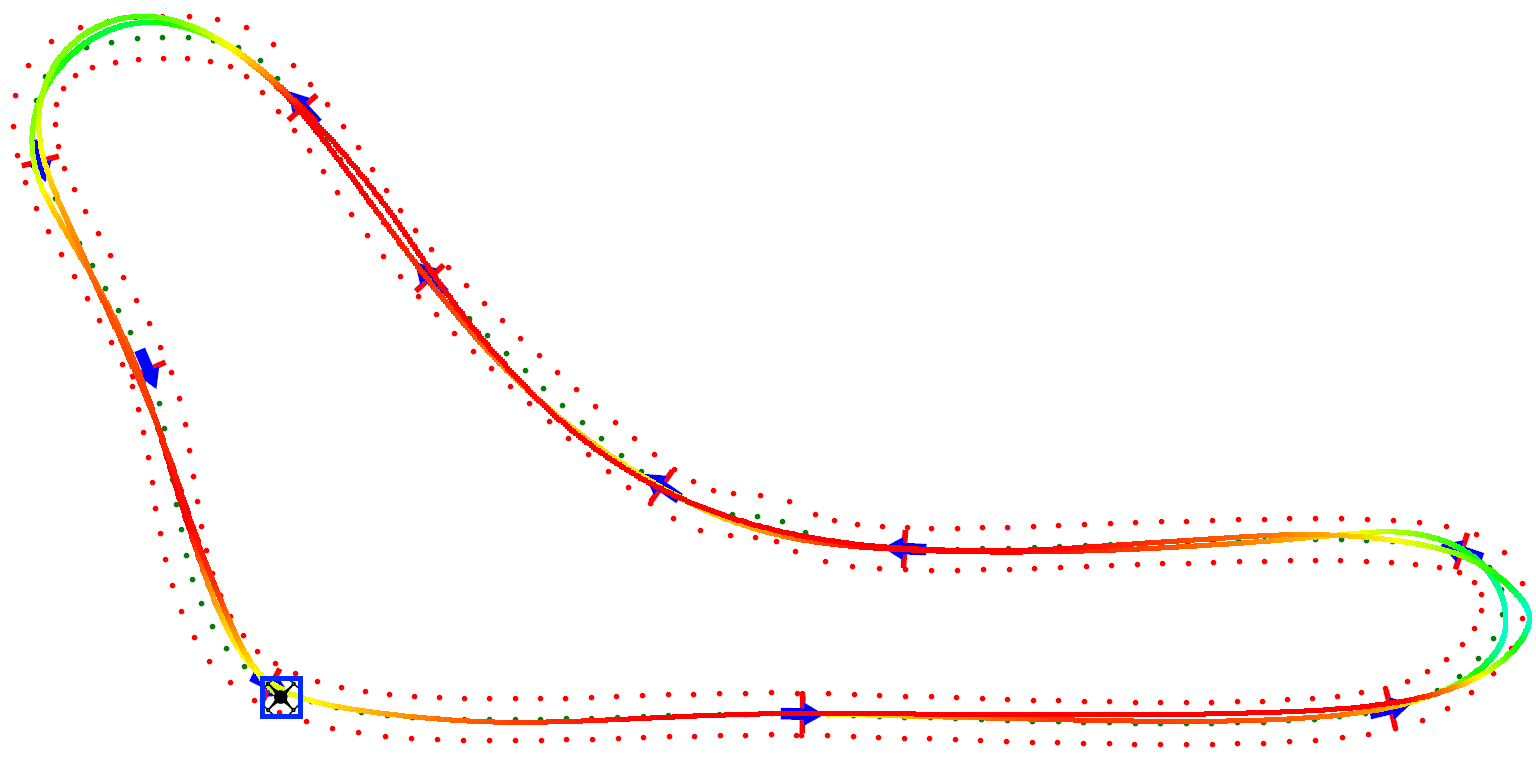}\\
		\small (g) Novice &
		\small (h) Intermediate &
		\small (i) Professional \\
		\includegraphics[height=3cm]{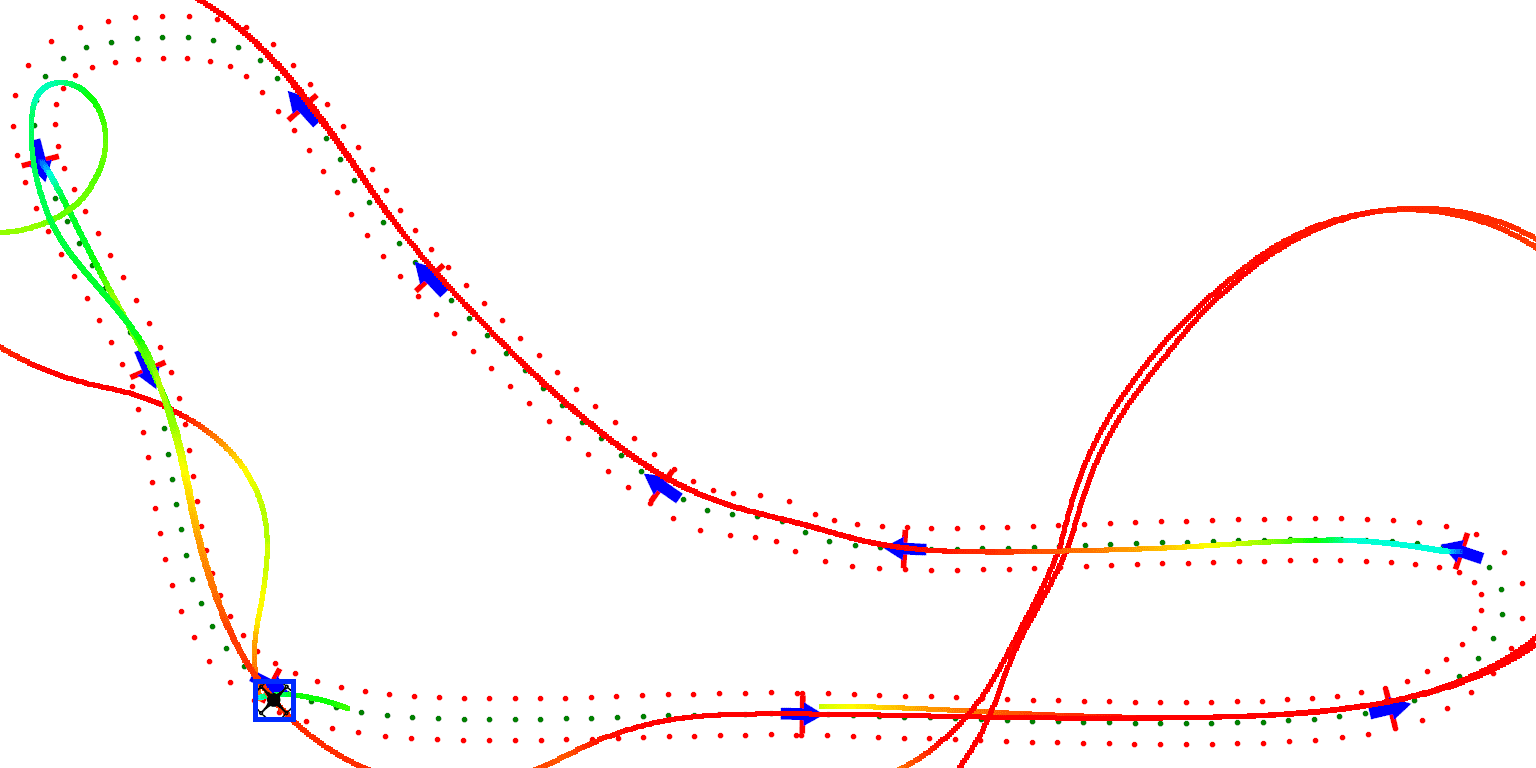} &
		\includegraphics[height=3cm]{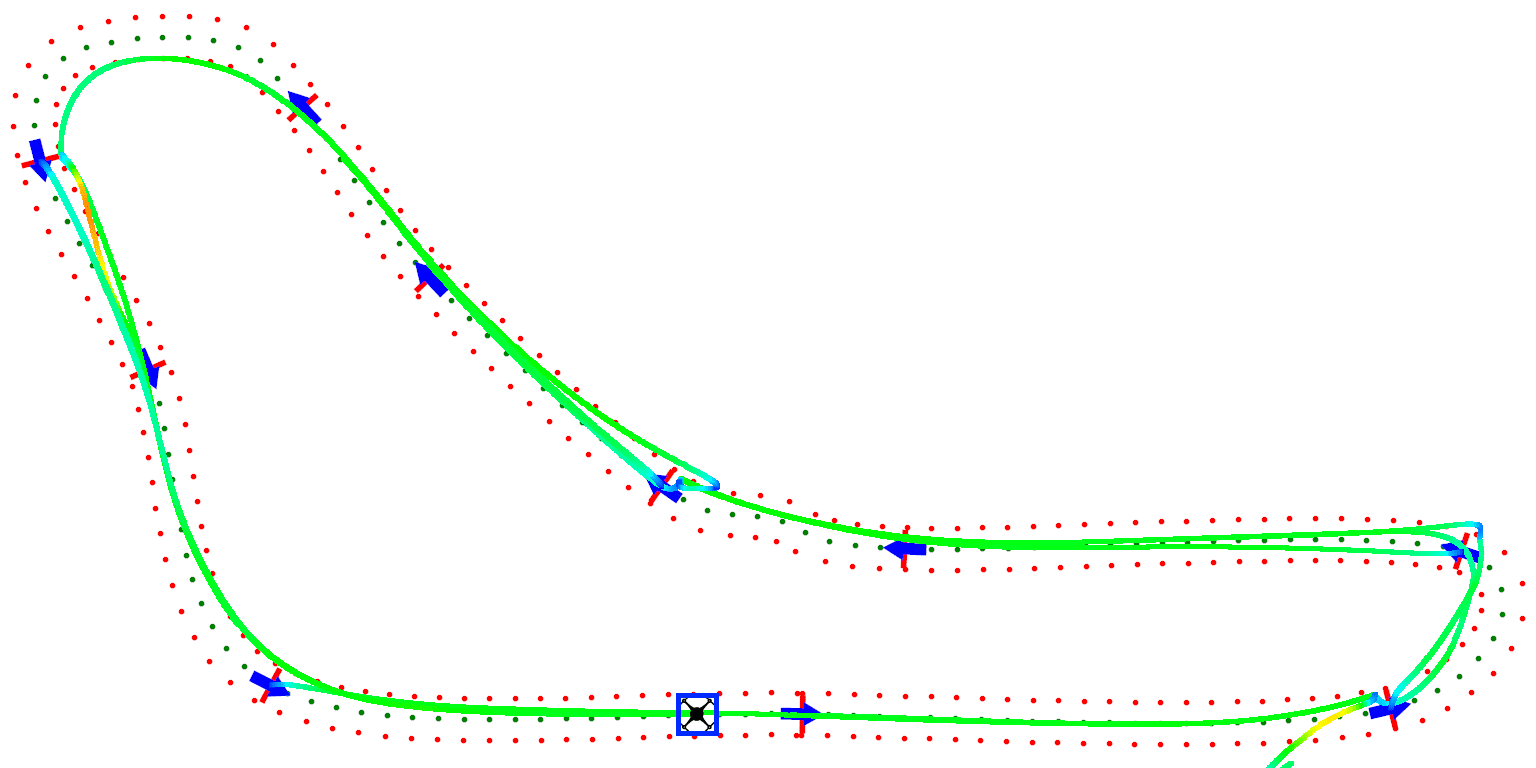} &
		\includegraphics[height=3cm]{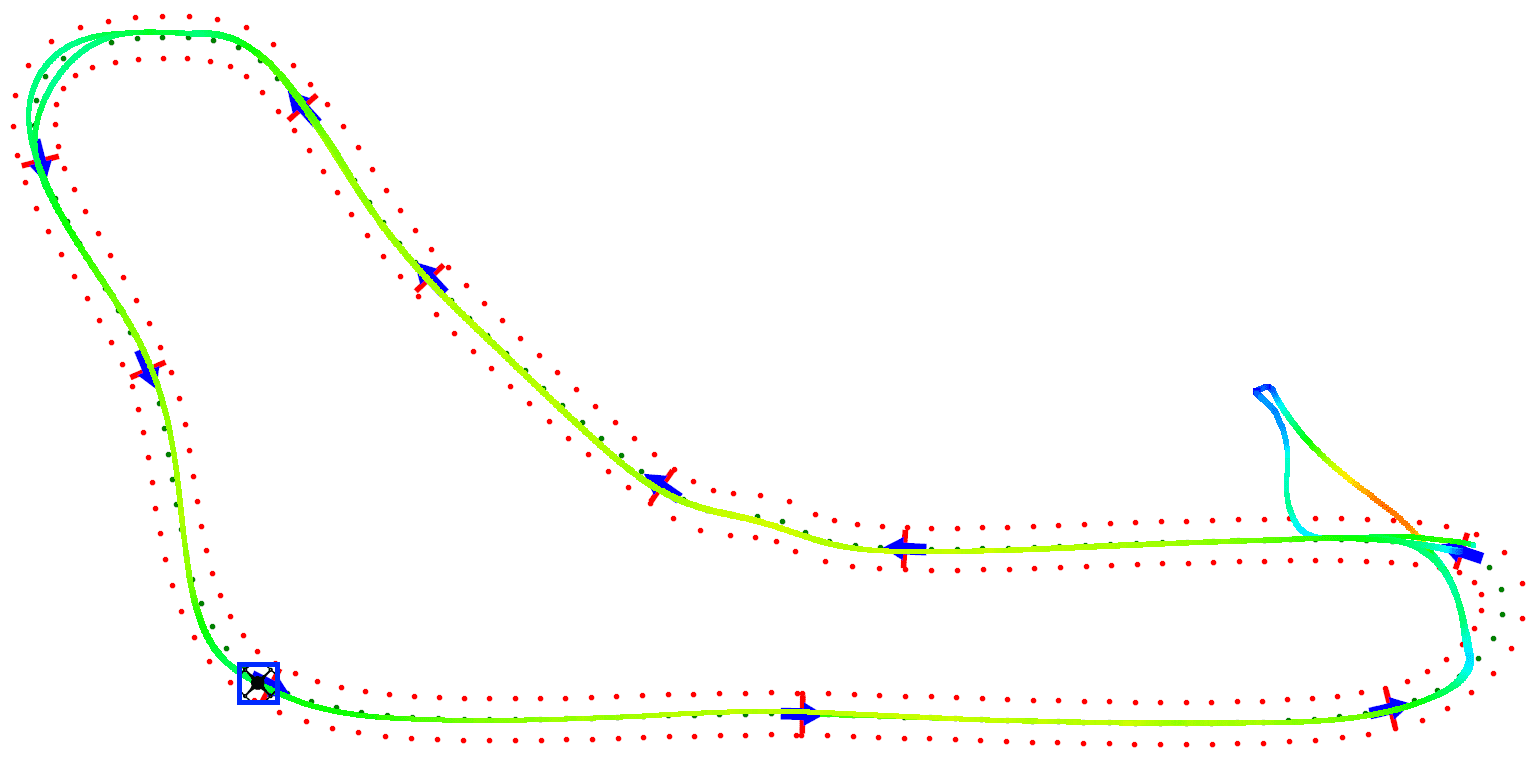}\\
		\small (j) Behaviour Cloning &
		\small (k) Dagger &
		\small (l) DDPG \\
       \multicolumn{3}{c}{\includegraphics[height=1.2cm]{figures/ColorScaleUAV.png}} 
\end{tabular}
\captionof{figure}{Comparison between our learned policy and its teachers (\emph{row 1,2}), human pilots (\emph{row 3}) and baselines (\emph{row 4}) on test track 2. Color encodes speed as a heatmap, where blue is the minimum speed and red is the maximum speed.}
\label{fig:qualitive_results_track2}
\end{figure*}

\begin{figure*}
\centering
\begin{tabular}{@{}c@{\hspace{1mm}}c@{\hspace{1mm}}c@{}}
		\includegraphics[height=3cm]{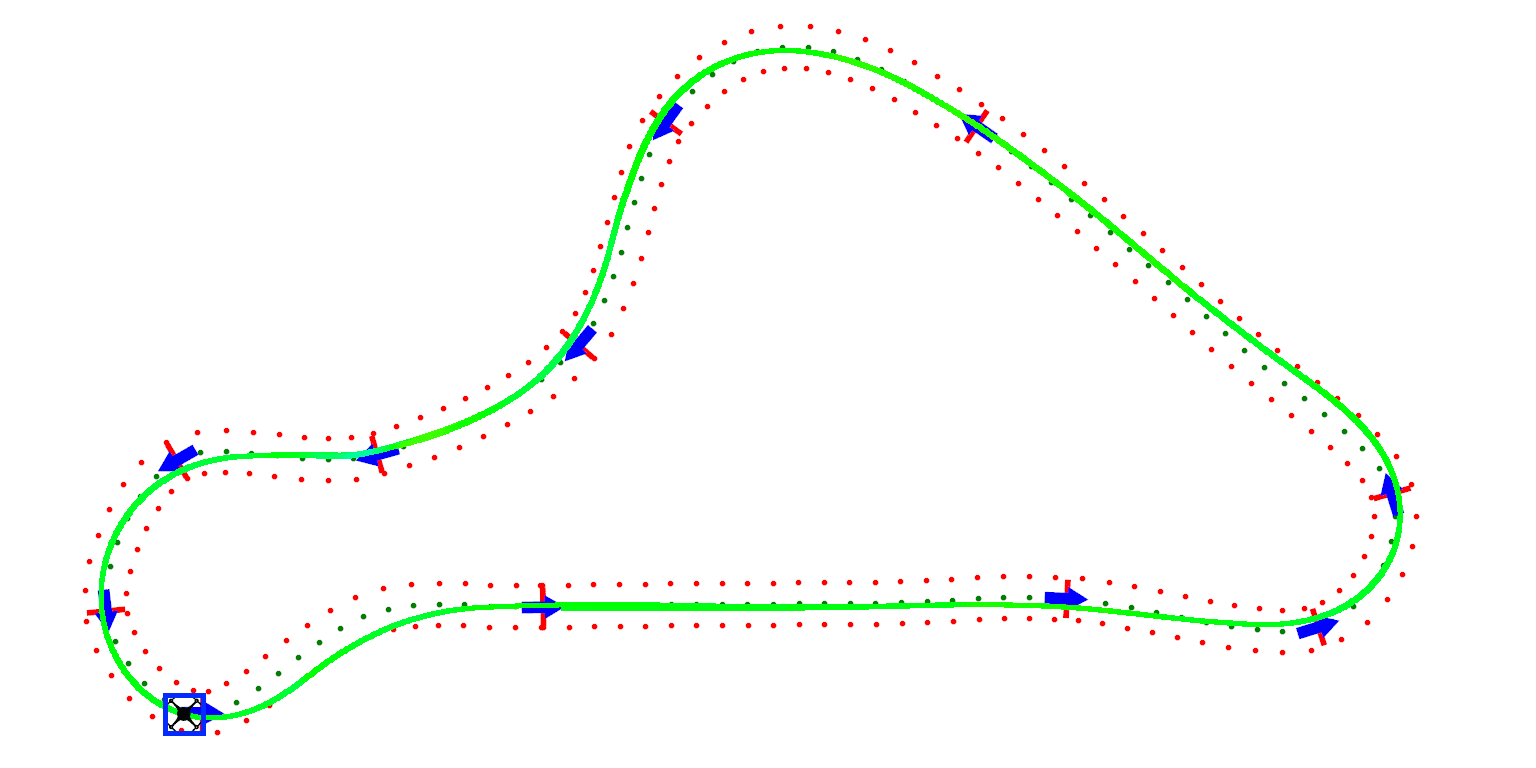} &
		\includegraphics[height=3cm]{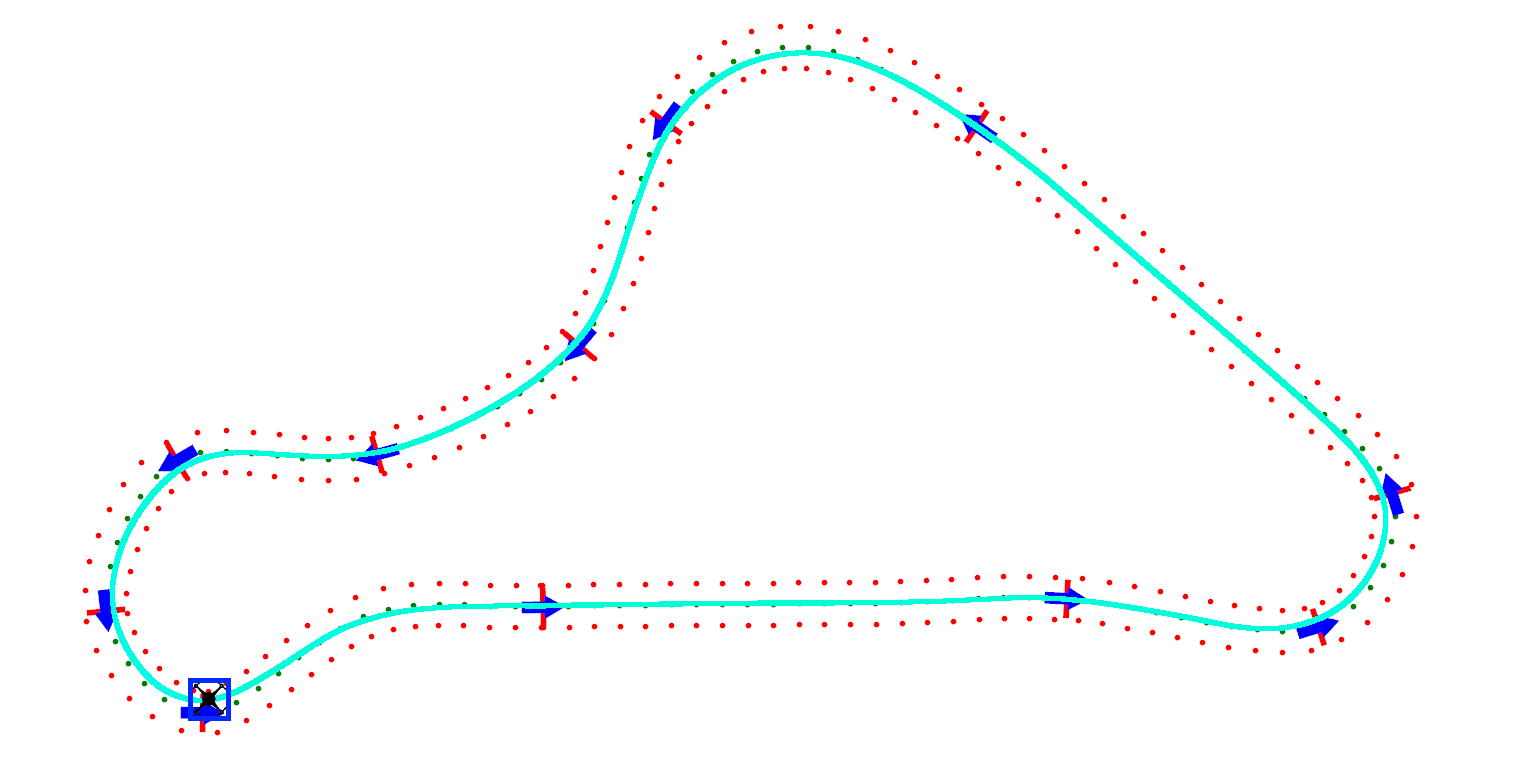} &
		\includegraphics[height=3cm]{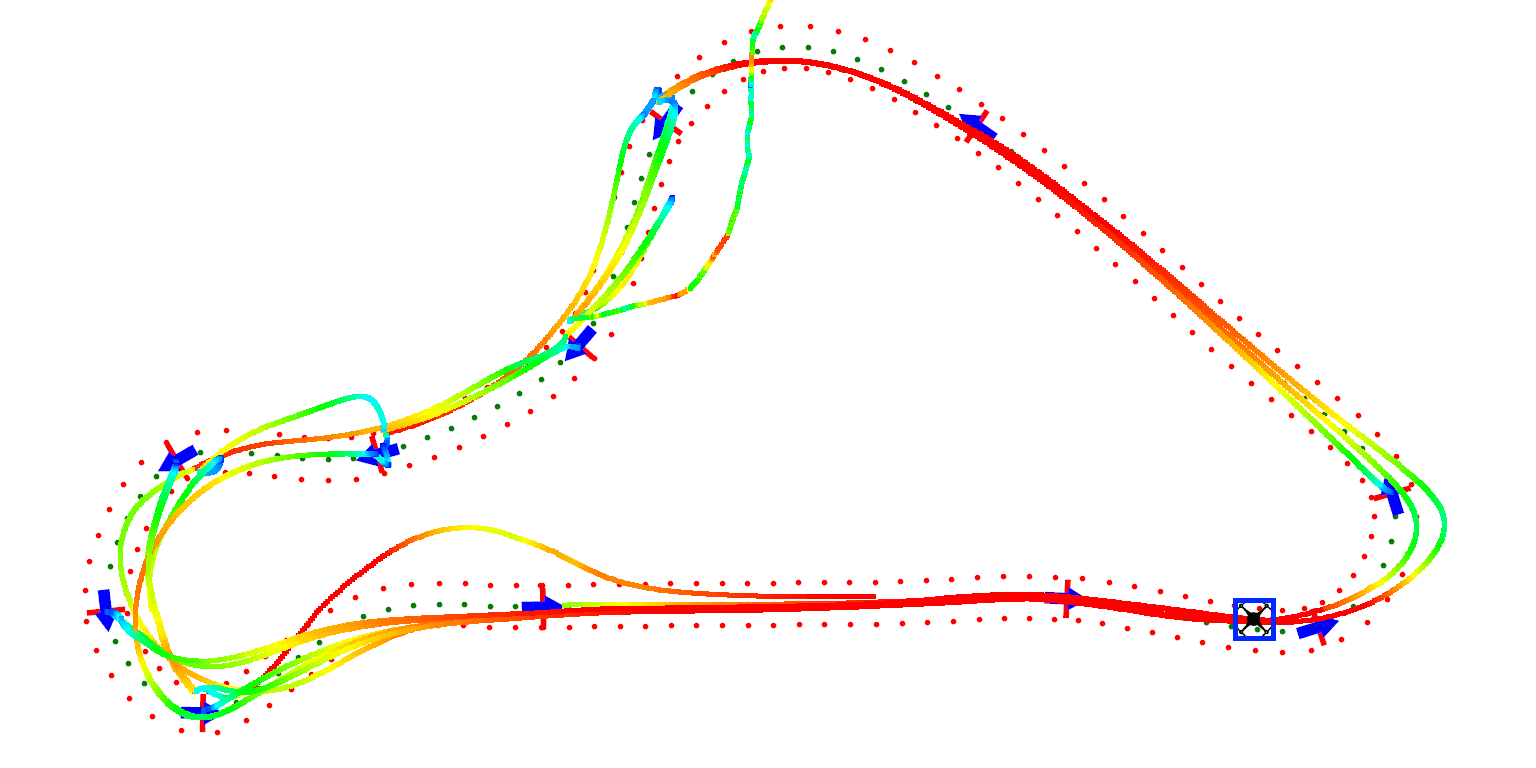} \\
		\small (a) OIL &
		\small (b) Teacher 1 &
		\small (c) Teacher 2 \\
		\includegraphics[height=3cm]{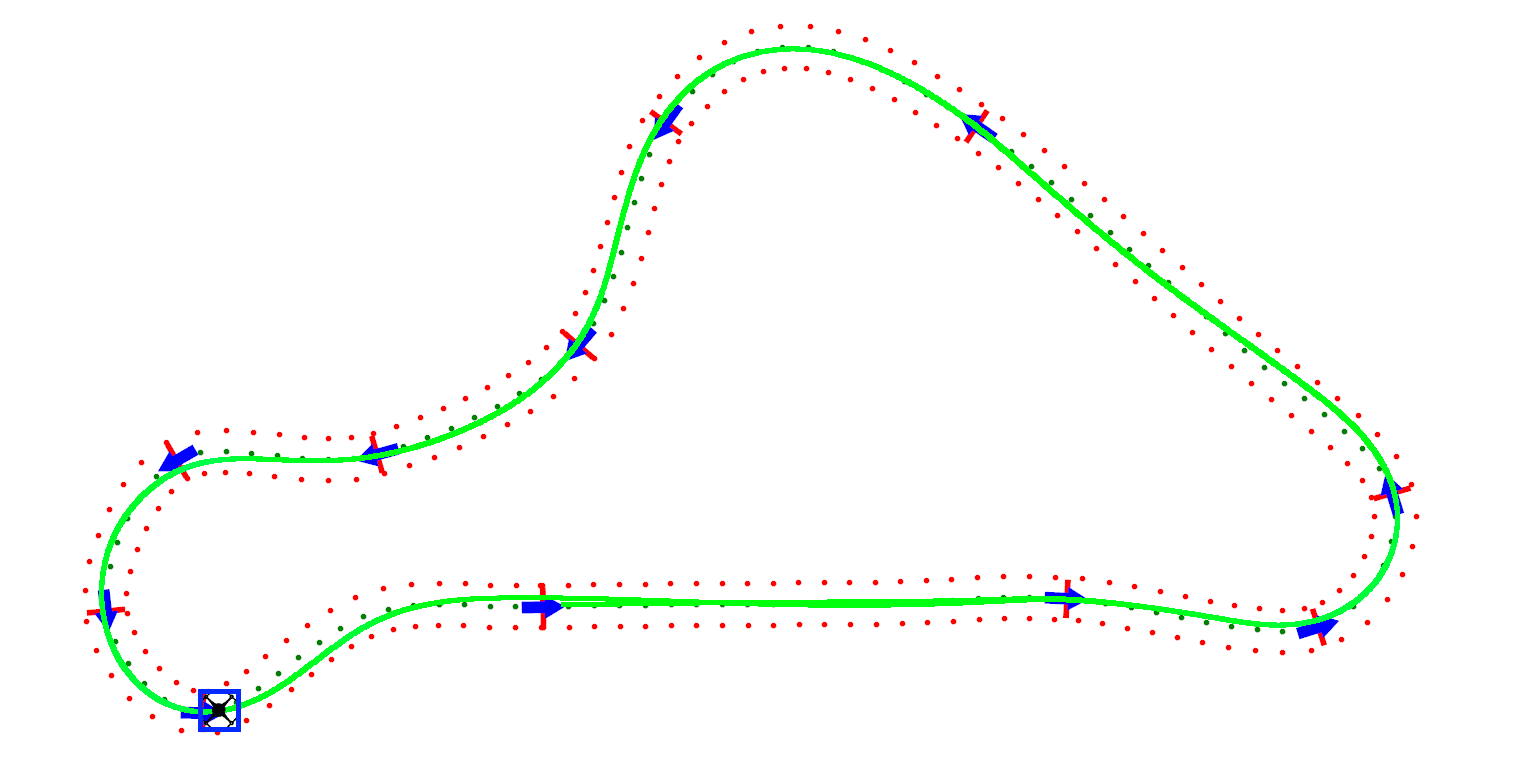} &
		\includegraphics[height=3cm]{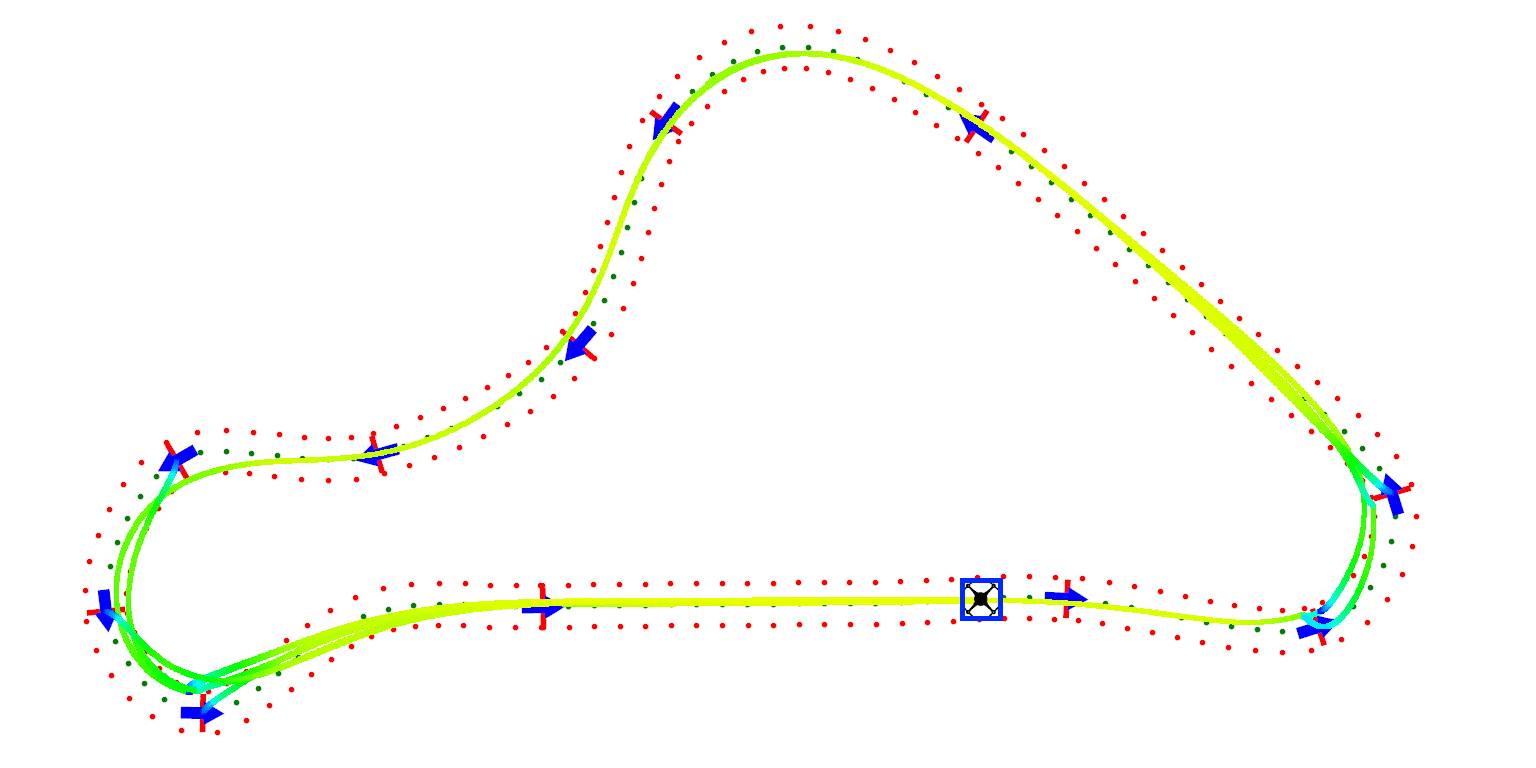} &
		\includegraphics[height=3cm]{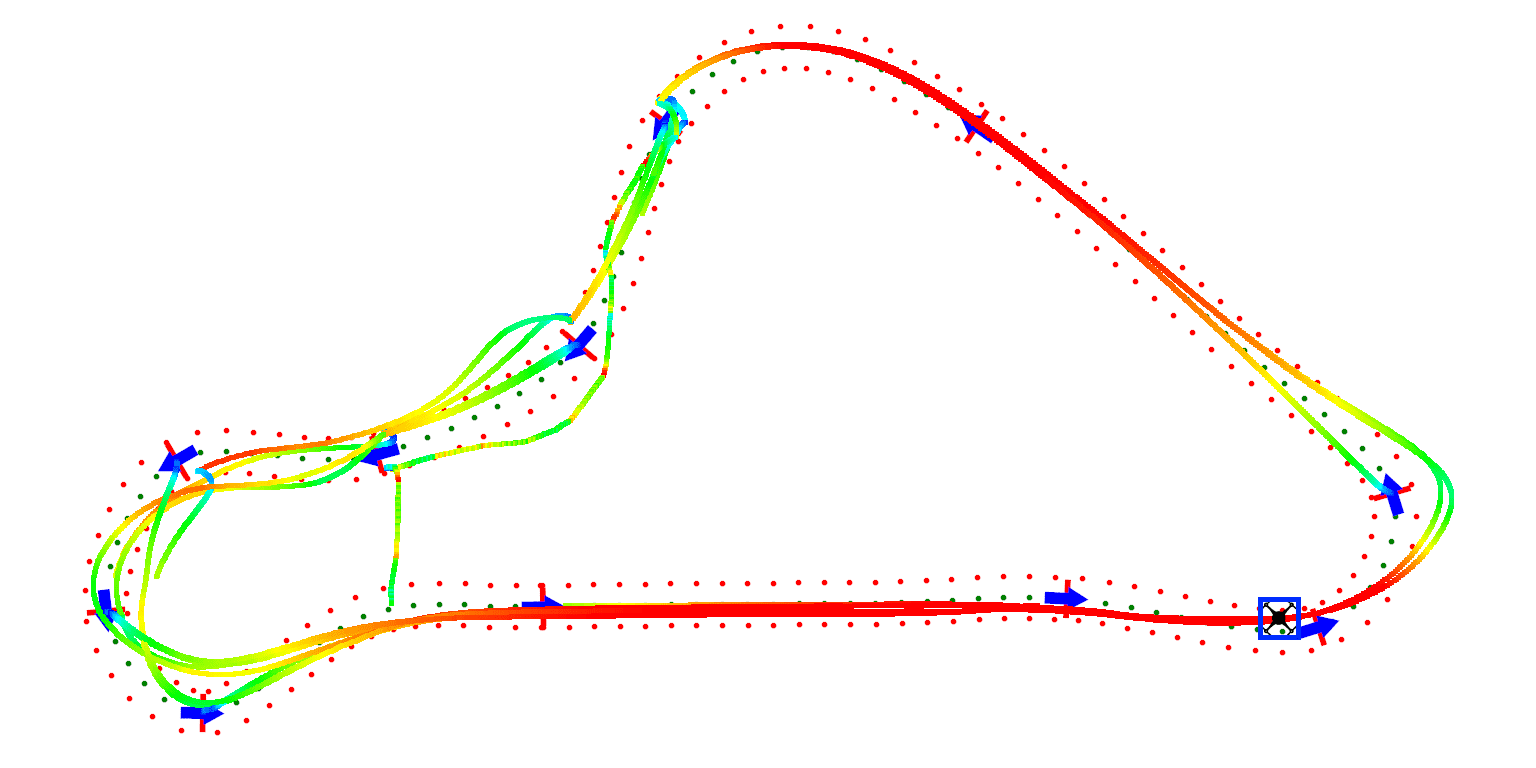} \\
		\small (d) Teacher 3 &
		\small (e) Teacher 4 &
		\small (f) Teacher 5 \\
		\includegraphics[height=3cm]{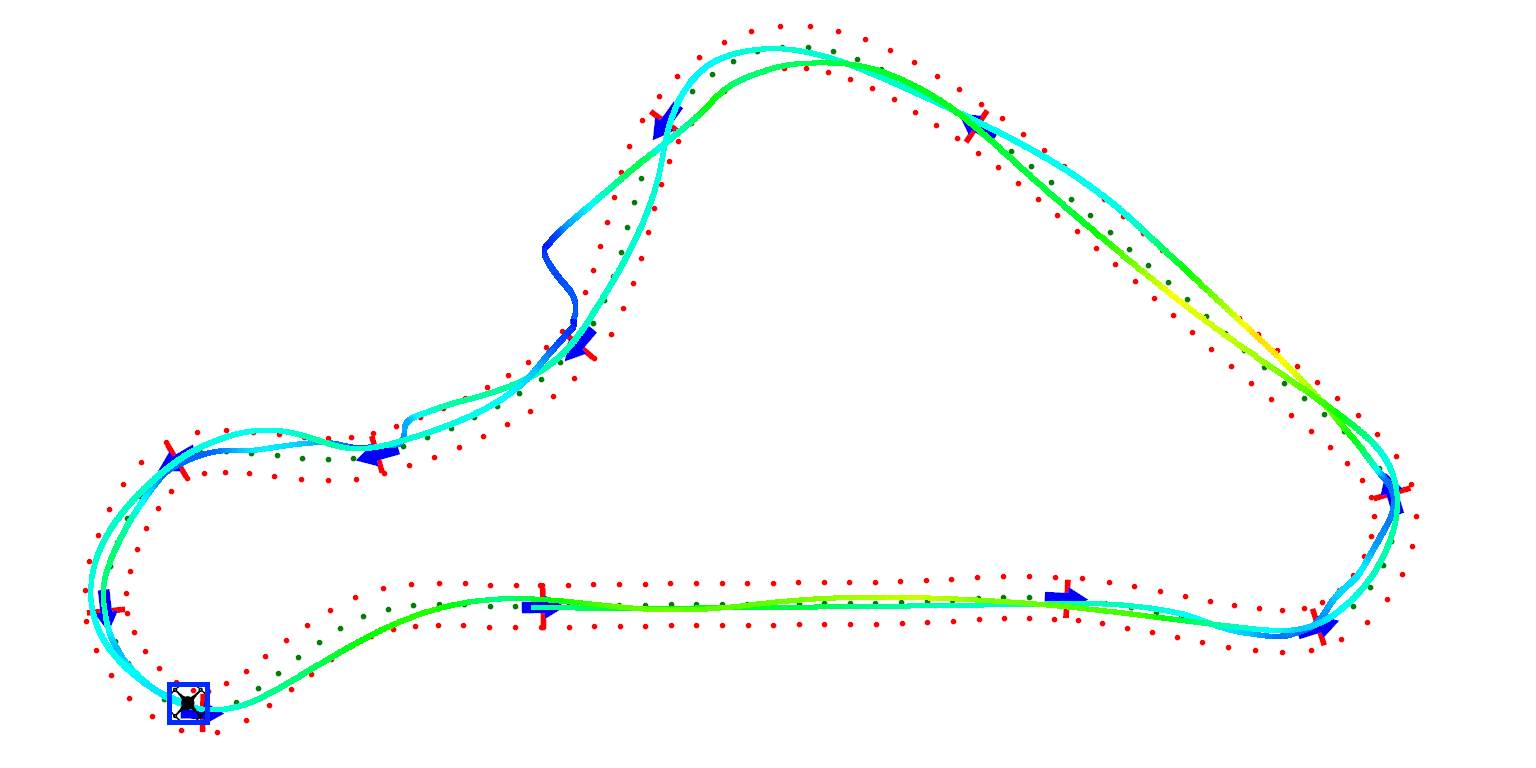} &
		\includegraphics[height=3cm]{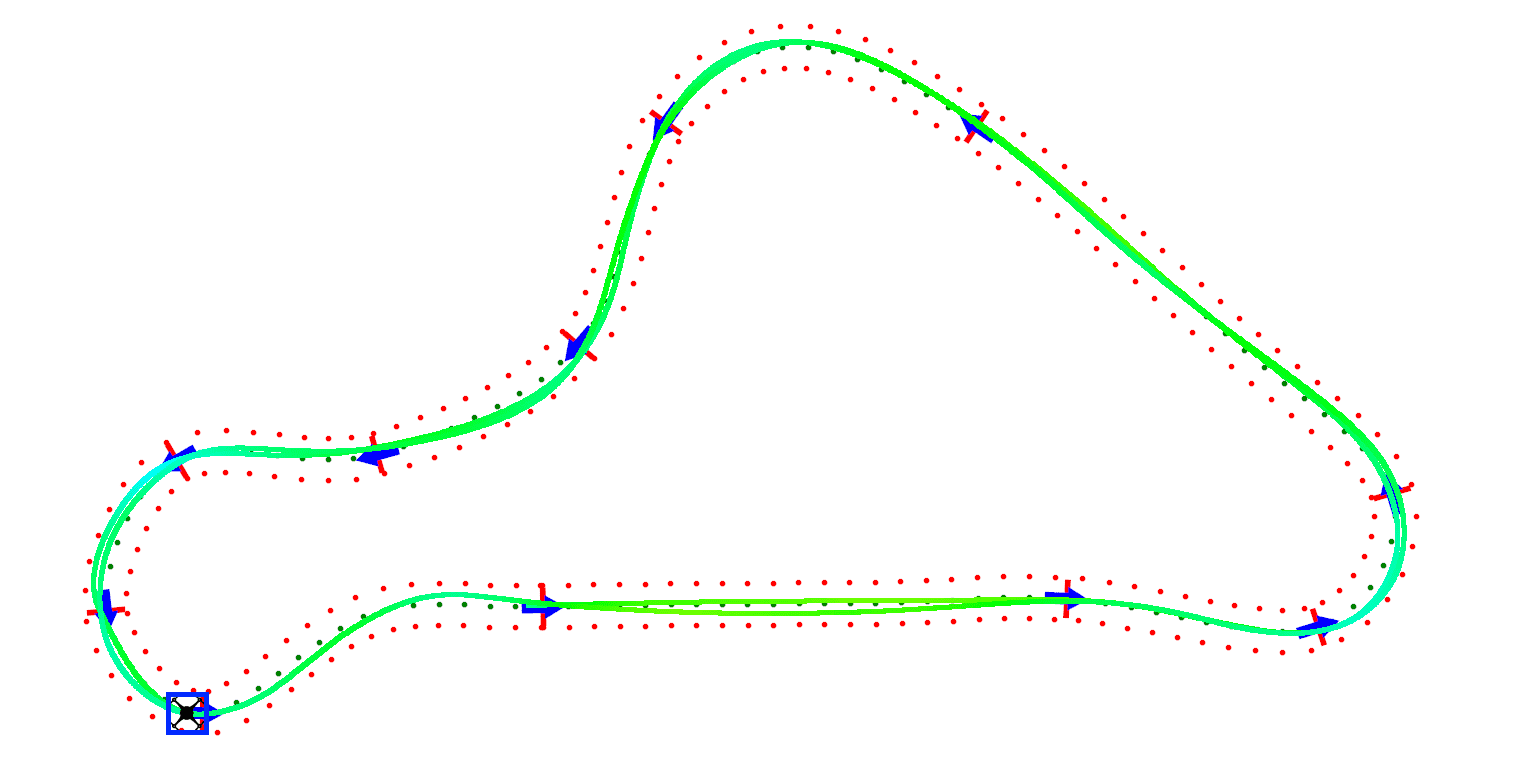} &
		\includegraphics[height=3cm]{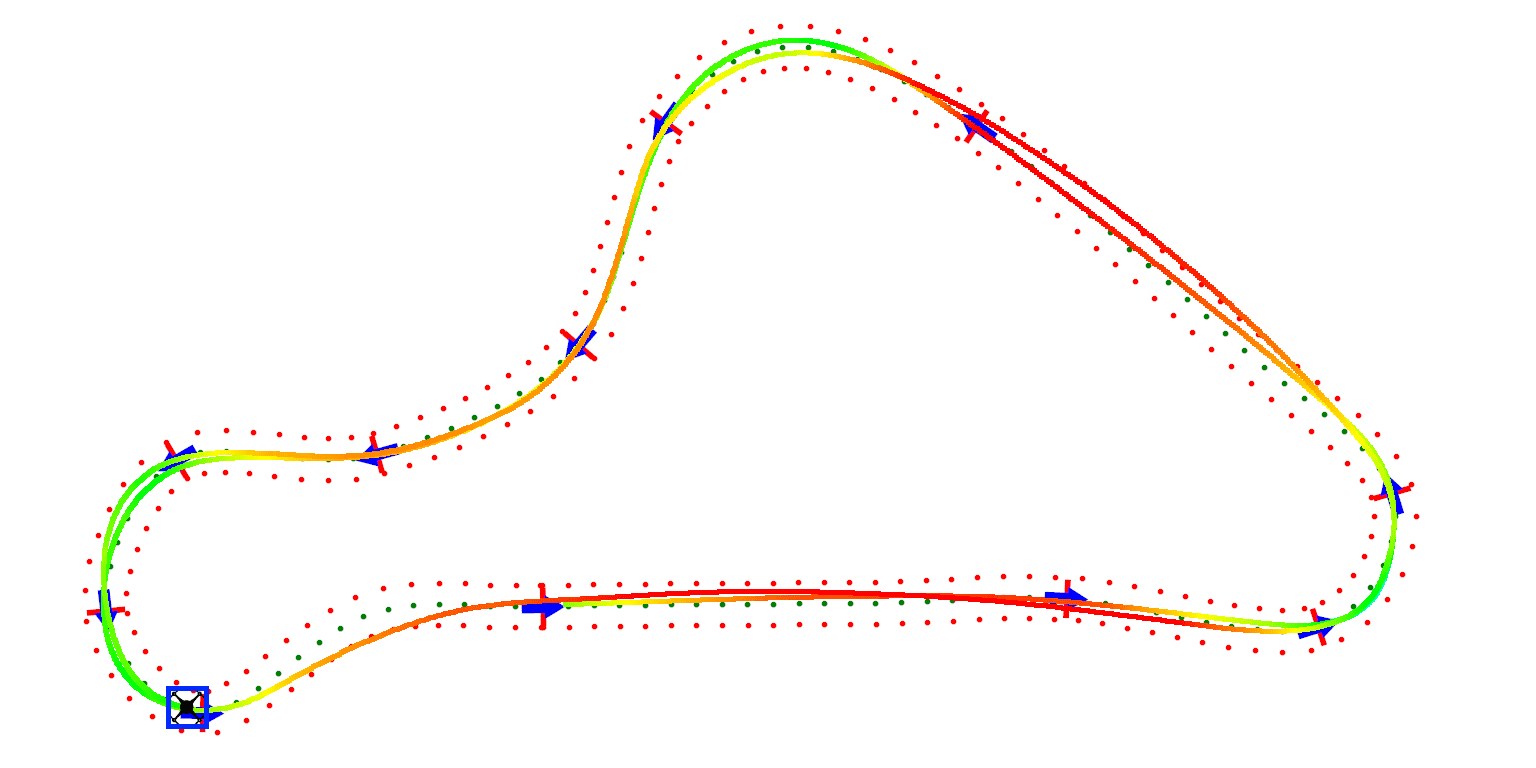}\\
		\small (g) Novice &
		\small (h) Intermediate &
		\small (i) Professional \\
		\includegraphics[height=3cm]{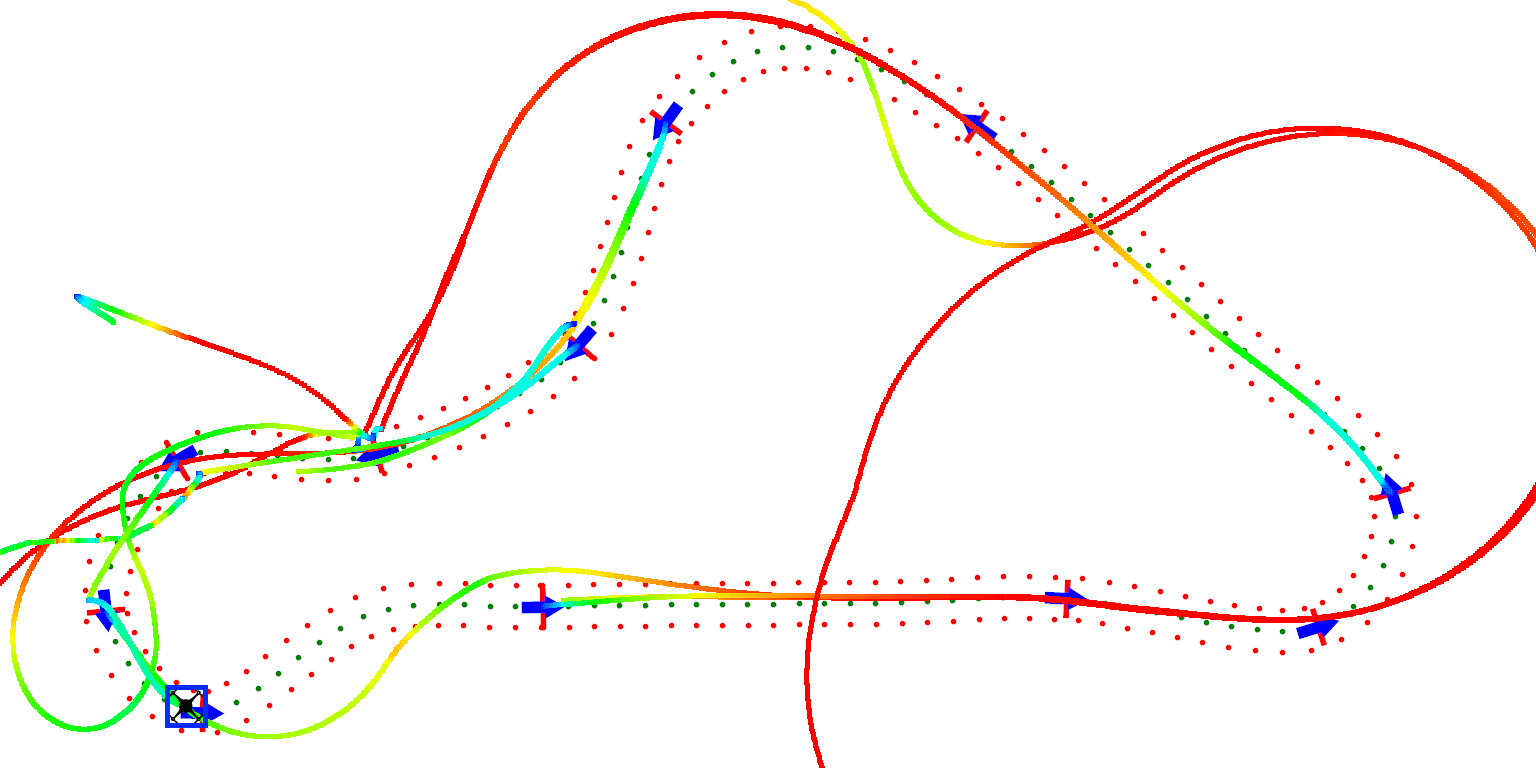} &
		\includegraphics[height=3cm]{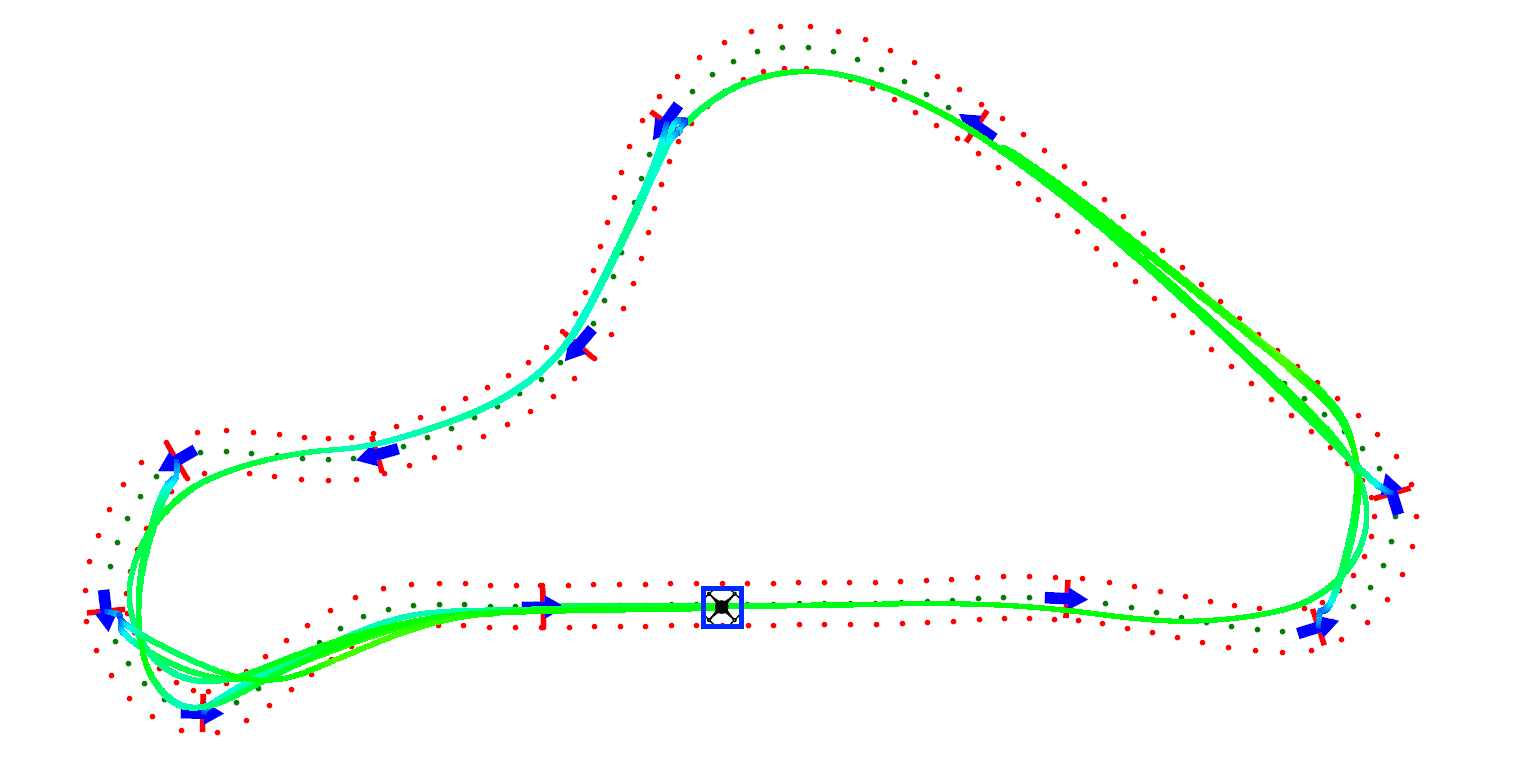} &
		\includegraphics[height=3cm]{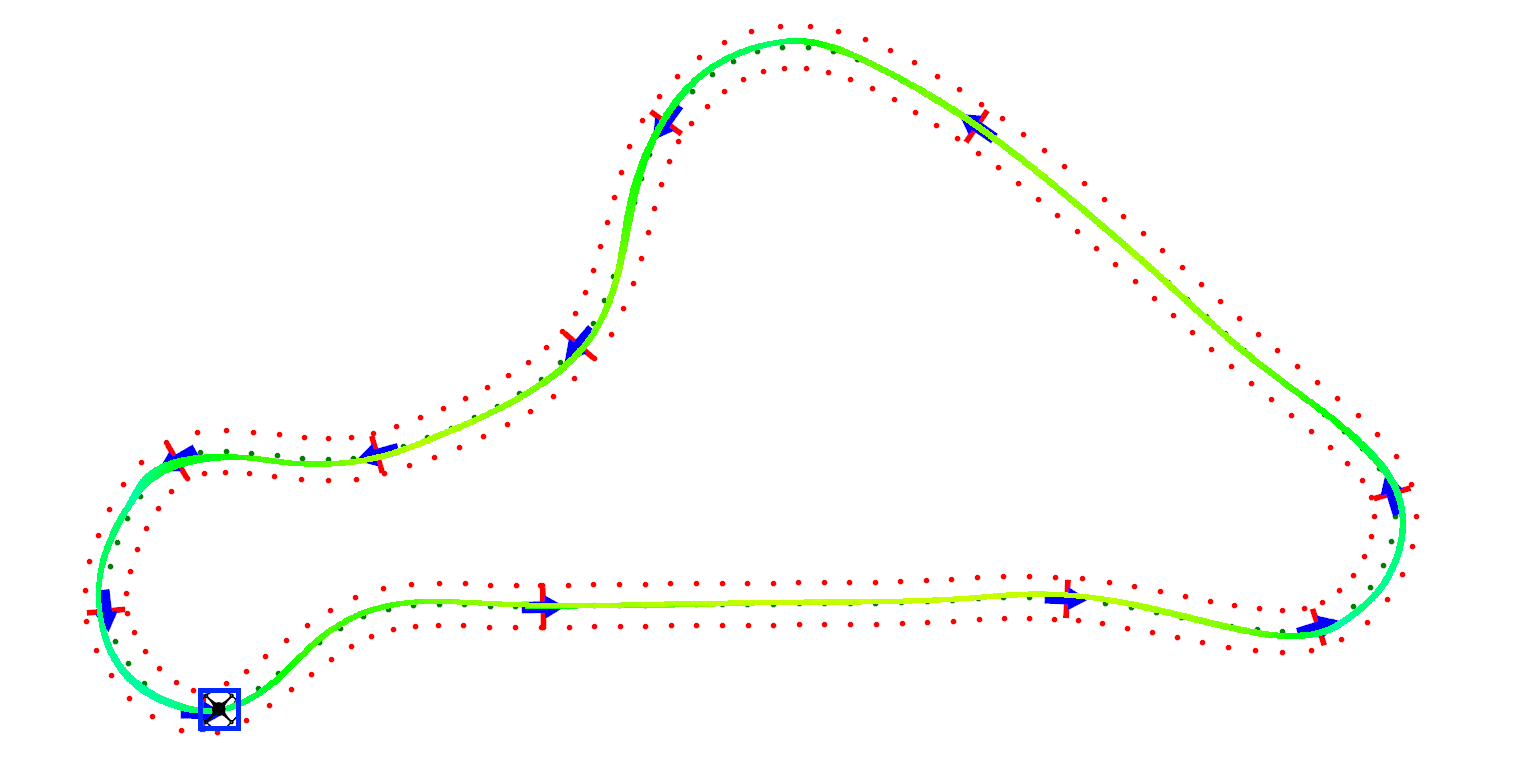}\\
		\small (j) Behaviour Cloning &
		\small (k) Dagger &
		\small (l) DDPG \\
       \multicolumn{3}{c}{\includegraphics[height=1.2cm]{figures/ColorScaleUAV.png}} 
\end{tabular}
\captionof{figure}{Comparison between our learned policy and its teachers (\emph{row 1,2}), human pilots (\emph{row 3}) and baselines (\emph{row 4}) on test track 3. Color encodes speed as a heatmap, where blue is the minimum speed and red is the maximum speed.}
\label{fig:qualitive_results_track3}
\end{figure*}

\begin{figure*}
\centering
\begin{tabular}{@{}c@{\hspace{1mm}}c@{\hspace{1mm}}c@{}}
		\includegraphics[height=3cm]{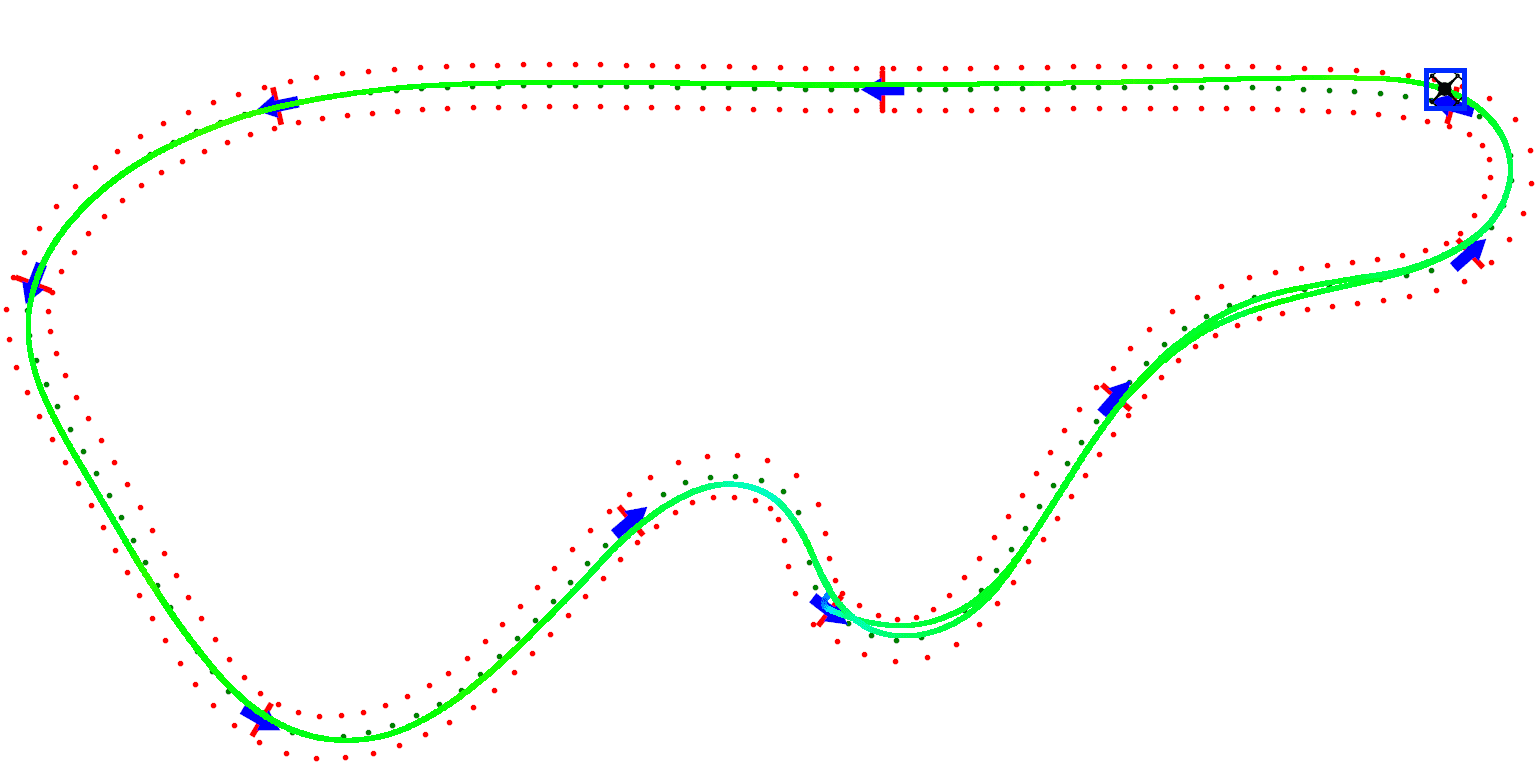} &
		\includegraphics[height=3cm]{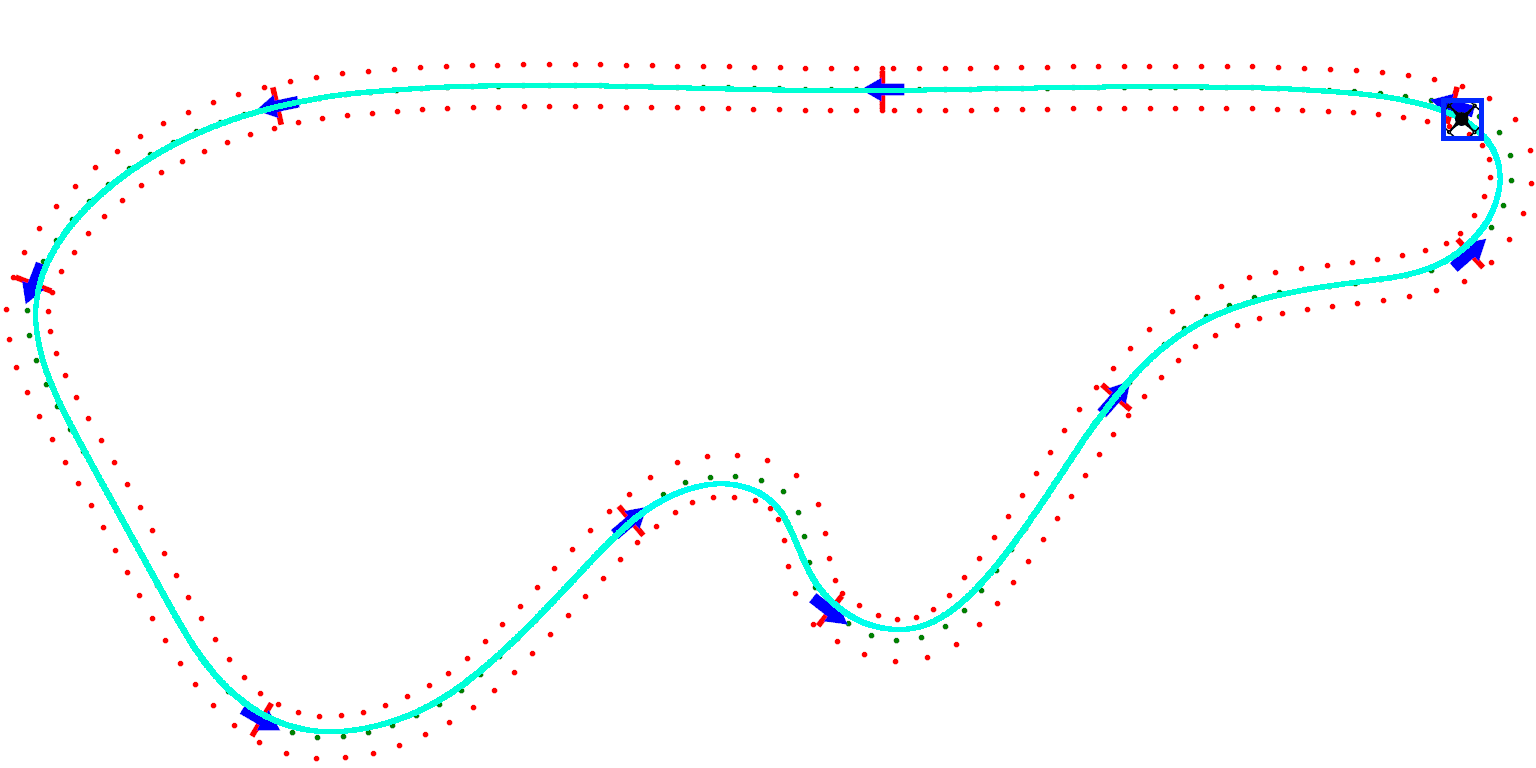} &
		\includegraphics[height=3cm]{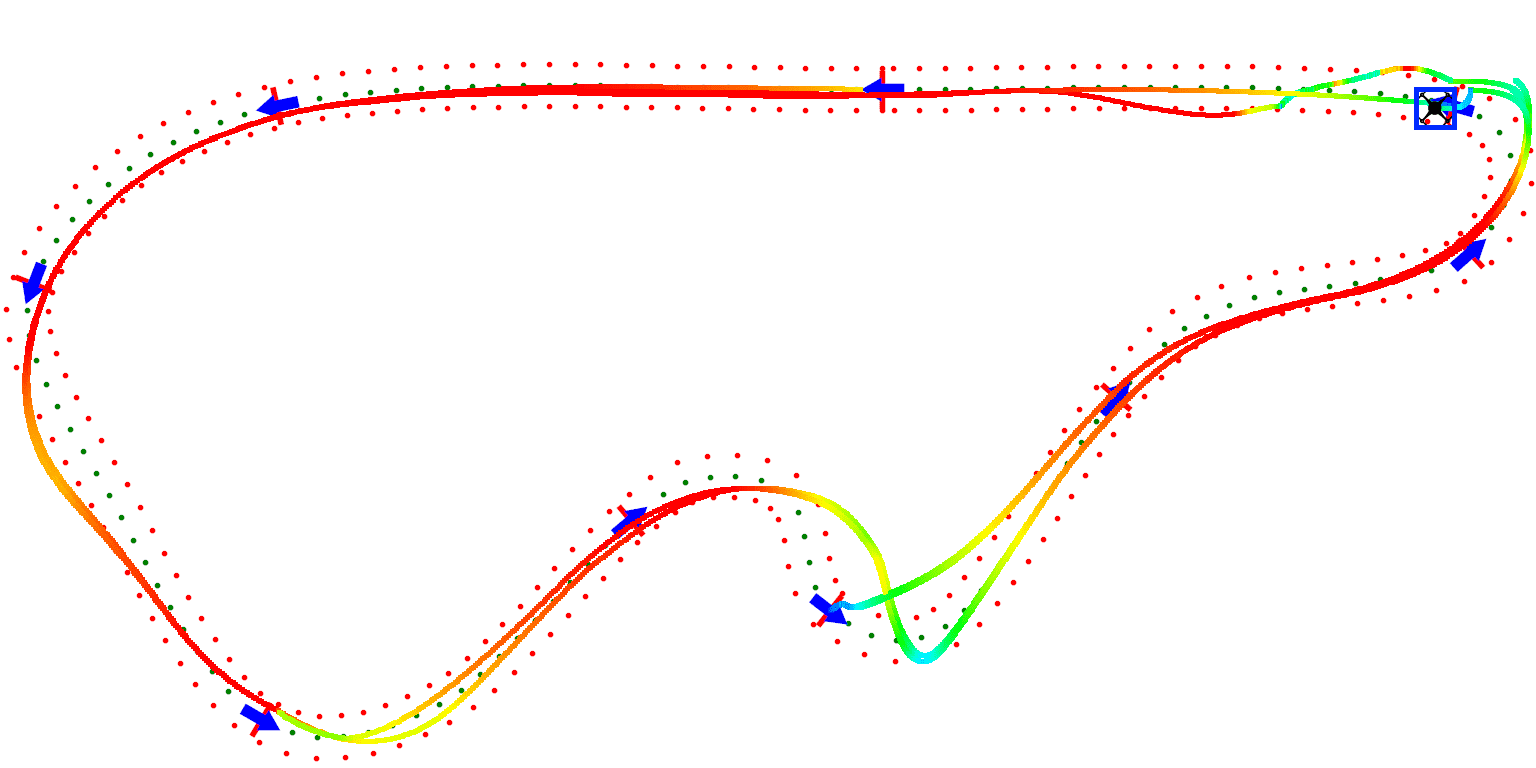} \\
		\small (a) OIL &
		\small (b) Teacher 1 &
		\small (c) Teacher 2 \\
		\includegraphics[height=3cm]{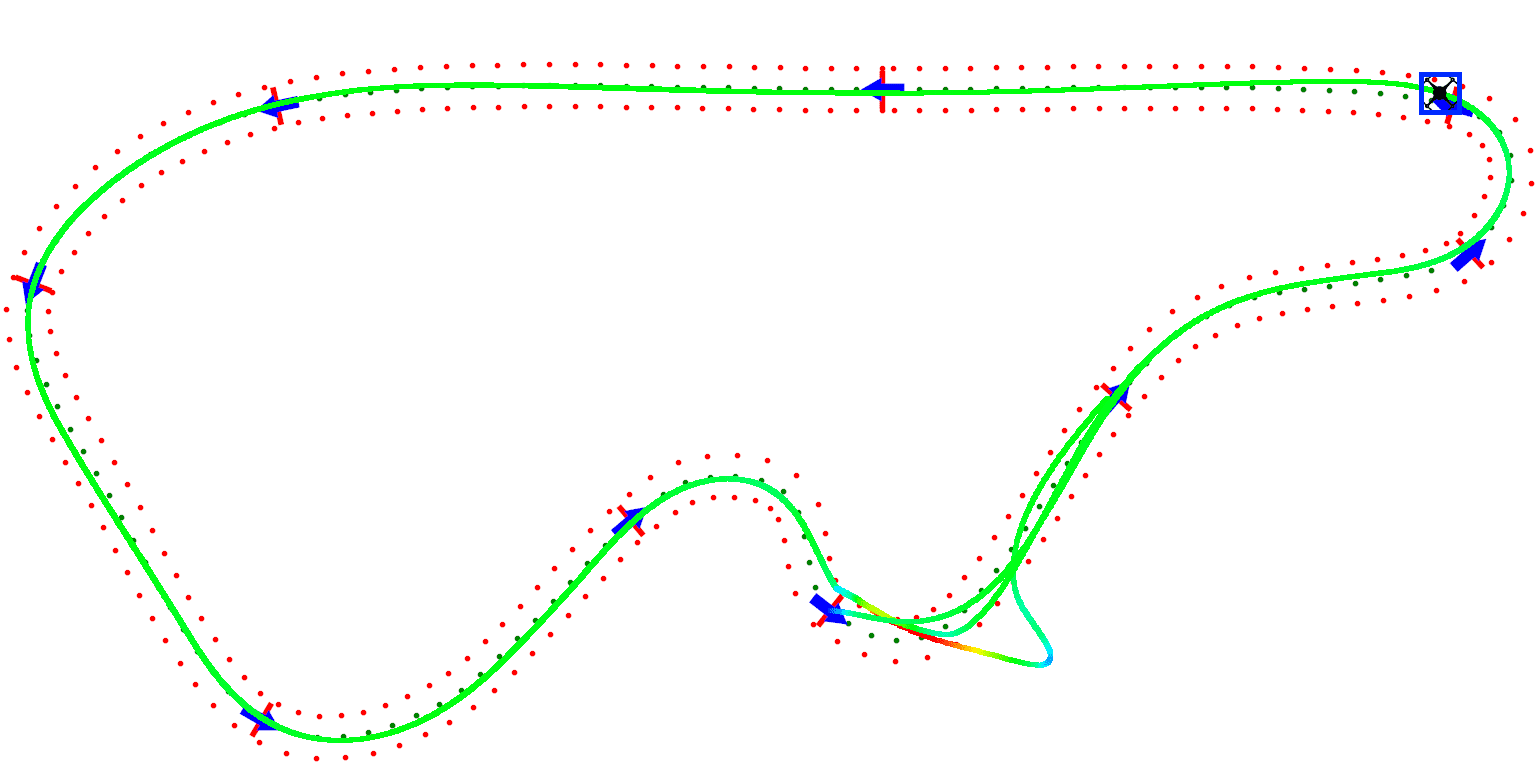} &
		\includegraphics[height=3cm]{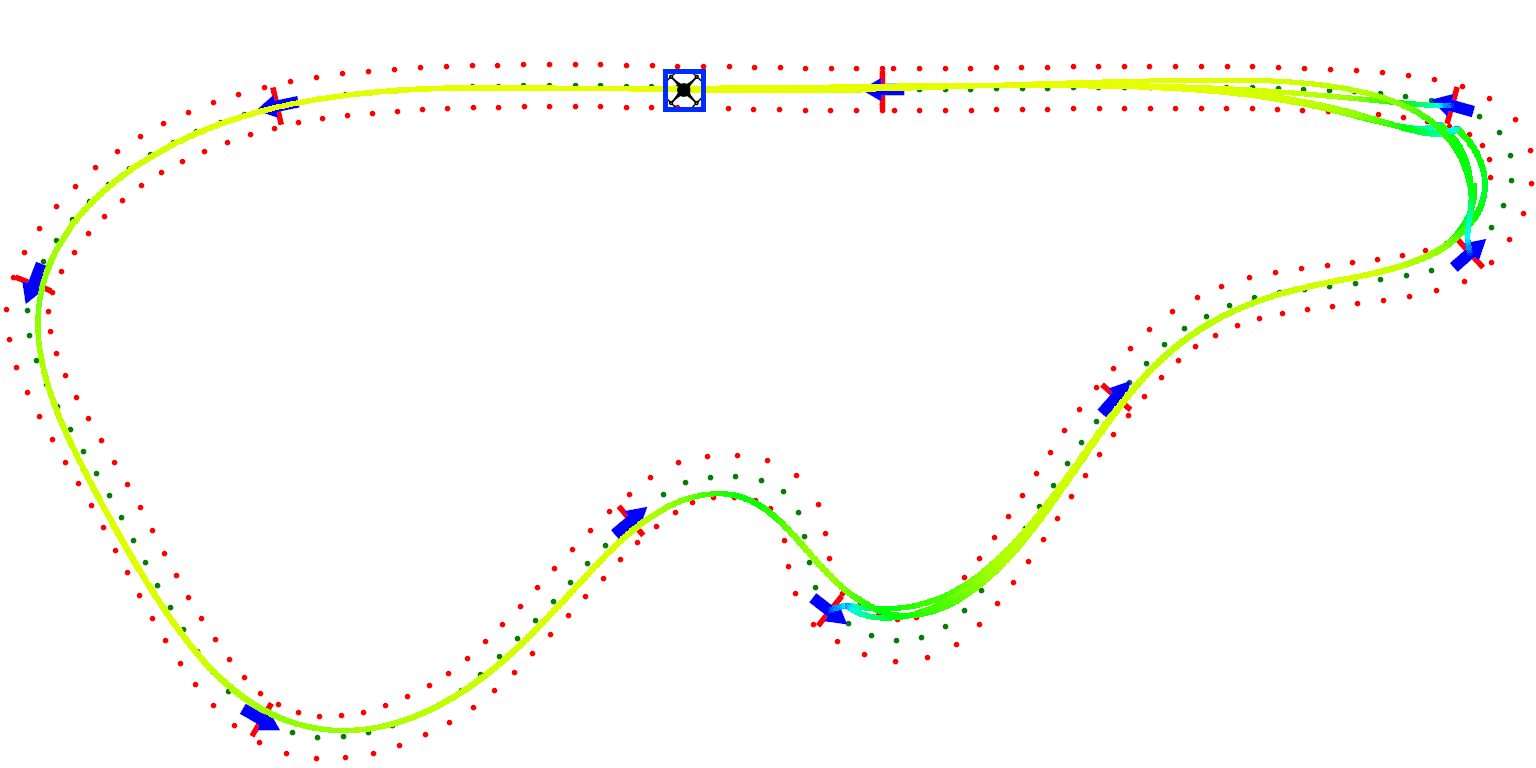} &
		\includegraphics[height=3cm]{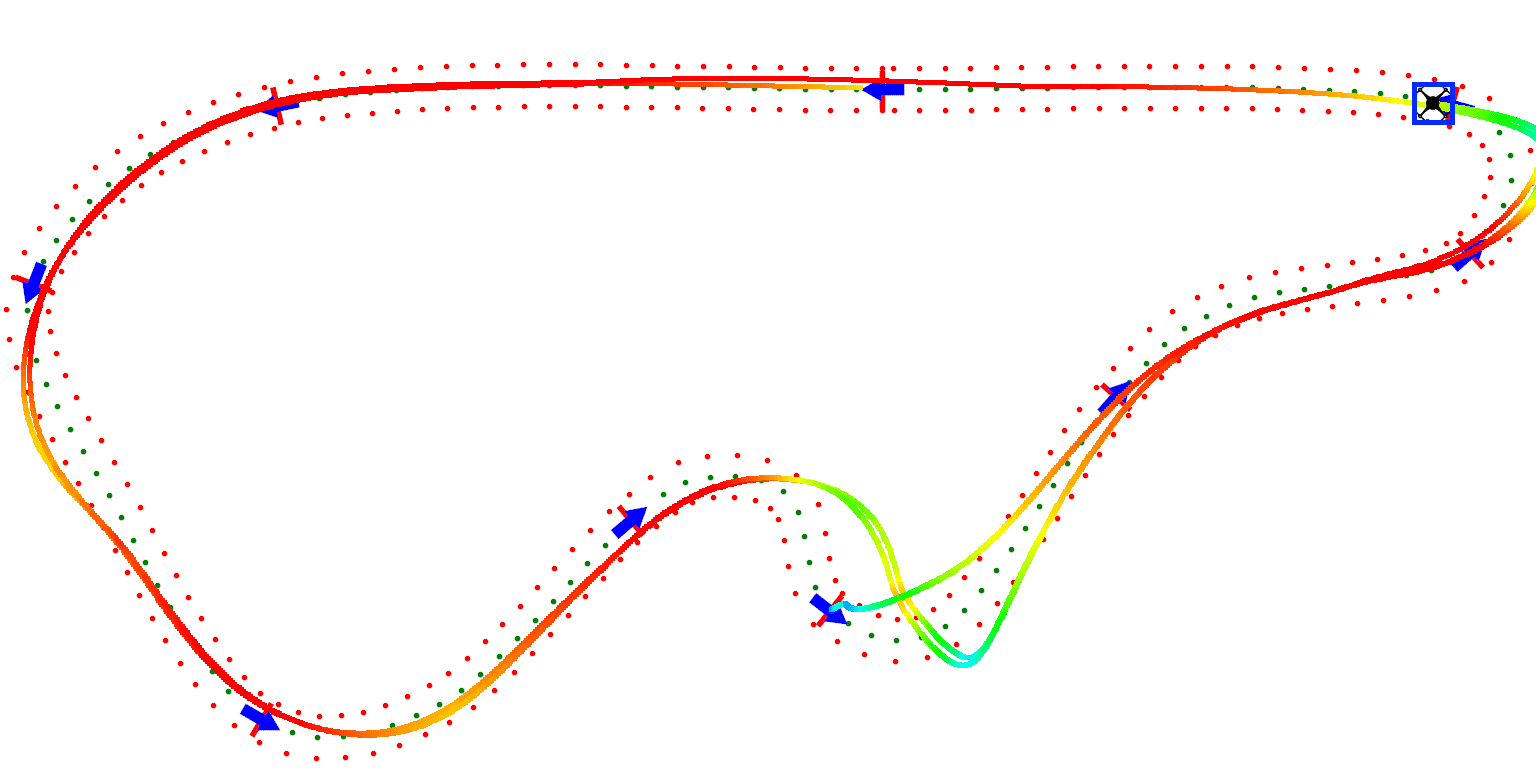} \\
		\small (d) Teacher 3 &
		\small (e) Teacher 4 &
		\small (f) Teacher 5 \\
		\includegraphics[height=3cm]{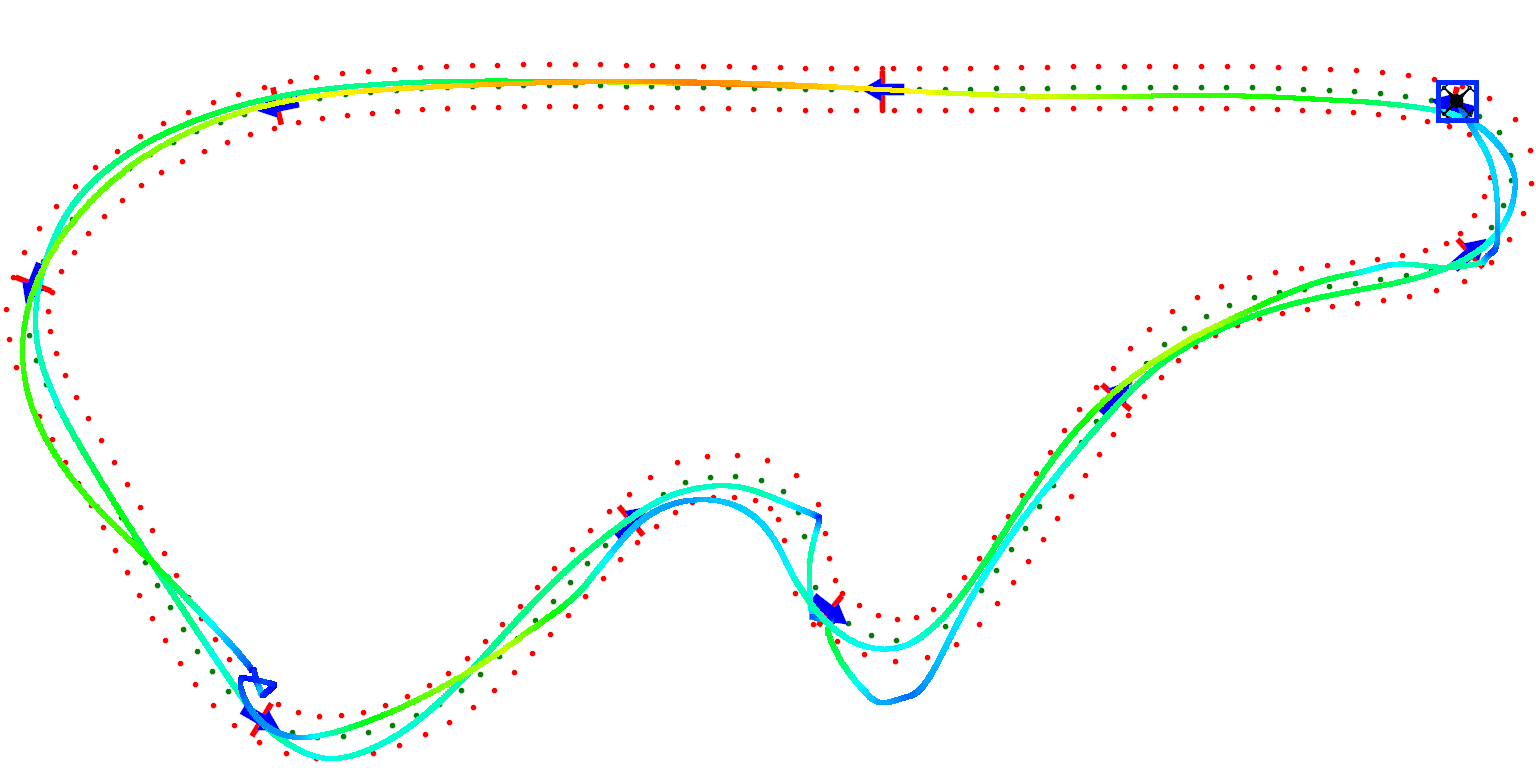} &
		\includegraphics[height=3cm]{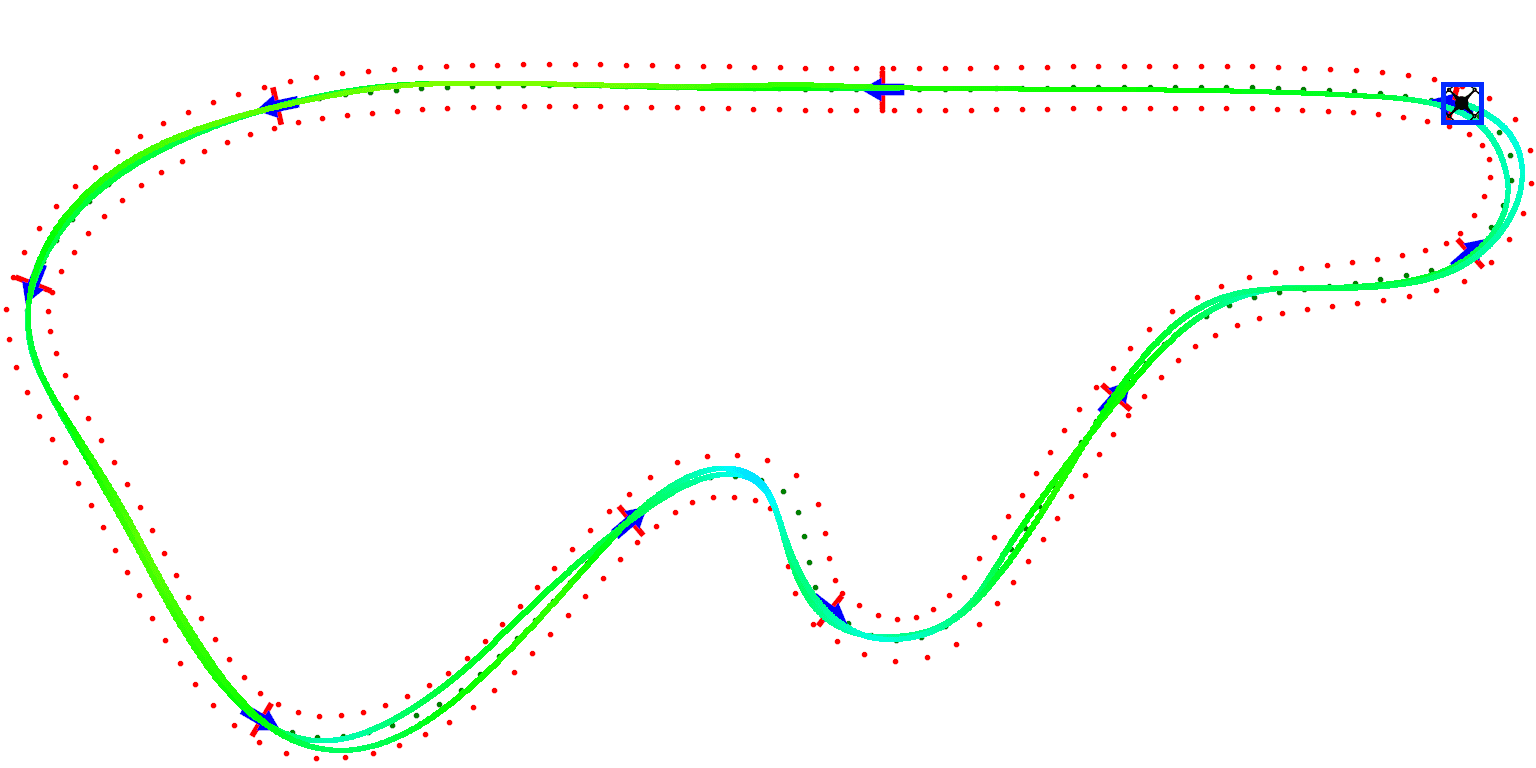} &
		\includegraphics[height=3cm]{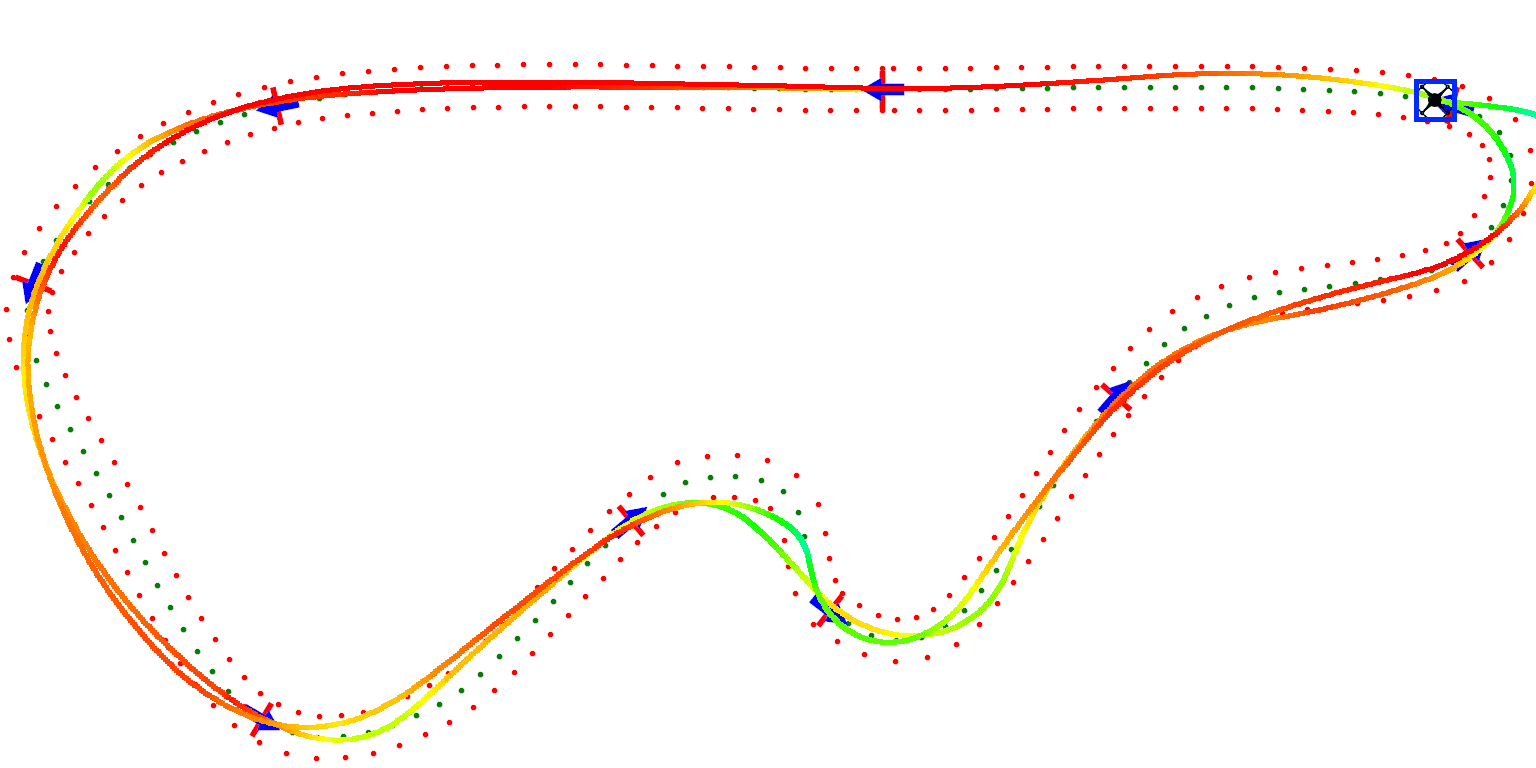}\\
		\small (g) Novice &
		\small (h) Intermediate &
		\small (i) Professional \\
		\includegraphics[height=3cm]{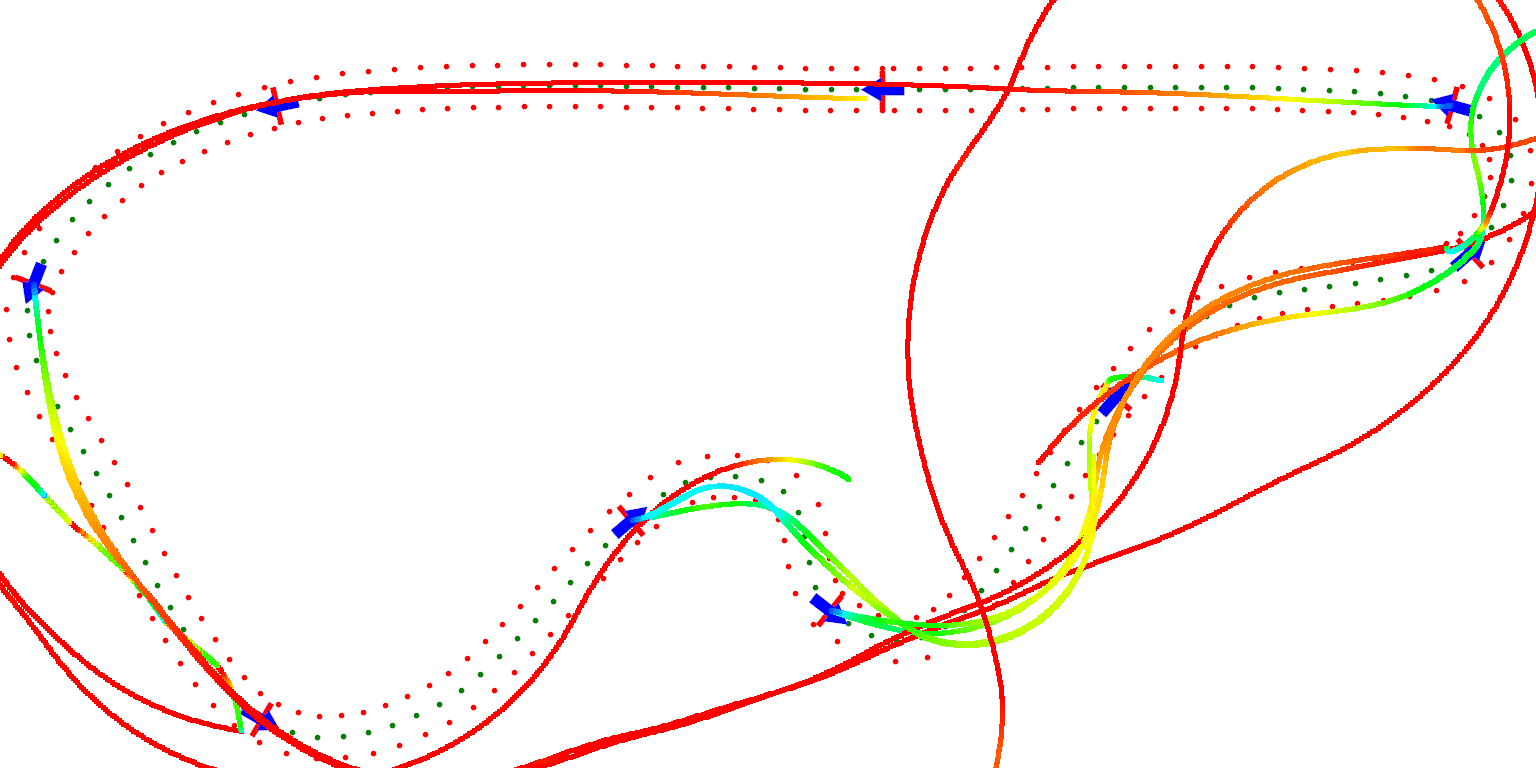} &
		\includegraphics[height=3cm]{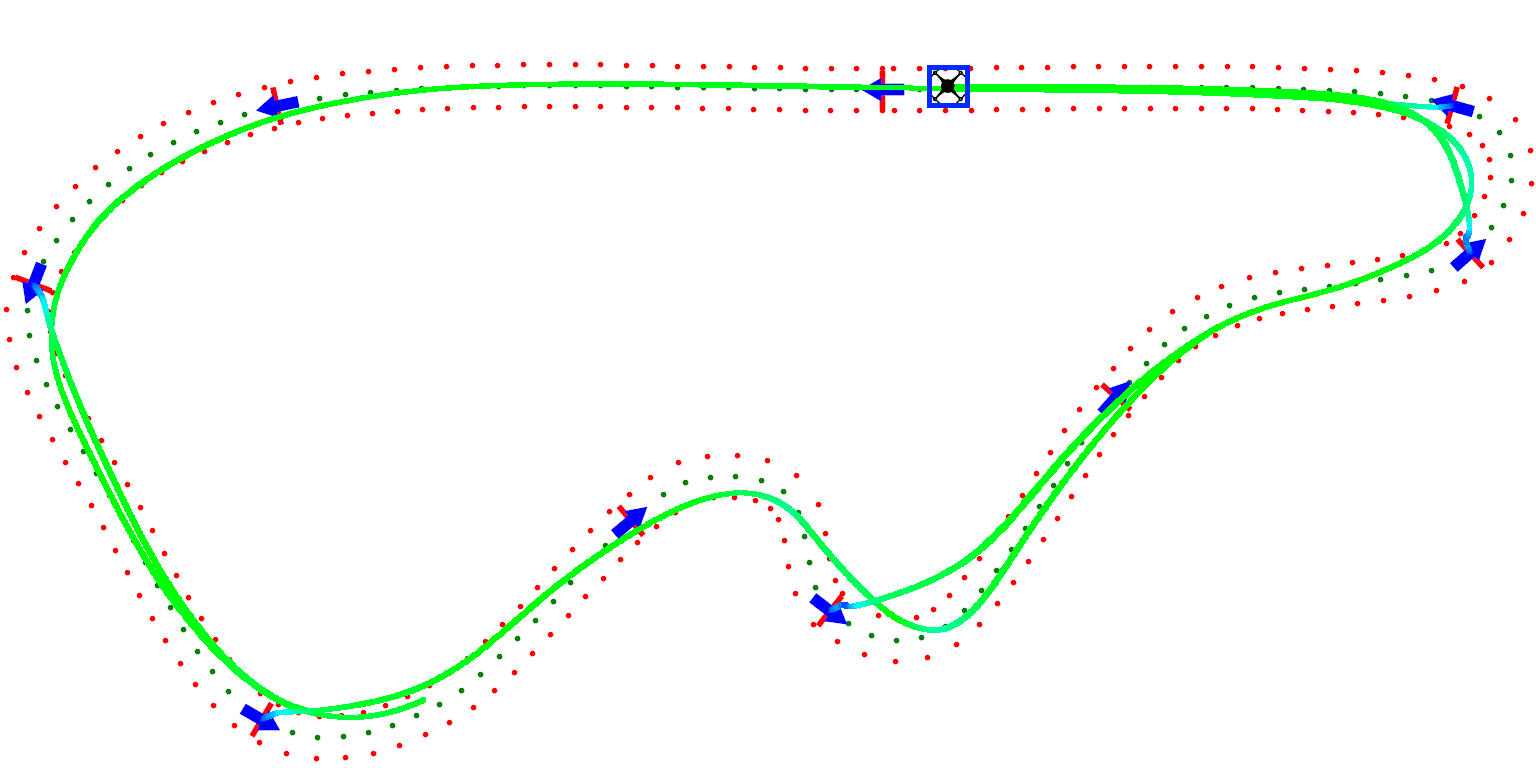} &
		\includegraphics[height=3cm]{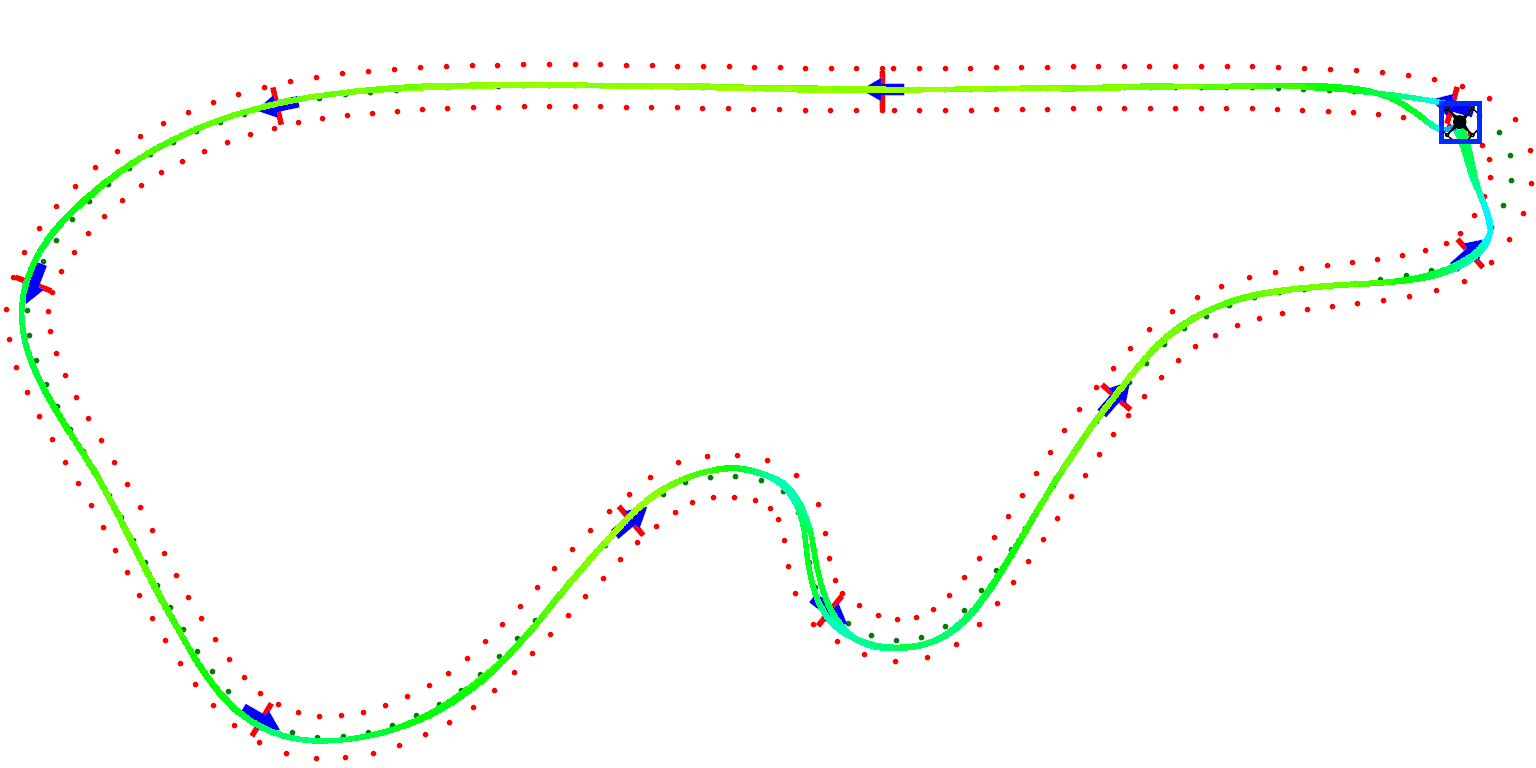}\\
		\small (j) Behaviour Cloning &
		\small (k) Dagger &
		\small (l) DDPG \\
       \multicolumn{3}{c}{\includegraphics[height=1.2cm]{figures/ColorScaleUAV.png}} 
\end{tabular}
\captionof{figure}{Comparison between our learned policy and its teachers (\emph{row 1,2}), human pilots (\emph{row 3}) and baselines (\emph{row 4}) on test track 4. Color encodes speed as a heatmap, where blue is the minimum speed and red is the maximum speed.}
\label{fig:qualitive_results_track4}
\end{figure*}

We show detailed results for the autonomous UAV racing experiments for each track and report the number of missed gates and the time to complete two laps in Tables \ref{tbl: regular baselines uav} and \ref{tbl: learned baselines uav}. In \tblLabel \ref{tbl: regular baselines uav}, we show the results for each of the teachers and provide human baselines. In \tblLabel \ref{tbl: learned baselines uav}, we compare our approach \emph{OIL} to several learned baselines. Figures \ref{fig:qualitive_results_track1}, \ref{fig:qualitive_results_track2}, \ref{fig:qualitive_results_track3} and \ref{fig:qualitive_results_track4} visualize the results.

\mysection{Transfer.~} The modular network architecture abstracts perception and control, and can be applied to many different types of complex vision-based sequential prediction problems. The perception network can be specific to the environment making a transfer of the learnt controller (\eg sim-to-real) easier but predicted trajectories and state estimation will still need to be consistent across environments. The transferability of the perception network can be tested with a conservative PID controller handling controls. The transferability of the control network can be tested using an IMU and differential GPS to determine waypoints and state. 

\clearpage


\begin{table*}[t!]
\scriptsize
\centering
\begin{tabular}{ccccccccccccccccc}
\toprule
\multicolumn{1}{l}{}& \multicolumn{10}{c}{PID}       & \multicolumn{6}{c}{Human}      \\ 
\midrule
\multicolumn{1}{l}{}& \multicolumn{2}{c}{teacher1}          & \multicolumn{2}{c}{teacher2} & \multicolumn{2}{c}{teacher3}     & \multicolumn{2}{c}{teacher4}     & \multicolumn{2}{c}{teacher5}     & \multicolumn{2}{c}{novice}     & \multicolumn{2}{c}{intermediate}          & \multicolumn{2}{c}{expert}     \\
\multicolumn{1}{l}{}& \multicolumn{1}{l}{error} & \multicolumn{1}{l}{time} & \multicolumn{1}{l}{error} & \multicolumn{1}{l}{time} & \multicolumn{1}{l}{error} & \multicolumn{1}{l}{time} & \multicolumn{1}{l}{error} & \multicolumn{1}{l}{time} & \multicolumn{1}{l}{error} & \multicolumn{1}{l}{time} & \multicolumn{1}{l}{error} & \multicolumn{1}{l}{time} & \multicolumn{1}{l}{error} & \multicolumn{1}{l}{time} & \multicolumn{1}{l}{error} & \multicolumn{1}{l}{time} \\ 
\midrule
Track1   & 22.26& 151.62  & 681.59& 113.82  & \textbf{13.61}& 84.30 & 102.21& 83.87  & 664.79& 113.12  & 53.67& 88.67  & 76.09& 64.18  & 34.27& \textbf{56.83}  \\
Track2   & \textbf{25.24}& 140.59  & 371.56& 89.40   & 28.00& \textbf{80.65}  & 67.29& 84.07  & 327.80& 89.00  & 100.89& 114.32  & 87.36& 118.69  & 57.60& 89.83  \\
Track3   & 22.91& 151.80  & 620.70& 116.24  & \textbf{15.57}& 82.98  & 42.49& 58.97  & 830.63& 102.19  & 96.80& 93.12  & 79.40& 75.53  & 50.51& \textbf{57.33}  \\
Track4   & 26.17& 161.57  & 484.35& 120.36  & \textbf{20.95}& 89.90  & 94.83& 79.84  & 450.24& 104.50  & 89.92& 106.74  & 79.63& 94.77  & 53.50& \textbf{77.58}  \\
\midrule
\textbf{Avg.} & 24.14 & 151.39  & 539.55 & 109.95  & \textbf{19.53} & 84.45  & 76.70 & 76.68  & 568.36 & 102.20  & 85.32 & 100.71  &  80.62   & 88.29  & 48.97 & \textbf{70.39}  \\ 
\bottomrule
\end{tabular}
\caption{Regular Baselines - CAR.}
\label{tbl: regular baselines car}
\end{table*}

\begin{table*}[t!]
\scriptsize
\centering
\begin{tabular}{ccccccccccccccc}
\toprule
\multicolumn{1}{l}{}& \multicolumn{4}{c}{Behaviour Cloning}   & \multicolumn{4}{c}{Dagger}      & \multicolumn{2}{c}{DDPG}          & \multicolumn{4}{c}{OIL}   \\ 
\midrule
\multicolumn{1}{l}{}& \multicolumn{2}{c}{best}          & \multicolumn{2}{c}{all}& \multicolumn{2}{c}{best}          & \multicolumn{2}{c}{all}& \multicolumn{2}{c}{random}        & \multicolumn{2}{c}{best}     & \multicolumn{2}{c}{all}       \\
\multicolumn{1}{l}{}& \multicolumn{1}{l}{error} & \multicolumn{1}{l}{time} & \multicolumn{1}{l}{error} & \multicolumn{1}{l}{time} & \multicolumn{1}{l}{error} & \multicolumn{1}{l}{time} & \multicolumn{1}{l}{error} & \multicolumn{1}{l}{time} & \multicolumn{1}{l}{error} & \multicolumn{1}{l}{time} & \multicolumn{1}{l}{error}       & \multicolumn{1}{l}{time} & \multicolumn{1}{l}{error} & \multicolumn{1}{l}{time}        \\ 
\midrule
Track1      & 11.67  & 87.92 & 411.11 & 127.77& 42.31  & 86.37 & 21.77  & 90.35 & 30.98 & 136.11 & \textbf{10.77} & 87.43 & 11.08  & \textbf{74.05} \\
Track2      & 16.57  & 85.38 & 367.69 & 110.20& 32.99  & 86.48 & 35.23  & 78.68 & 93.45 & 135.96 & \textbf{14.58} & 85.68 & 27.93  & \textbf{70.20} \\
Track3      & 11.97  & 86.63 & 325.79 & 115.42& 41.46  & 84.95 & 22.40  & 88.73 & 35.94 & 135.76 & \textbf{10.39} & 86.05 & 11.15  & \textbf{73.53} \\
Track4      & 15.22  & 94.48 & 449.74 & 97.37 & 37.82  & 95.25 & 24.01  & 93.50 & 69.10 & 150.45 & \textbf{13.83} & 94.82 & 20.20  & \textbf{79.08} \\ 
\midrule
\textbf{Avg.} & 13.85  & 88.60 & 388.58 & 112.69& 38.64  & 88.26 & 25.85  & 87.82& 57.37 & 139.57   & \textbf{12.39} & 88.50 & 17.59  & \textbf{74.22} \\ 
\bottomrule
\end{tabular}
\caption{Learned Baselines - CAR.}
\label{tbl: learned baselines car}
\end{table*}

\begin{table*}[t!]
\scriptsize
\centering
\begin{tabular}{ccccccccccccccc}
\toprule
\multicolumn{1}{l}{}& \multicolumn{8}{c}{Trajectory Length}     & \multicolumn{6}{c}{Number of Teachers}    \\ 
\midrule
\multicolumn{1}{l}{}& \multicolumn{2}{c}{N=60}       & \multicolumn{2}{c}{N=180}      & \multicolumn{2}{c}{\textbf{N=300}}      & \multicolumn{2}{c}{N=600}      & \multicolumn{2}{c}{one (3)}    & \multicolumn{2}{c}{three (1,3,4)}         & \multicolumn{2}{c}{\textbf{five (1-5)}} \\
\multicolumn{1}{l}{}& \multicolumn{1}{l}{error} & \multicolumn{1}{l}{time} & \multicolumn{1}{l}{error} & \multicolumn{1}{l}{time} & \multicolumn{1}{l}{error} & \multicolumn{1}{l}{time} & \multicolumn{1}{l}{error} & \multicolumn{1}{l}{time} & \multicolumn{1}{l}{error} & \multicolumn{1}{l}{time} & \multicolumn{1}{l}{error} & \multicolumn{1}{l}{time} & \multicolumn{1}{l}{error} & \multicolumn{1}{l}{time} \\ 
\midrule
Track1&10.21&74.35&12.93&79.33&11.08&74.05&7.48&81.80&10.77&87.43&12.24&73.90&11.08&74.05    \\
Track2&19.18&86.32&16.47&80.91&27.93&70.20&21.95&78.13&14.58&85.68&43.26&70.70&27.93&70.20    \\
Track3&10.66&74.15&17.39&78.03&11.15&73.53&11.08&81.00&10.39&86.05&18.42&72.10&11.15&73.53\\
Track4&21.16&87.12&17.37&84.62&20.20&79.08&14.68&87.77&13.83&94.82&29.26&77.70&20.20&79.08\\ \midrule
\textbf{Avg.}&15.30&80.48&16.04&80.72&17.59&74.22&13.79&82.17&12.39&88.50&25.79&73.60&17.59&74.22    \\ 
\bottomrule
\end{tabular}
\caption{Ablation Study - CAR. The hyperparameter selection for our final model is highlighted in bold.}
\label{tbl: ablation study car}
\end{table*}

\begin{table*}[]
\scriptsize
\centering
\begin{tabular}{ccccccccccccccccc}
\toprule
\multicolumn{1}{l}{}& \multicolumn{10}{c}{PID}       & \multicolumn{6}{c}{Human}      \\ 
\midrule
\multicolumn{1}{l}{}& \multicolumn{2}{c}{teacher1}          & \multicolumn{2}{c}{teacher2} & \multicolumn{2}{c}{teacher3}     & \multicolumn{2}{c}{teacher4}     & \multicolumn{2}{c}{teacher5}     & \multicolumn{2}{c}{novice}     & \multicolumn{2}{c}{intermediate}          & \multicolumn{2}{c}{expert}     \\
\multicolumn{1}{l}{}& \multicolumn{1}{l}{score} & \multicolumn{1}{l}{time} & \multicolumn{1}{l}{score} & \multicolumn{1}{l}{time} & \multicolumn{1}{l}{score} & \multicolumn{1}{l}{time} & \multicolumn{1}{l}{score} & \multicolumn{1}{l}{time} & \multicolumn{1}{l}{score} & \multicolumn{1}{l}{time} & \multicolumn{1}{l}{score} & \multicolumn{1}{l}{time} & \multicolumn{1}{l}{score} & \multicolumn{1}{l}{time} & \multicolumn{1}{l}{score} & \multicolumn{1}{l}{time} \\ 
\midrule
Track1   & 12/12& 130.76  & 12/12& \textbf{40.04}  & 12/12& 81.10  & 12/12& 57.60  & 12/12& 46.30  & 12/12& 87.44  & 12/12& 62.80  & 12/12& 40.50  \\
Track2   & 20/20& 136.19  & 17/20& 77.41  & 20/20& 86.41  & 18/20& 80.95  & 18/20& 71.97  & 20/20& 166.11  & 20/20& 88.21  & 20/20& \textbf{49.23}  \\
Track3   & 22/22& 121.54  & 11/22& 149.45  & 22/22& 77.77  & 16/22& 109.76  & 10/22& 160.42  & 21/22& 118.41  & 22/22& 82.17  & 22/22& \textbf{47.67}  \\
Track4   & 18/18& 139.09  & 15/18& 81.08  & 16/18& 104.99  & 12/18& 112.83  & 10/18& 119.99  & 17/18& 126.47  & 18/18& 91.53  & 18/18& \textbf{50.10}  \\
\midrule
\textbf{Avg.} & \textbf{100\%} & 131.90  & 76.38\% & 87.00  & 97.22\% & 87.57  & 80.56\% & 90.29  & 69.44\% & 99.67  & 97.22\% & 124.61  & \textbf{100\%}    & 81.18  & \textbf{100\%} & \textbf{46.88}  \\ 
\toprule
\end{tabular}
\caption{Regular Baselines - UAV.}
\label{tbl: regular baselines uav}
\end{table*}

\begin{table*}[]
\scriptsize
\centering
\begin{tabular}{ccccccccccccccc}
\toprule
\multicolumn{1}{l}{}& \multicolumn{4}{c}{Behaviour Cloning}  & \multicolumn{4}{c}{Dagger}      & \multicolumn{2}{c}{DDPG}          & \multicolumn{4}{c}{OIL}         \\ 
\midrule
\multicolumn{1}{l}{}& \multicolumn{2}{c}{best}          & \multicolumn{2}{c}{all}       & \multicolumn{2}{c}{best}          & \multicolumn{2}{c}{all}& \multicolumn{2}{c}{random}        & \multicolumn{2}{c}{best}      & \multicolumn{2}{c}{all}     \\
\multicolumn{1}{l}{}& \multicolumn{1}{l}{score} & \multicolumn{1}{l}{time} & \multicolumn{1}{l}{score} & \multicolumn{1}{l}{time}        & \multicolumn{1}{l}{score} & \multicolumn{1}{l}{time} & \multicolumn{1}{l}{score} & \multicolumn{1}{l}{time} & \multicolumn{1}{l}{score} & \multicolumn{1}{l}{time} & \multicolumn{1}{l}{score}       & \multicolumn{1}{l}{time} & \multicolumn{1}{l}{score}       & \multicolumn{1}{l}{time}        \\ \midrule
Track1      & 12/12  & 133.38& 12/12  & \textbf{42.72} & 12/12  & 132.70& 12/12  & 83.45 & 12/12  & 63.48 & 12/12        & 132.81& 12/12        & 78.94        \\
Track2      & 18/20  & 152.04& 15/20  & 109.19       & 20/20  & 138.09& 11/20  & 149.51& 18/20  & 103.97& 20/20        & 138.16& 20/20        & \textbf{82.34} \\
Track3      & 22/22  & 117.54& 15/22  & 127.90       & 22/22  & 123.55& 11/22  & 171.19& 22/22  & 72.96 & 22/22        & 122.97& 22/22        & \textbf{75.52} \\
Track4      & 16/18  & 155.32& 10/18  & 126.45       & 18/18  & 141.89& 8/18   & 156.21& 17/18  & 98.01 & 18/18        & 141.75& 18/18        & \textbf{88.55} \\ 
\midrule
\textbf{Avg.} & 94.44\%& 139.57& 72.22\%& 101.57      & 100\%& 134.05& 58.33\%& 140.09& 95.83\%& 84.61 & \textbf{100\%} & 133.92& \textbf{100\%} & \textbf{81.33} \\ 
\bottomrule
\end{tabular}
\caption{Learned Baselines - UAV.}
\label{tbl: learned baselines uav}
\end{table*}

\end{document}